\documentclass{article}

 \usepackage[main, final]{neurips_2026}
\usepackage[utf8]{inputenc} % allow utf-8 input
\usepackage[T1]{fontenc}    % use 8-bit T1 fonts
\usepackage{hyperref}       % hyperlinks
\usepackage{url}            % simple URL typesetting
\usepackage{booktabs}       % professional-quality tables
\usepackage{amsfonts}       % blackboard math symbols
\usepackage{nicefrac}       % compact symbols for 1/2, etc.
\usepackage{microtype}      % microtypography
\usepackage{xcolor}         % colors
\usepackage{graphicx}

\usepackage{amsmath}
\usepackage{amssymb}
\usepackage{amsthm}
\usepackage{kotex}
\usepackage{todonotes}

\newtheorem{theorem}{Theorem}[section]
\newtheorem{proposition}[theorem]{Proposition}

\newtheorem{corollary}[theorem]{Corollary}

\theoremstyle{definition}
\newtheorem{definition}[theorem]{Definition}

\theoremstyle{remark}

\usepackage{subfigure}
\usepackage{wrapfig}
\usepackage{multicol}
\usepackage{multirow}
\usepackage{graphicx}
\usepackage{tabularx}
\title{Attention Sinks as Spectral Spikes: \\ A Mechanism Analysis of Gated Attention}

\author{%
  Seojin Kim\thanks{Equal contribution.} ~~~ Yehjin Shin\footnotemark[1]  ~~~ Noseong Park\thanks{Corresponding author.} \\
  KAIST \\
  \texttt{\{seojinkim, yehjin.shin, noseong\}@kaist.ac.kr} \\
}

\begin{document}

\maketitle

\begin{abstract}
Transformer attention exhibits attention sinks, where a small subset of source tokens absorbs disproportionate attention mass across layers. Recent work shows that post-SDPA output gating strongly reduces attention-sink behavior, while separate studies link attention sinks to gradient sinks and massive activations through backward training dynamics. Yet how the same gate affects forward sink structure, backward gradient concentration, and training-time representation signatures remains unclear. We address this gap by formulating attention sinks as sink-induced low-rank modes in token-space covariance, with directions aligned to high-mass attention columns. Under this view, post-SDPA gating acts as a mode-dependent contraction: it attenuates sink-induced spectral energy in the forward pass and locally reduces value-path gradients routed through sink columns in the backward pass. Empirically, random-matrix diagnostics show that gating reduces outlier mass, weakens sink-subspace alignment, and increases effective rank, while backward analyses show active-gate attenuation of first-token value gradients. Training-time diagnostics further show that gating suppresses the co-emergence of attention concentration, value-gradient amplification, and representation dominance.
\end{abstract}

\section{Introduction} \label{sec:introduction}
Transformers have become the backbone of modern large language models, with self-attention enabling flexible token interactions across long contexts~\citep{vaswani2017attention}. Yet recent studies have shown that this same mechanism can give rise to attention sinks: a small number of tokens that absorb disproportionate attention mass across layers, often independently of semantic relevance~\citep{xiao2023efficient,gu2025attention}. Such sink behavior is not merely an artifact of attention visualization: it has been linked to distorted representation geometry, massive activations, and training instability~\citep{sun2026spike,sun2024massive}. Understanding and mitigating attention sinks has therefore become an important step toward more stable Transformer architectures~\citep{bu2025value,zhu2025transformers,byun2025bounded,huang2026threshold}. 

Recently, a strikingly simple mitigation was proposed: applying a sigmoid gate after the scaled dot-product attention (SDPA) output~\cite{qiu2025gated}. In its most effective form, this gate is head-specific and elementwise, and it nearly eliminates attention sinks, reducing first-token attention from 46.7\% to 4.8\%, while substantially improving training stability. This mechanism has since been adopted in production-scale architectures such as Qwen3-Next~\citep{qwen3next2025}, underscoring the practical relevance of understanding why it works. The authors attribute this effect to query-dependent sparsity, arguing that sparse gating filters out context irrelevant to the current query. However, this explanation remains largely phenomenological. It leaves open a central question: how can a standard attention-output gate selectively suppress sink-aligned structure, rather than merely shrink the attention output uniformly?

This question is especially important because post-SDPA gating is not merely a post-hoc correction to a trained attention map. It is used throughout training, shaping both the forward representations and the gradients by which they are learned. Its effect therefore appears at multiple coupled levels: it changes the forward representation geometry associated with sink tokens, attenuates sink-directed value-path gradients in the backward dynamics, and suppresses training-time sink signatures. Recent work~\citep{chen2026gradient} explains one part of this picture, showing that attention sinks can induce value-path gradient sinks under causal masking and that massive activations may arise as an RMSNorm-mediated response to this localized gradient pressure. However, this backward account has not yet been connected to the standard post-SDPA gate.

We bridge this gap by recasting an attention sink as a sink-induced low-rank mode in token-space covariance. Under this view, a sink is not merely a token that receives large attention mass; rather, a high-mass attention column can induce a structured token-space direction in the post-attention representation. Post-SDPA gating then has a precise structural interpretation: it induces an effective contraction of this sink-induced mode. 
Under an idealized isotropic reference model, this contraction predicts that a supercritical spike can be pushed below the Baik--Ben Arous--Péché (BBP) threshold and merge back into the Marchenko--Pastur bulk. In the backward pass, the gate locally attenuates value-path gradients routed through sink columns, replacing original sink-column statistics with gated effective counterparts. Thus, post-SDPA gating is not simply a generic damping operation; it attenuates sink-induced structure in both forward spectral geometry and backward gradient flow.

Taken together, we argue that post-SDPA gating goes beyond generic sparsity or uniform attenuation: it weakens sink-induced token-space structure in the forward pass and locally reduces value-path gradient flow through high-mass sink columns. We make the following contributions:

\begin{itemize}
    \item \textbf{Spectral characterization of attention sinks.}
    We formulate attention sinks as sink-induced low-rank modes in token-space covariance, with directions aligned to high-mass attention columns, connecting attention concentration to spectral outliers and providing a random-matrix perspective on sink emergence.

    \item \textbf{Dual contraction mechanism of post-SDPA gating.}
    We show that post-SDPA gating suppresses sink-induced structure in two coupled ways: it attenuates forward spectral energy  and it locally contracts value-path gradient aggregation through sink columns.

    \item \textbf{Training-time sink signatures.} 
    We show that models without gating develop first-token attention concentration and first-token representation dominance during training, whereas post-SDPA gated models substantially attenuate both signatures.
\end{itemize}

\section{From Attention Sinks to Spectral Spikes}
This section recasts attention sinks as spectral objects in the post-attention representation. We first connect high-mass attention columns to token-space covariance directions, then introduce a sink-induced spiked model and the post-SDPA gate as a row-wise diagonal contraction.

\subsection{Attention Sinks as Column Imbalance} \label{sec:prelim_attention_sink}
Transformer attention computes token interactions through non-negative, row-stochastic attention weights. For a single attention head, let $Q,K,V\in\mathbb{R}^{n\times d}$ denote the query, key, and value matrices. Under a causal mask, the attention weights $A$ and corresponding output $Y$ are defined as follows:
\begin{equation}
A_{ij}
=
\begin{cases}
\displaystyle
\frac{
\exp(q_i^\top k_j/\sqrt{d})
}{
\sum_{\ell\le i}\exp(q_i^\top k_\ell/\sqrt{d})
},
& j\le i, \\[1.2ex]
0, & j>i,
\end{cases}
\qquad
y_i = \sum_{j\le i} A_{ij}v_j,
\qquad
Y = AV \in \mathbb{R}^{n \times d}.
\end{equation}
Thus \(A_{ij}\ge 0\) and $\sum_{j\le i} A_{ij}=1$. Due to the non-negativity and row-wise normalization of the softmax, attention mass must be assigned to some previous token even when no semantically relevant context is needed. As a result, certain source tokens can accumulate attention mass across many query positions and layers. This phenomenon is known as an \emph{attention sink}. 

Empirically, early-position tokens, especially the first token, often act as sinks, receiving disproportionately large attention weights independent of their semantic relevance. Formally, we measure the incoming attention mass of a source token \(j\) by its causal column mass, the average attention it receives over all \(n\) query positions:
\begin{equation}
    m_j = \frac{1}{n}\sum_{i=j}^{n} A_{ij}.
    \label{eq:column_mass}
\end{equation}
A token \(j\) is referred to as an attention sink when \(m_j\) is significantly larger than the typical non-sink scale. Large values of \(m_j\) indicate a column-wise imbalance in the attention matrix, reflecting persistent aggregation of attention toward a small number of source positions. While \(m_j\) identifies where attention mass concentrates in the map \(A\), our goal is to understand how this column imbalance appears in the representation \(Y = AV\).

\subsection{From Sink Columns to Token-Space Covariance}
\label{sec:sink_columns_covariance}

The standard definition of an attention sink is based on column-wise concentration in the attention matrix. To study how this phenomenon affects representations, we shift attention from the map \(A\) to the token-space covariance of the post-attention output:
\begin{equation}
    C_Y
    :=
    \frac{1}{d}YY^\top
    =
    \frac{1}{d}AVV^\top A^\top,
    \qquad
    Y=AV\in\mathbb{R}^{n\times d}.
\end{equation}
Because \(C_Y\) is a token-space covariance, its leading eigenvectors describe token-position modes of concentrated representation energy. Under an isotropic value approximation, 
\(\mathbb{E}[VV^\top/d\mid A]=\sigma_V^2 I\), and therefore
\begin{equation}
    \mathbb{E}[C_Y\mid A]
    =
    \frac{1}{d}A\,\mathbb{E}[VV^\top\mid A]\,A^\top
    =
    \sigma_V^2 AA^\top
    =
    \sigma_V^2\sum_j A_{:,j}A_{:,j}^\top .
\end{equation}
This column-wise decomposition identifies the directions through which attention sinks can enter the token-space covariance. When a source token \(j\) has high incoming attention mass, its attention column contributes the rank-one term \(A_{:,j}A_{:,j}^{\top}\), whose direction is the normalized column
\(
    \bar a_j := {A_{:,j}}/{\|A_{:,j}\|_2}.
\)
This observation provides the bridge from attention-map concentration to representation geometry: a persistent sink column can appear as a dominant token-space mode in \(C_Y\).

\subsection{Spectral Spikes and Sink-Induced Representations}
\label{sec:rmt_preliminary}

The covariance view above reduces sink structure to the question of whether \(C_Y\) contains a dominant token-space direction aligned with high-mass attention columns. Random matrix theory provides a baseline for separating such structured directions from high-dimensional noise. When the representation is purely isotropic noise, we write \(Y=\sigma Z\), where \(Z\in\mathbb{R}^{n\times d}\) has i.i.d. standard Gaussian entries. Then the spectrum of \(C_Y\) converges, in the high-dimensional limit \(n,d\to\infty\) with \(n/d\to c\), to the Marchenko--Pastur (MP) law~\citep{marvcenko1967distribution}, whose support is given by
\begin{equation}
    \lambda_\pm = \sigma^2(1\pm\sqrt{c})^2,
    \qquad c=\frac{n}{d}.
\end{equation}
Eigenvalues within this bulk are consistent with isotropic noise, whereas eigenvalues separated above \(\lambda_+\) indicate excess structured variance. A standard way to represent such structure is a rank-one spiked model~\cite{johnstone2001distribution,paul2007asymptotics,johnstone2009consistency},
\begin{equation} \label{eq:rank_one_spike}
    Y = \sqrt{d}\,a s u^\top + \sigma Z,
\end{equation}
where \(s\in\mathbb{R}^{n}\) and \(u\in\mathbb{R}^{d}\) are unit vectors, and \(a>0\) is the spike amplitude. Here \(s\) specifies a token-space direction and \(u\) specifies its feature-space direction. Under the Baik--Ben Arous--Péché (BBP) transition \citep{baik2005phase,benaych2012singular}, the rank-one component produces a separated spectral outlier when $a^2 > \sigma^2\sqrt{c}$.
Thus, the MP edge serves as a noise threshold, and a spectral outlier signals a detectable low-rank mode in the token representations.

We now specialize this generic spiked model to attention sinks. From the column-covariance relation in Section~\ref{sec:sink_columns_covariance}, high-mass sink columns induce token-space directions \(\bar a_j\). We therefore identify sink-induced spikes by the alignment of the spike direction \(s\) with the subspace spanned by these directions.

\begin{definition}[Sink-induced spiked representation model]
\label{def:sink_induced_spike}
Consider a rank-one spiked post-attention representation
\[
    Y = \sqrt{d}\,a s u^\top + \sigma Z,
\]
where \(s\in\mathbb{R}^{n}\) and \(u\in\mathbb{R}^{d}\) are unit vectors, \(a>0\) is the spike amplitude, and \(Z\) is isotropic noise. 
For sink tokens \(J_{\mathrm{sink}}\), let
\(
    \mathcal{S}_{\mathrm{sink}}
    =
    \operatorname{span}\{\bar a_j:j\in J_{\mathrm{sink}}\},
\)
and let \(P_{\mathrm{sink}}\) be the orthogonal projector onto this subspace. We call the spike sink-induced if
\[
    \|P_{\mathrm{sink}}s\|_2^2 \ge 1-\varepsilon .
\]
For a single first-token sink, this becomes \(\langle s,\bar a_1\rangle^2\ge 1-\varepsilon\).
\end{definition}

This model connects attention-map concentration with representation geometry: a persistent sink column can manifest as a low-rank token-space mode in \(C_Y\). When this mode is supercritical, it appears as a spectral outlier above the MP bulk.

\subsection{Post-SDPA Gating}
\label{sec:post_sdpa_gating}

A recent line of work~\citep{qiu2025gated} mitigates attention sinks by applying a query-dependent multiplicative gate to the output of scaled dot-product attention. We describe the post-SDPA gated attention variant for a single attention head and write its head dimension as \(d\). The attention output is transformed as
\begin{equation}
    \widetilde y_i
    =
    G_i \odot y_i,
    \qquad
    G_i = \sigma(x_i W_g),
    \qquad
    G_i\in[0,1]^d,
    \qquad
    \widetilde Y = Y\odot G,
\end{equation}
where \(x_i\) denotes the hidden state used to compute the gate, \(W_g\) is a learned gating projection, \(\sigma\) is the sigmoid function, \(\odot\) denotes element-wise multiplication, and \(G\in[0,1]^{n\times d}\) collects the gate values row-wise. Equivalently, with \(\Gamma_i:=\operatorname{diag}(G_i)\), we have
\begin{equation}
    \widetilde y_i = \Gamma_i y_i,
    \qquad
    0\preceq \Gamma_i\preceq I ,
\end{equation}
so the gate acts on each token row as a diagonal contraction. The post-SDPA gate is therefore not a scalar shrinkage of \(Y\), but a token- and feature-dependent diagonal contraction. Its effect on a sink-induced spike depends on how the spike's token positions and feature direction interact with the row-wise operators \(\Gamma_i\).

\section{Forward Pass: Spectral Suppression} \label{sec:forward_spectral}
\subsection{Theory: Sink-mode Contraction under Post-SDPA Gating}
We analyze the effect of post-SDPA gating on a sink-induced spectral mode. Let
\(Y_{\mathrm{sig}}=\sqrt d\,a s u^\top\)
denote a sink-induced signal component of the SDPA output, where \(s\in\mathbb{R}^n\) is aligned with the sink-column subspace in Definition~\ref{def:sink_induced_spike} and \(u\in\mathbb{R}^d\) is the corresponding feature direction. The analysis is deterministic and makes no distributional assumption on the learned gate.

Using the row-wise gate operators \(\Gamma_i\) defined in Section~\ref{sec:post_sdpa_gating}, define the effective gate on the feature direction \(u\) by
\begin{equation}
    g_i
    :=
    \|\Gamma_i u\|_2,
    \qquad
    D
    :=
    \operatorname{diag}(g_1,\ldots,g_n).
\end{equation}
Since \(\|u\|_2=1\) and \(0\preceq \Gamma_i\preceq I\), we have \(0\le g_i\le 1\).

\begin{theorem}[Sink-induced mode contraction under post-SDPA gating]
\label{thm:sink_mode_contraction}
For the sink-induced signal \(Y_{\mathrm{sig}}=\sqrt d\,a s u^\top\), the gated signal satisfies
\begin{equation}
    \frac{1}{d}
    \|\widetilde Y_{\mathrm{sig}}\|_F^2
    =
    a^2\|Ds\|_2^2
    =
    a^2\alpha_s
    \le
    a^2,
\end{equation}
where
\begin{equation}
    \alpha_s
    :=
    \|Ds\|_2^2
    =
    \sum_{i=1}^{n}s_i^2\|\Gamma_i u\|_2^2 .
\end{equation}
Moreover, the gated signal covariance
\(
    \widetilde C_{\mathrm{sig}}
    :=
    \frac{1}{d}
    \widetilde Y_{\mathrm{sig}}\widetilde Y_{\mathrm{sig}}^\top
\)
satisfies
\begin{equation}
    \lambda_{\max}(\widetilde C_{\mathrm{sig}})
    \le
    \operatorname{tr}(\widetilde C_{\mathrm{sig}})
    =
    a^2\alpha_s .
\end{equation}
\end{theorem}

Theorem~\ref{thm:sink_mode_contraction} shows that \(\alpha_s\) quantifies how much of the sink-induced mode is preserved by the gate. Thus, gating suppresses a sink-induced spectral spike when it contracts the query positions and feature direction participating in the corresponding attention-column mode.

To connect this contraction to spectral outlier behavior, we use the classical spiked-covariance model as an idealized reference. The deterministic contraction above applies to learned gates, but a general learned gate also transforms the noise component and therefore does not preserve the exact assumptions of a standard spiked Wishart model. We therefore isolate a simplified isotropic thinning setting in which the same contraction mechanism can be related exactly to the BBP phase transition.

\begin{proposition}[BBP suppression under isotropic thinning]
\label{prop:bbp_suppression_thinning}
Assume the spiked representation model in Eq.~\ref{eq:rank_one_spike}.
Let \(P_{\mathcal I}\in\{0,1\}^{n'\times n}\) be a row-selection matrix independent of \(Z\), where \(n'=|\mathcal I|\), and define $\alpha_s:=\|P_{\mathcal I}s\|_2^2$, $c':=\frac{n'}{d}$.
If \(\alpha_s>0\), then
\begin{equation}
    P_{\mathcal I}Y
    =
    \sqrt da\sqrt{\alpha_s}\bar s u^\top
    +
    \sigma Z',
    \qquad
    \bar s
    :=
    \frac{P_{\mathcal I}s}{\|P_{\mathcal I}s\|_2},
\end{equation}
where \(Z'\in\mathbb{R}^{n'\times d}\) has i.i.d. standard Gaussian entries.
Hence the thinned covariance
\begin{equation}
    \widetilde C
    =
    \frac{1}{d}
    (P_{\mathcal I}Y)(P_{\mathcal I}Y)^\top
\end{equation}
is a rank-one spiked Wishart model with effective spike strength \(a^2\alpha_s\). Consequently, the sink-induced spectral outlier disappears when
\begin{equation}
    a^2\alpha_s
    \le
    \sigma^2\sqrt{c'}.
\end{equation}
When \(a^2\alpha_s>\sigma^2\sqrt{c'}\), a spectral outlier persists, but its effective spike strength is reduced to \(a^2\alpha_s\).
\end{proposition}
Proposition~\ref{prop:bbp_suppression_thinning} should be read as an idealized spectral reference rather than an exact theorem for learned gates. For a general post-SDPA gate, the sink-induced signal component is contracted as in Theorem~\ref{thm:sink_mode_contraction}, but the noise component is also modulated by the row-wise feature gates \(\Gamma_i\). The resulting covariance is generally anisotropic and need not be a standard spiked Wishart matrix. Nevertheless, the BBP reference captures the mechanism-level prediction: when gating reduces the effective strength of a sink-induced mode, the associated spectral outlier should weaken, move closer to the MP bulk, or disappear. 
We empirically demonstrate that this happens in the learned post-SDPA gate setting. 
Proofs of Theorem~\ref{thm:sink_mode_contraction} and Proposition~\ref{prop:bbp_suppression_thinning} can be found in Appendix~\ref{app:spectral_proofs}.

\subsection{Empirical: Spectral Suppression and Sink-Subspace Alignment}
\label{sec:forward_empirical}

We empirically test the forward-pass prediction of Theorem~\ref{thm:sink_mode_contraction}: if post-SDPA gating contracts structured token-space modes, then the post-gate representation should exhibit weaker spectral concentration. In addition, following Definition~\ref{def:sink_induced_spike}, we ask whether the suppressed spectral structure is aligned with sink-induced attention-column subspaces.

\paragraph{Analysis setting.}
We evaluate the post-SDPA gated Qwen-1B model following previous work~\citep{qiu2025gated} and record both the pre-gate output \(Y_{\mathrm{raw}}\) and the post-gate output \(Y_{\mathrm{gate}}\) at each layer. All spectral statistics are computed layer-wise and averaged across sixteen samples of sequence length 512 from WikiText dataset ~\citep{wikitext}.

For each layer, we compute the token-space covariance
\(
    C_Y
    =
    \frac{1}{D}\bar Y\bar Y^\top,
\)
where \(\bar Y\in\mathbb{R}^{T\times D}\) is obtained by centering and normalizing each feature dimension across tokens. This removes trivial feature-scale effects and focuses the diagnostic on token-space correlation structure. 
We use the theoretical MP edge \(\widehat{\lambda}_+ = (1+\sqrt{T/D})^2\) of the unit-variance isotropic null as a fixed reference and summarize each spectrum by
\begin{equation}
    \mathrm{OutMass}(C_Y)
    =
    \frac{\sum_{\lambda_k>\widehat{\lambda}_+}\lambda_k}
         {\sum_i \lambda_i},
    \qquad
    \frac{\lambda_1(C_Y)}{\widehat{\lambda}_+},
    \qquad
    r_{\mathrm{eff}}(C_Y)
    =
    \exp\!\left(-\sum_i p_i\log p_i\right),
\end{equation}
where \(p_i=\lambda_i/\sum_j\lambda_j\). These metrics measure out-of-bulk spectral mass, leading-mode dominance, and the spread of variance across token-space directions.
To connect spectral suppression to attention sinks, we construct a sink-column subspace from the attention matrix. 
For each layer and sample, 
let \(J_{\mathrm{sink}}\) denote the source positions whose head-averaged column mass \(\bar M_j=\frac{1}{T}\sum_i \bar A_{ij}\) is at least half its maximum (at most \(k=8\)).
For each \(j\in J_{\mathrm{sink}}\), define the normalized attention-column mode
\(
    \bar a_j
    = \bar A_{:,j} / \|\bar A_{:,j}\|_2
\)
and let \(P_{\mathrm{sink}}\) be the orthogonal projector onto \(\operatorname{span}\{\bar a_j:j\in J_{\mathrm{sink}}\}\). If \(q_1(C_Y)\) denotes the leading eigenvector of \(C_Y\), we measure sink-subspace alignment by
\begin{equation}
    \operatorname{Align}_{\mathrm{sink}}(C_Y)
    =
    \|P_{\mathrm{sink}}q_1(C_Y)\|_2^2 .
\end{equation}
The random baseline for a \(k\)-dimensional subspace is approximately \(k/T\) where $T=512$. 

\begin{figure}[t]
    \centering
    \subfigure[Outlier mass]{
    \includegraphics[width=0.23\linewidth]{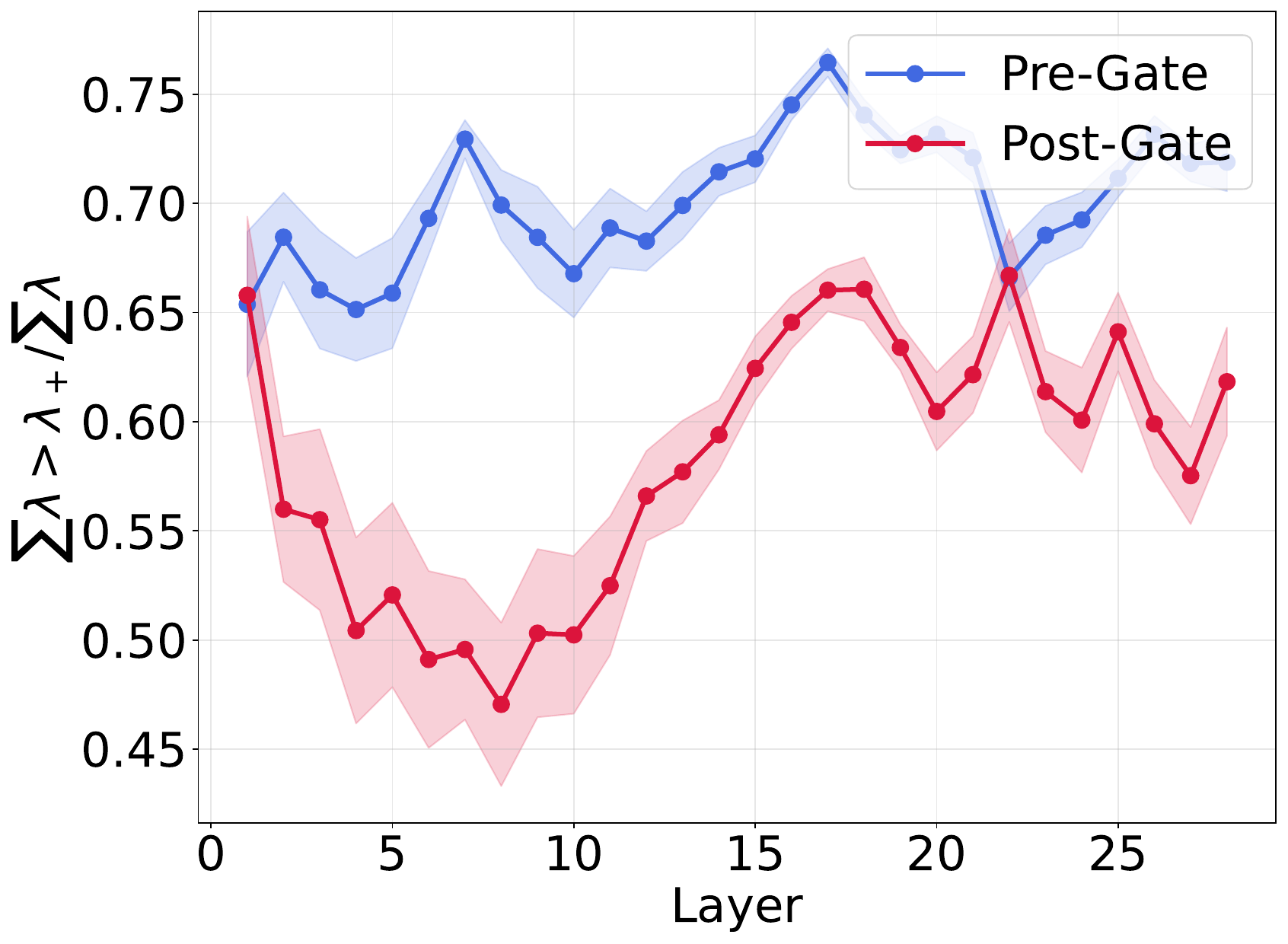}}
    \subfigure[\(\lambda_{\max}\) vs. MP edge]{
        \includegraphics[width=0.23\linewidth]{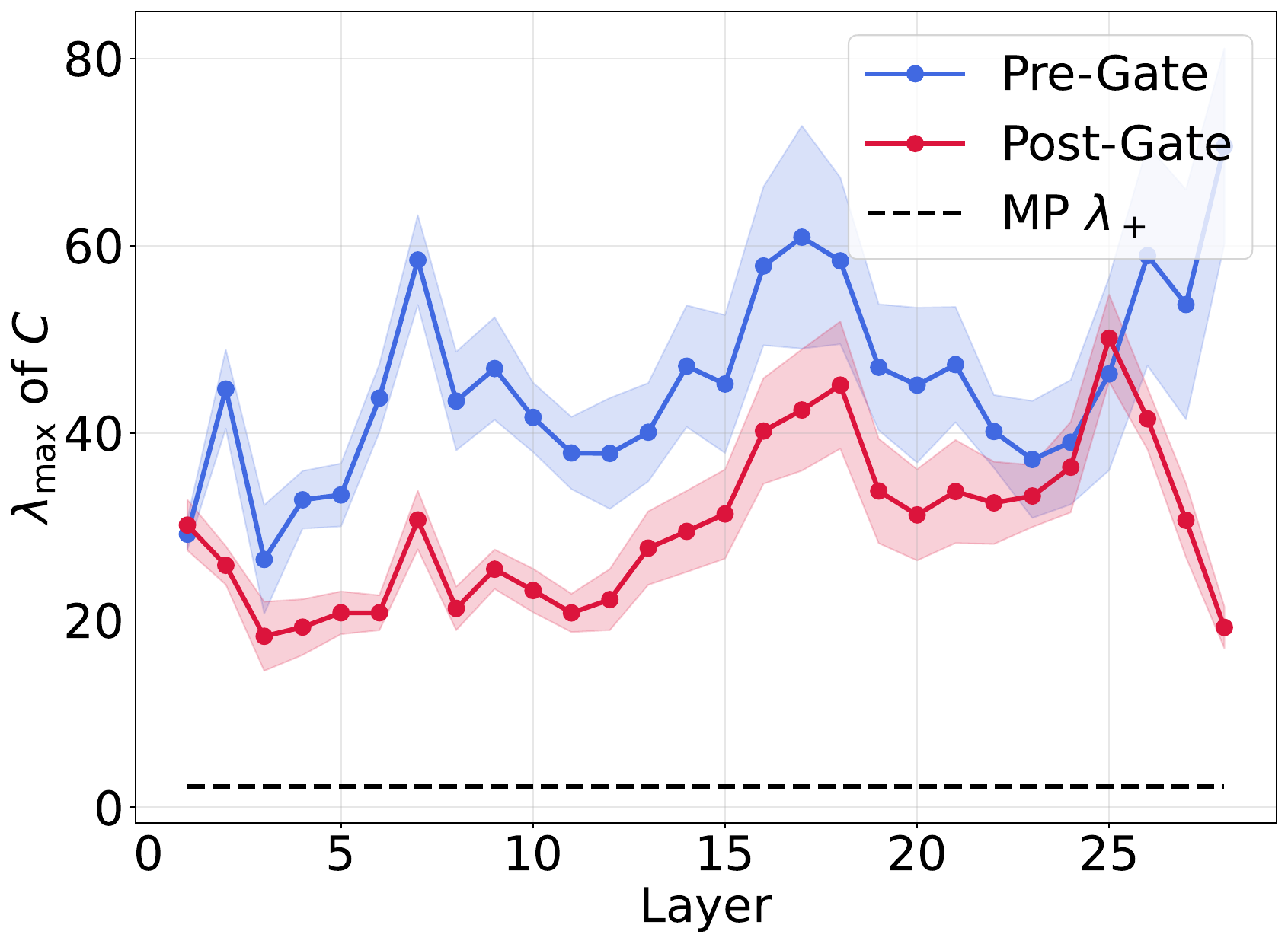}}
    \subfigure[Effective rank]{
        \includegraphics[width=0.23\linewidth]{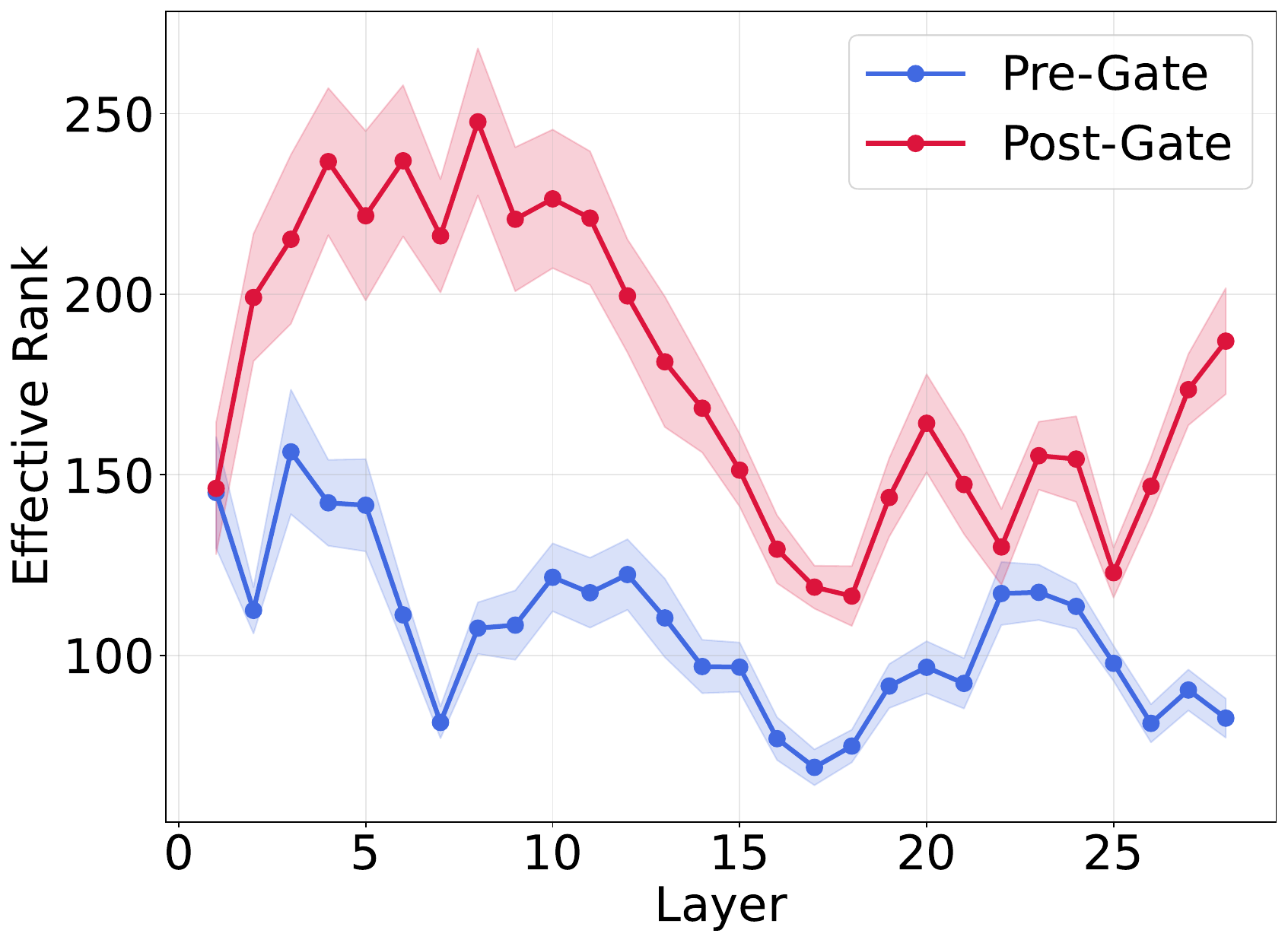}}
    \subfigure[Sink Alignment\label{fig:rmt_sink_alingment}]{
        \includegraphics[width=0.23\linewidth]{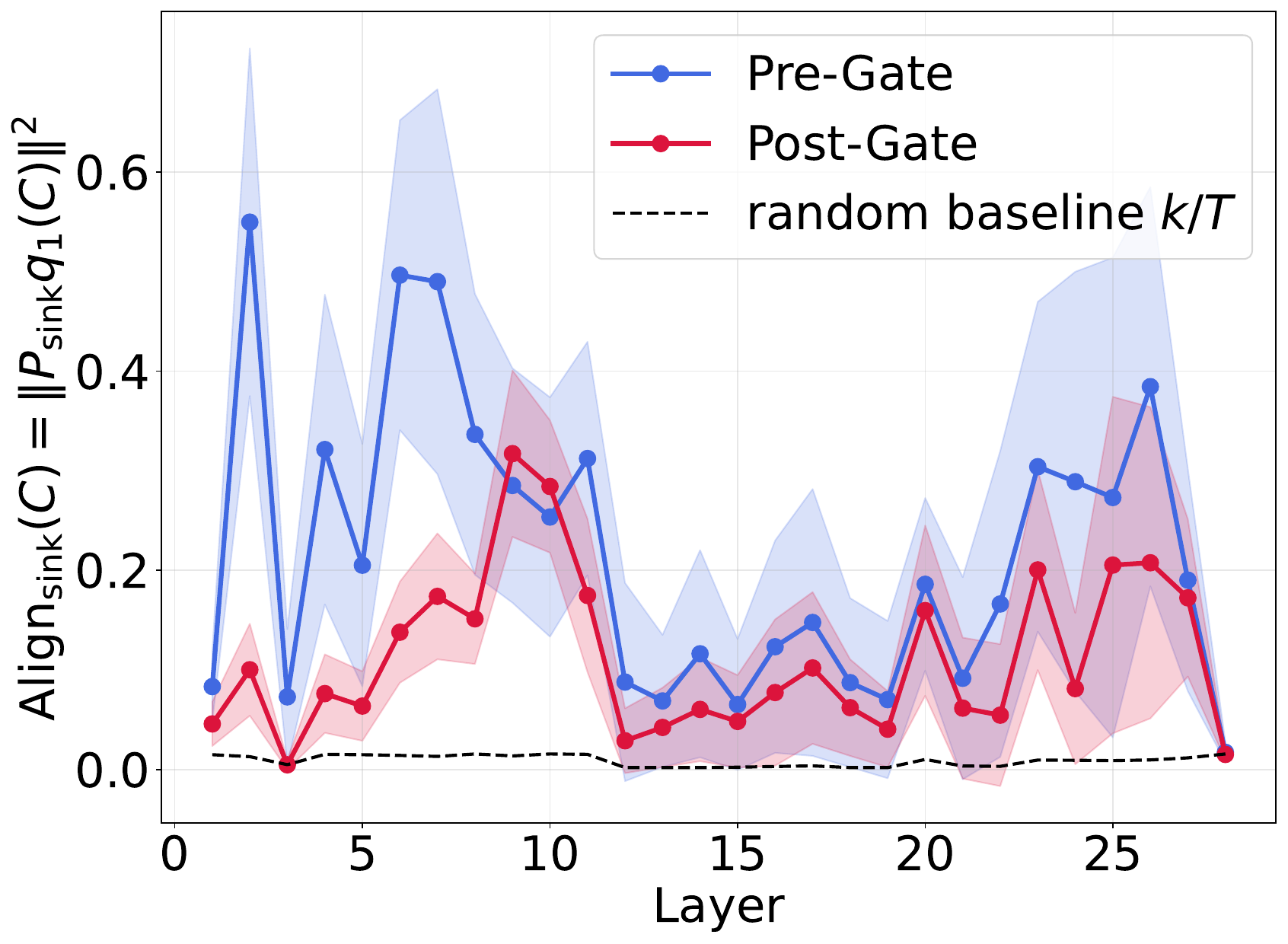}}
    \caption{
    \textbf{Post-SDPA gating suppresses token-space spectral structure.}
    (a) Outlier mass above the MP edge decreases.
    (b) The leading eigenvalue moves closer to the MP edge.
    (c) Effective rank increases.
    (d) Sink-subspace alignment \(\operatorname{Align}_{\mathrm{sink}}(C)\) decreases after gating; the dashed line shows the random-subspace baseline (layer-wise mean of $k/T$).
}
    \label{fig:rmt_suppression}
\end{figure}

\paragraph{Gating suppresses dominant token-space spectral structure.}
Figure~\ref{fig:rmt_suppression}(a-c) shows that post-SDPA gating consistently reduces spectral concentration. The gated output has lower outlier mass and a smaller leading eigenvalue relative to the MP edge, while its effective rank increases. Thus, after gating, the representation is less dominated by a small number of token-space directions. 
Since the covariance is computed after feature-wise standardization, which removes any per-feature scaling shared across tokens, this reduction is not a consequence of uniform feature-scale shrinkage; rather, the token-dependent variation of the learned gate changes the token-space correlation structure. Robustness checks using non-standardized covariance and scalar gating are reported in Appendix~\ref{app:rmt_nonstd}. 

\paragraph{Gating reduces sink-subspace alignment.}
Figure~\ref{fig:rmt_sink_alingment} further shows that the leading pre-gate spectral mode is often aligned with the sink-column subspace, substantially above the random \(k/T\) baseline. After gating, this alignment is strongly reduced across layers. This supports the sink-induced spike formulation in Definition~\ref{def:sink_induced_spike}: high-mass attention columns are not only attention-map artifacts, but can correspond to structured low-rank directions in the post-attention representation. Post-SDPA gating attenuates these sink-aligned directions together with the overall spectral outlier mass. Additional sink-subspace counterfactuals, which further probe the role of sink-induced modes, are reported in Appendix~\ref{app:rmt_controls}.

\paragraph{Gating weakens sink-aligned covariance structure.}

\begin{wraptable}{r}{0.43\linewidth}
\centering
\vspace{-1.5em}
\caption{\textbf{Sink-aligned mode retention.} Retention of highly and weakly sink-aligned directions in Qwen-1B.  The scalar control uses feature-averaged gates.}
\label{tab:main_gate_selectivity}
\small
\setlength{\tabcolsep}{4pt}

\begin{tabular}{lcc}
\toprule
Metric & High align. & Low align. \\
\midrule
Raw energy ($\alpha_k$) & 0.047 & 0.044 \\
Std. covariance ($R_k$) & 0.740 & 0.945 \\
Scalar gate ($R_k$) & 0.957 & 0.990 \\
\bottomrule
\end{tabular}
\vspace{-1em}
\end{wraptable}

We further examine how post-SDPA gating affects spectral directions with different degrees of sink alignment. As shown in Table~\ref{tab:main_gate_selectivity}, highly sink-aligned directions retain substantially less covariance energy than weakly aligned directions. This gap largely disappears when the elementwise post-SDPA gate is replaced by its feature-averaged scalar counterpart, while raw-energy retention remains similar across the two groups. These results indicate that post-SDPA gating preferentially weakens sink-aligned covariance structure beyond uniform magnitude reduction, consistent with the spectral suppression observed in Figure~\ref{fig:rmt_suppression}. Further details are provided in Appendix~\ref{app:gate_selectivity}.

\section{Backward Pass: Gradient Valve}
\label{sec:backward}

Having shown that gating attenuates sink-induced structure in the forward geometry, we next analyze its effect on backward dynamics. Building on the gradient-sink view~\cite{chen2026gradient}, we show that under fixed upstream signals, post-SDPA gating replaces sink-column gradient statistics with gated effective counterparts, acting as an implicit value-path gradient valve.

\subsection{Theory: A Gated Extension of the Gradient-Sink Bound}
\label{sec:backward_theory}

Consider a single causal attention head, $y_t = \sum_{j\le t} a_{tj}v_j$, where \(a_{tj}\) is the attention weight from query position \(t\) to source position \(j\), and \(v_j\in\mathbb{R}^d\) is the value vector. 
Since post-SDPA gate \(\Gamma_t\) is diagonal and contractive, define $\kappa_t := \|\Gamma_t\|_{\mathrm{op}}\le 1$. 
For a candidate sink token \(s\), define the sink column statistics without gate and corresponding gated effective statistics as 
\begin{equation}
    M_s := \sum_{t=s}^{T} a_{ts},
    \qquad
    S_s := \sum_{t=s}^{T} a_{ts}^2,
    \qquad
    M_s^{\mathrm{eff}}
    :=
    \sum_{t=s}^{T} a_{ts}\kappa_t,
    \qquad
    S_s^{\mathrm{eff}}
    :=
    \sum_{t=s}^{T} a_{ts}^2\kappa_t^2 .
\end{equation}
By construction,
\begin{equation}
    M_s^{\mathrm{eff}}\le M_s,
    \qquad
    S_s^{\mathrm{eff}}\le S_s .
\end{equation}

\begin{theorem}[Gated value-path gradient bound]
\label{thm:gated_value_path_gradient_bound}
Let \(\delta_t:=\nabla_{\widetilde y_t}\mathcal L\) be the upstream gradient at
the post-SDPA gated attention output. Then the gradient with respect to the source value
\(v_s\) is
\begin{equation}
    \nabla_{v_s}\mathcal L
    =
    \sum_{t=s}^{T} a_{ts}\Gamma_t^\top\delta_t .
\end{equation}
Consequently,
\begin{equation}
    \|\nabla_{v_s}\mathcal L\|
    \le
    \sum_{t=s}^{T} a_{ts}\kappa_t\|\delta_t\|.
\end{equation}
In particular, if \(\|\delta_t\|\le B\) for all \(t\), then
\begin{equation}
    \|\nabla_{v_s}\mathcal L\|
    \le
    B M_s^{\mathrm{eff}}
    \le
    B M_s .
\end{equation}
\end{theorem}

The coherent bound above controls the worst-case aligned contribution of future-query gradients. A complementary second-moment statement applies when the upstream gradients behave incoherently. 

\begin{corollary}[Second-moment attenuation]
\label{cor:gated_second_moment}
Conditioned on the attention weights and gates, assume that the upstream
gradients satisfy
\begin{equation}
    \mathbb E[\delta_t\delta_r^\top]=0
    \quad (t\ne r),
    \qquad
    \mathbb E\|\delta_t\|^2\le B^2.
\end{equation}
Then
\begin{equation}
    \mathbb E\|\nabla_{v_s}\mathcal L\|^2
    \le
    B^2 S_s^{\mathrm{eff}}
    \le
    B^2 S_s.
\end{equation}
\end{corollary}

Theorem~\ref{thm:gated_value_path_gradient_bound} and
Corollary~\ref{cor:gated_second_moment} show that post-SDPA gating attenuates value-path gradient aggregation under both coherent and incoherent regimes. Importantly, this contraction does not require changing the attention matrix itself. Even if the same attention weights \(a_{ts}\) are fixed, each backpropagated contribution from query \(t\) to source value \(v_s\) is filtered by the gate. Thus, the backward strength of a sink column is governed by \(M_s^{\mathrm{eff}}\) and \(S_s^{\mathrm{eff}}\), not only by \(M_s\) and \(S_s\). Proofs for Theorem~\ref{thm:gated_value_path_gradient_bound} and Corollary~\ref{cor:gated_second_moment} can be found in Appendix~\ref{app:gradient_proofs}.

\subsection{Empirical: Gradient Sink under Post-SDPA Gating}
\label{sec:backward_empirical}

\paragraph{Practical proxies.}
Theorem~\ref{thm:gated_value_path_gradient_bound} is a local Jacobian statement: conditioned on fixed attention weights and upstream gradients, post-SDPA gating contracts value-path contributions routed through a sink column. We evaluate this prediction using two complementary diagnostics. First, we compare Qwen-1B models with and without post-SDPA gating under their actual forward-backward passes. This measures the outcome of training with post-SDPA gates. Second, within post-SDPA gated model, we isolate the local value-path effect of the gate by keeping the attention weights and upstream gradients fixed, and recomputing the first-token value-gradient contribution with and without the gate. This fixed-upstream-gradient counterfactual directly tests the valve effect predicted by Theorem~\ref{thm:gated_value_path_gradient_bound}.

We focus on the first-token sink, which is the primary sink position in this setting. We measure value-gradient amplification at source position \(s=1\) by
\begin{equation}
    \rho_{1}
    =
    \frac{\|\nabla_{v_{1}}\mathcal L\|}
    {(T-1)^{-1}\sum_{j\ne 1}\|\nabla_{v_j}\mathcal L\|}.
    \label{eq:rho1}
\end{equation}
To isolate the local value-path effect of the active gate, we cache \(\delta_t=\nabla_{\widetilde y_t}\mathcal L\) from the actual gated backward pass and compute
\begin{equation}
    g^{\mathrm{actual}}_{v_1}
    =
    \sum_{t=1}^{T}a_{t1}\Gamma_t^\top\delta_t,
    \qquad
    g^{\mathrm{cf}}_{v_1}
    =
    \sum_{t=1}^{T}a_{t1}\delta_t,
    \qquad
    A^{\mathrm{local}}_1
    =
    \frac{\|g^{\mathrm{actual}}_{v_1}\|_2}{\|g^{\mathrm{cf}}_{v_1}\|_2}.
\end{equation}
As a simple gate-mass proxy, we also compute the RMS effective-mass ratio
\begin{equation}
    \widetilde M_{1}^{\mathrm{rms}}
    =
    \sum_{t=1}^{T} a_{t1} r_t,
    \qquad
    R_{1}^{\mathrm{rms}}
    =
    \frac{\widetilde M_{1}^{\mathrm{rms}}}{M_{1}},
    \qquad
    M_{1}
    =
    \sum_{t=1}^{T}a_{t1},
    \qquad
    r_t
    =
    \left(\frac{1}{d}\sum_{c=1}^{d}G_{t,c}^2\right)^{1/2}.
    \label{eq:rms}
\end{equation}
This RMS quantity is an isotropic-gradient proxy for average feature-wise attenuation.

\begin{figure}[t]
    \centering
    \subfigure[Actual gradient amplification\label{fig:backward_sink_amp}]{
        \includegraphics[width=0.32\linewidth]{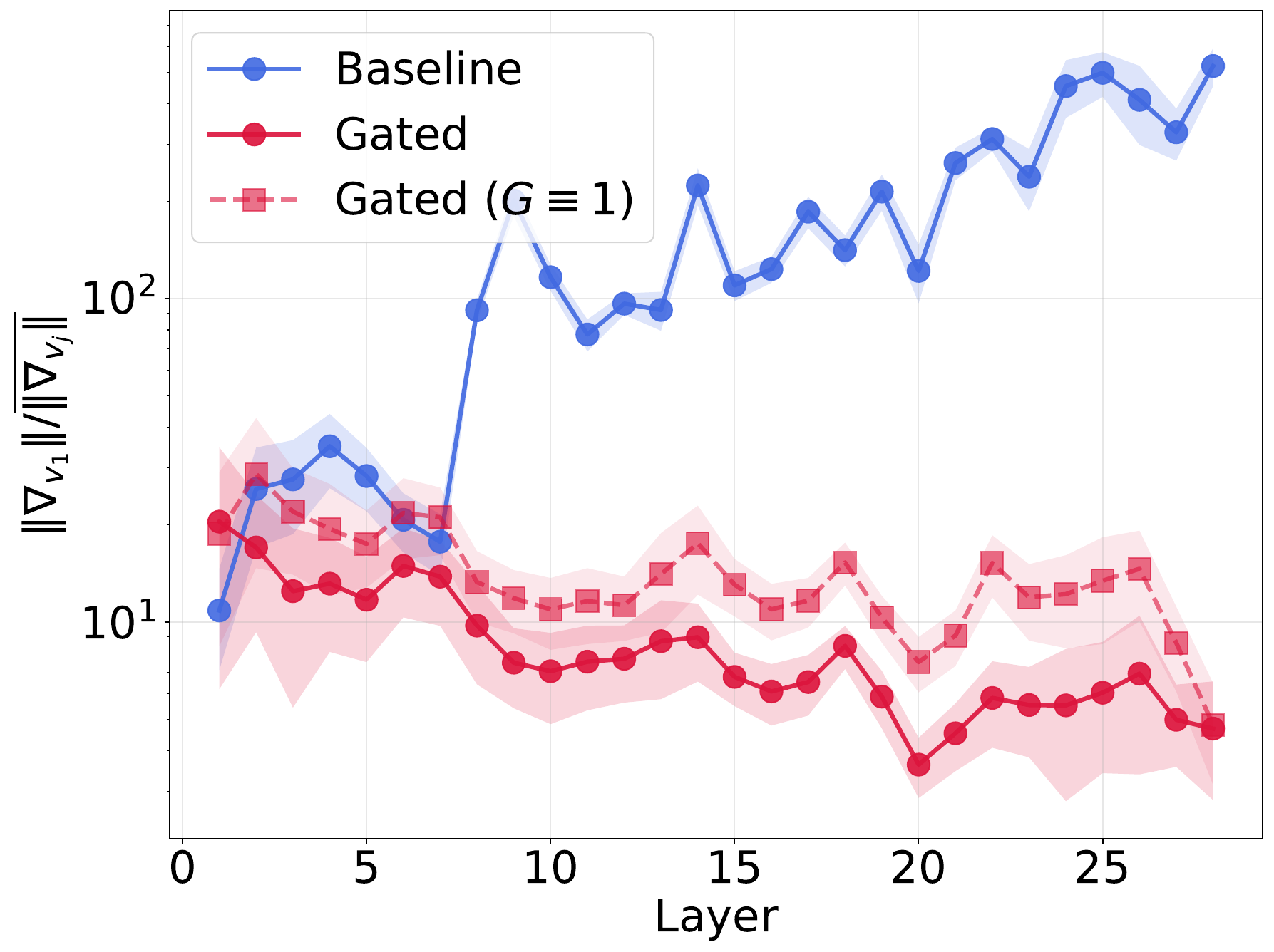}}
    \hfill
    \subfigure[Local gradient attenuation]{
        \includegraphics[width=0.32\linewidth]{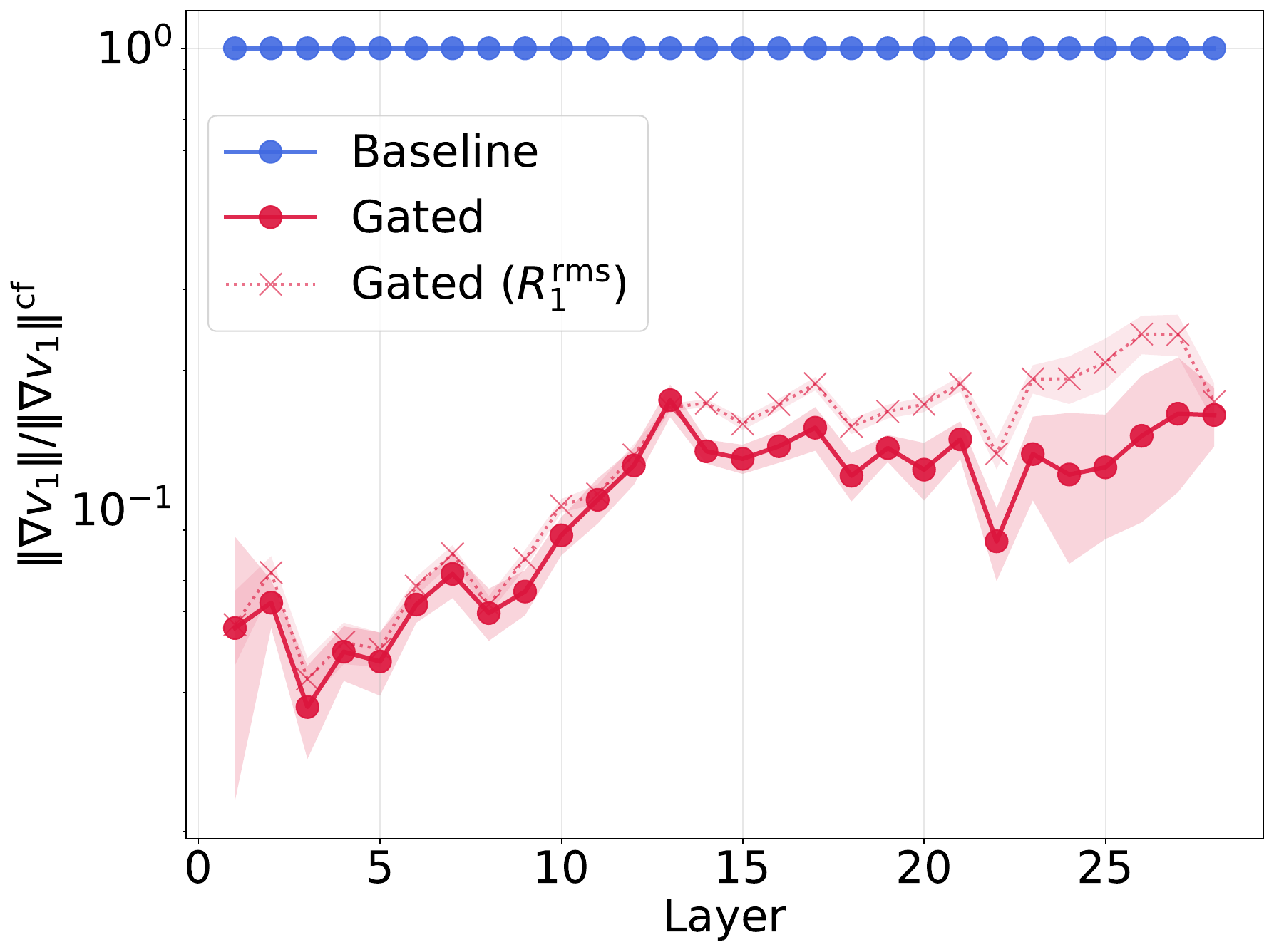}}
    \hfill
    \subfigure[Attenuation vs. effective mass]{
        \includegraphics[width=0.32\linewidth]{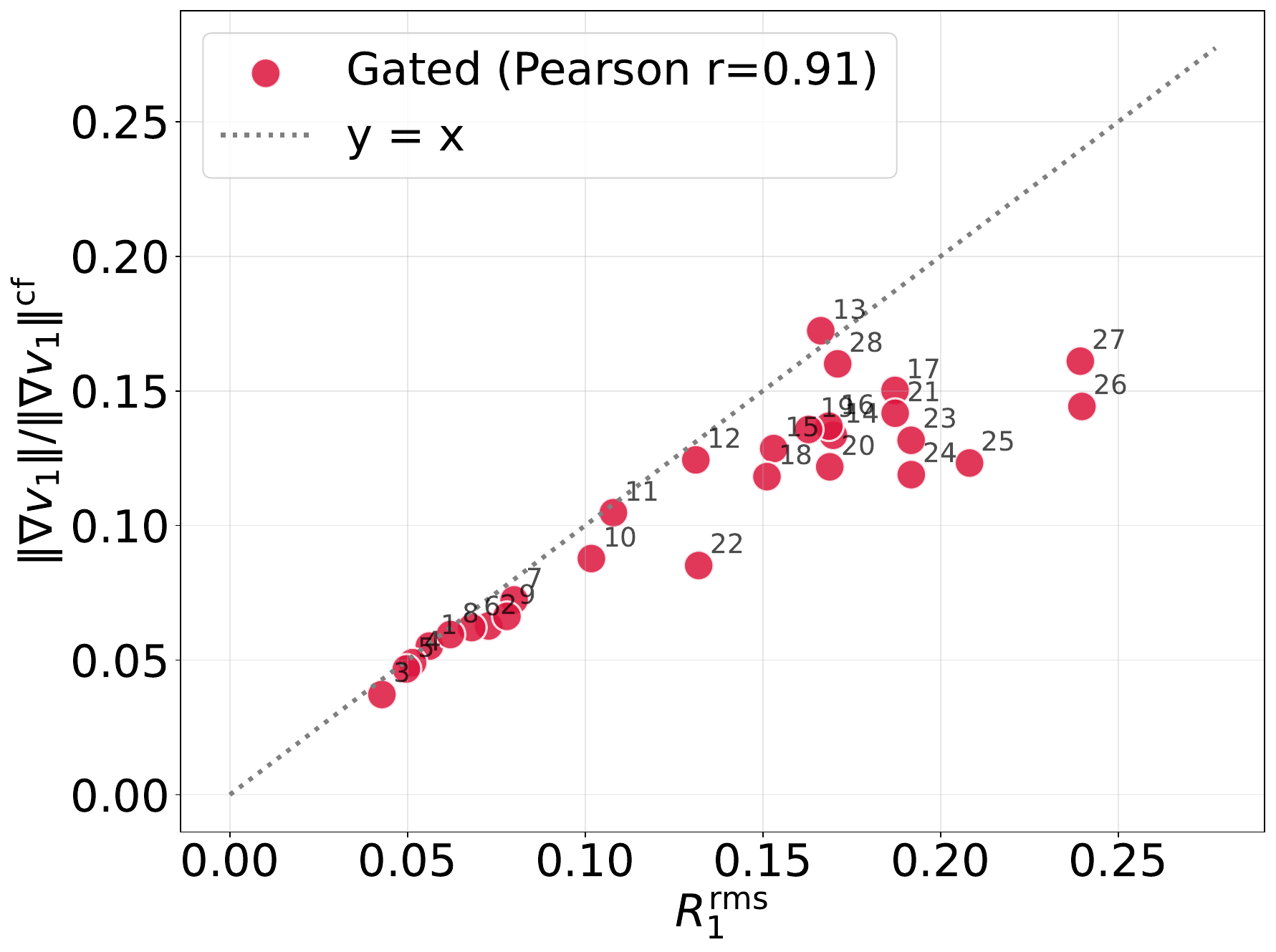}}
    \caption{
    \textbf{Post-SDPA gating attenuates first-token value-path gradients.}
    (a) Post-SDPA gate suppresses first-token value-gradient amplification.
    (b) Local attenuation tracks the RMS effective-mass proxy \(R_{1}^{\mathrm{rms}}\).
    (c) Their layer-wise correlation supports the fixed-upstream-gradient valve mechanism.
    }
    \label{fig:backward_sink}
\end{figure}

\paragraph{Results.}
Figure~\ref{fig:backward_sink_amp} reports the trained-model outcome. Baseline (without post-SDPA gating) develops large first-token value-gradient amplification in mid-to-late layers, whereas the gated model remains close to the non-sink scale. The counterfactual curve in the model with post-SDPA gating (Gated (G $\equiv$ 1)) shows that removing the gate increases the first-token gradient even without changing the trained weights, indicating that the active gate provides an additional attenuation mechanism beyond the learned representation itself. Figures~\ref{fig:backward_sink}(b,c) isolate this local valve effect. Holding the attention weights and upstream gradients fixed, we recompute the first-token value-gradient contribution with and without the gate. The resulting attenuation ratio \(A^{\mathrm{local}}_1\) is far below one across layers and closely tracks the RMS effective-mass proxy \(R_{1}^{\mathrm{rms}}\). 
The layer-wise scatter strongly correlates these quantities, supporting the Jacobian prediction of Theorem~\ref{thm:gated_value_path_gradient_bound} that post-SDPA gating attenuates value gradients according to the effective gated mass through the first-token sink column.

\subsection{Implication for the Attention-Sink--Gradient-Sink--Massive-Activation Pathway}
\label{sec:backward_implication}

\begin{wraptable}{r}{0.375\textwidth}
\centering
\vspace{-1.75em}
\caption{\textbf{AS--GS--MA suppression.} Final-checkpoint diagnostics for baseline and post-SDPA gated models on Pythia-410M. Detailed measurements are provided in Appendix~\ref{app:massive_activations}.}
\vspace{0.5em}
\label{tab:main_as_gs_ma}
\small
\begin{tabular}{ccc}
\toprule
Metric & Baseline & Gated \\
\midrule
AS ($m_1$) & 0.084 & 0.013 \\
GS ($\rho_1$) & 25.40 & 5.67 \\
MA ($M_\infty$) & 320.0 & 67.4 \\
\bottomrule
\end{tabular}
\vspace{-1em}
\end{wraptable}
The forward attention sink can create a backward gradient sink because many future query positions route their value-path gradients through the same source token. Previous work~\citep{chen2026gradient} suggests that this localized gradient pressure mediates the emergence of massive activations during training.
Our analysis shows that post-SDPA gating weakens precisely this value-path link. The gate does not need to remove the attention sink column; it attenuates each future-query contribution to the sink-token value gradient, reducing the effective sink-column mass even with fixed attention weights and upstream gradients. Thus, standard gated attention implements an implicit gradient valve using the same learned gate that suppresses forward spectral structure. This provides the backward counterpart to the spectral suppression mechanism in Section~\ref{sec:forward_spectral}, connecting the attenuation of sink-induced representation modes with reduced sink-directed gradient flow.

The quantitative diagnostics in Table~\ref{tab:main_as_gs_ma} provide complementary evidence: gating reduces attention concentration, value-gradient amplification, and massive activations in Pythia-410M. These signatures are most pronounced in the middle-to-late layers of the baseline model, while gating substantially suppresses their emergence during training. Qwen-1B further exhibits strong massive-activation suppression, supporting the consistency of this effect across models. Together, these findings are consistent with attenuation along the AS--GS--MA pathway. Further layer-wise and training-time analyses and details of metrics are provided in Appendix~\ref{app:massive_activations}.

\section{Training-Time Analyses of Sink Signatures} \label{sec:training_analysis}

\subsection{Training-Time Sink Emergence}
In this section, we examine whether the sink signatures analyzed in Sections~\ref{sec:forward_spectral} and~\ref{sec:backward} also emerge during training. Using checkpoints of baseline and post-SDPA gated models that we train with the Pythia-410M architecture~\cite{biderman2023pythia}, we track first-token attention concentration and representation dominance across layers. Details and additional results are in Appendices~\ref{app:analysis_setting} and~\ref{app:training_diagnostics}.

\begin{figure}[t]
    \centering
    \def\trainh{0.145\textheight}
    \subfigure[Baseline attention mass]{\includegraphics[height=\trainh,keepaspectratio]{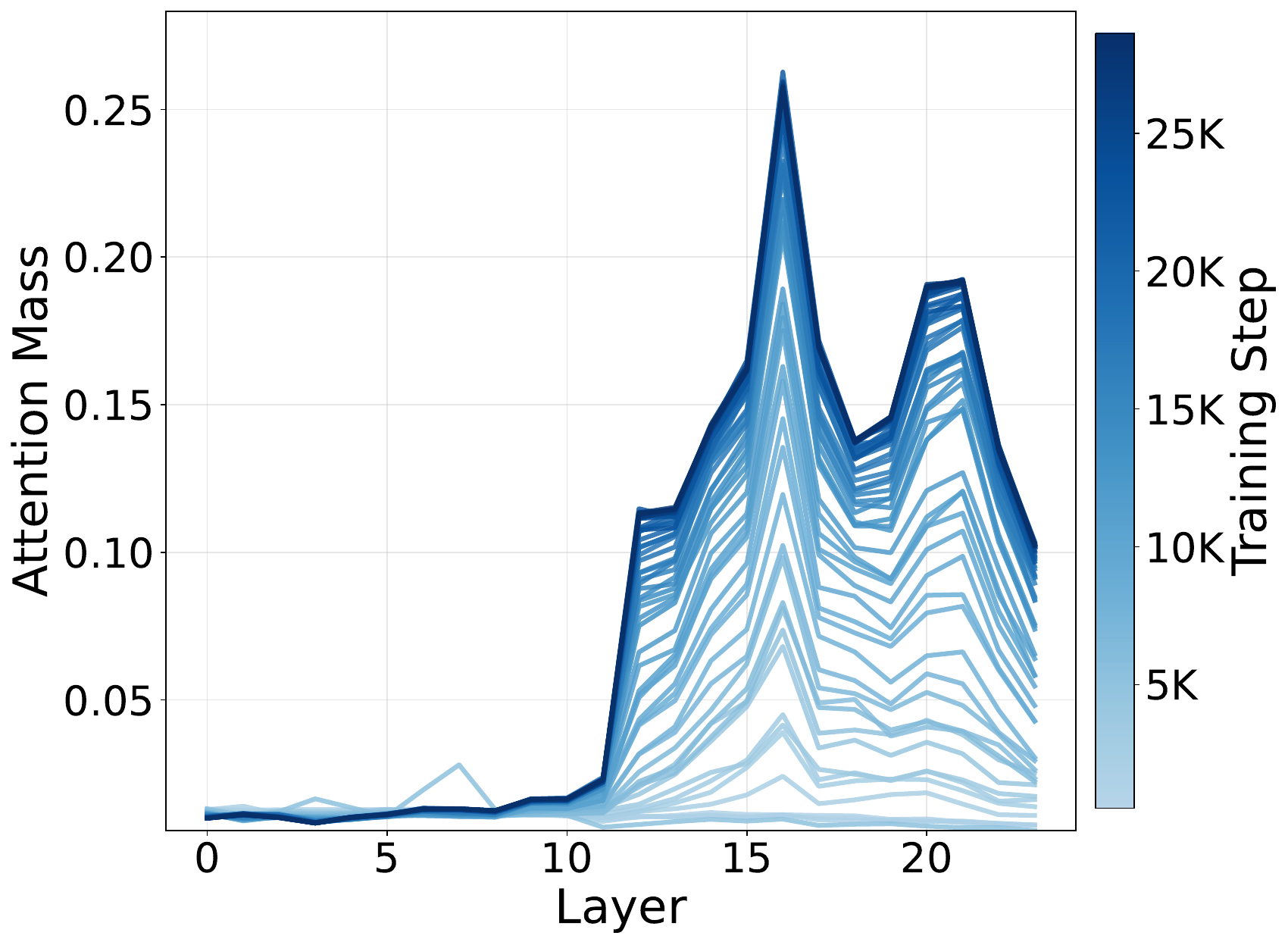}\label{fig:m_bos_baseline}}
    \subfigure[Gated attention mass]{\includegraphics[height=\trainh,keepaspectratio]{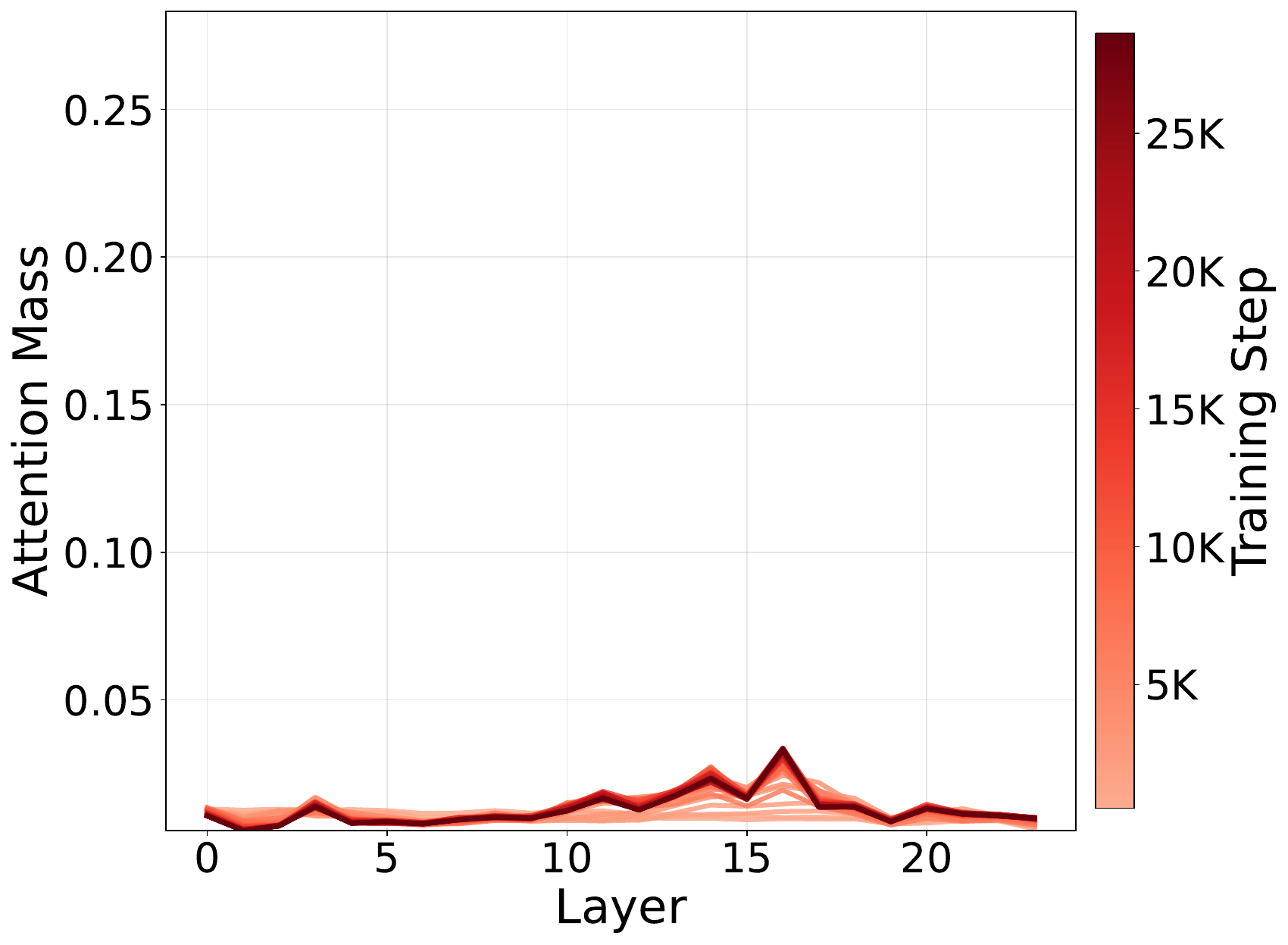}\label{fig:m_bos_gated}}
    \subfigure[Representation dominance]{\includegraphics[height=\trainh,keepaspectratio]{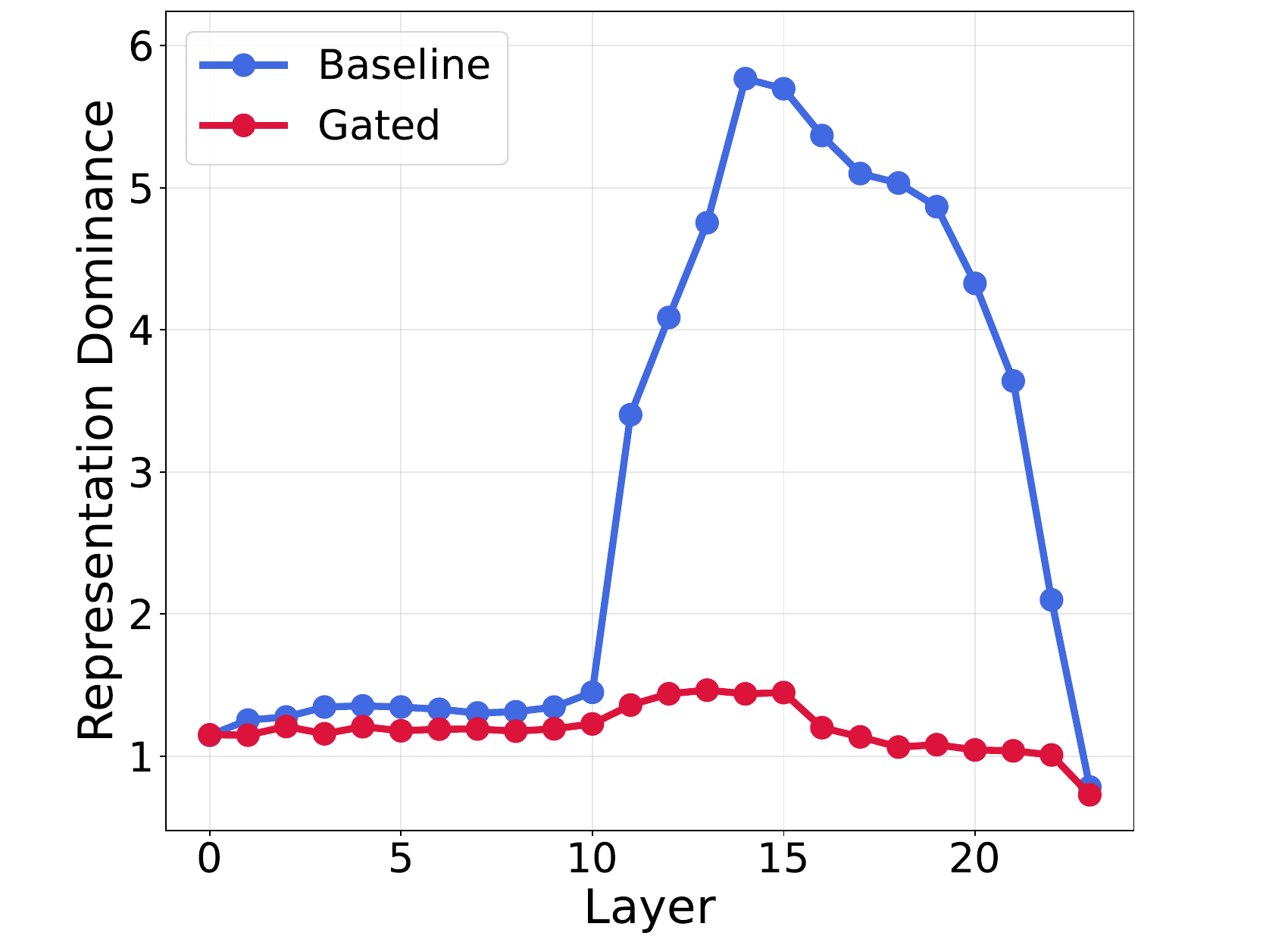}\label{fig:bos_dominance}}
    \\
    \subfigure[Baseline gradient ratio]{\includegraphics[height=\trainh,keepaspectratio]{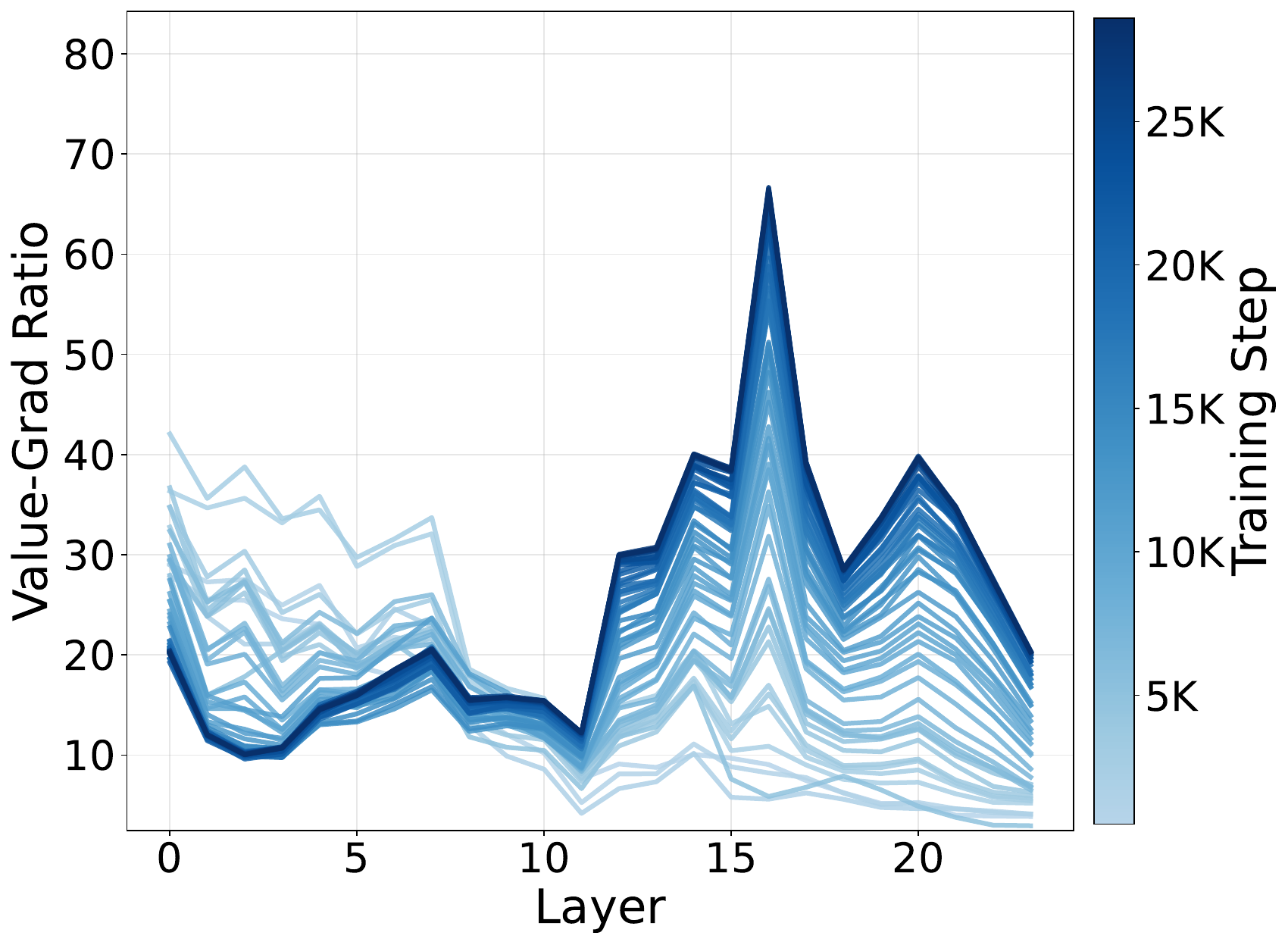}\label{fig:rho_baseline}}
    \subfigure[Gated gradient ratio]{\includegraphics[height=\trainh,keepaspectratio]{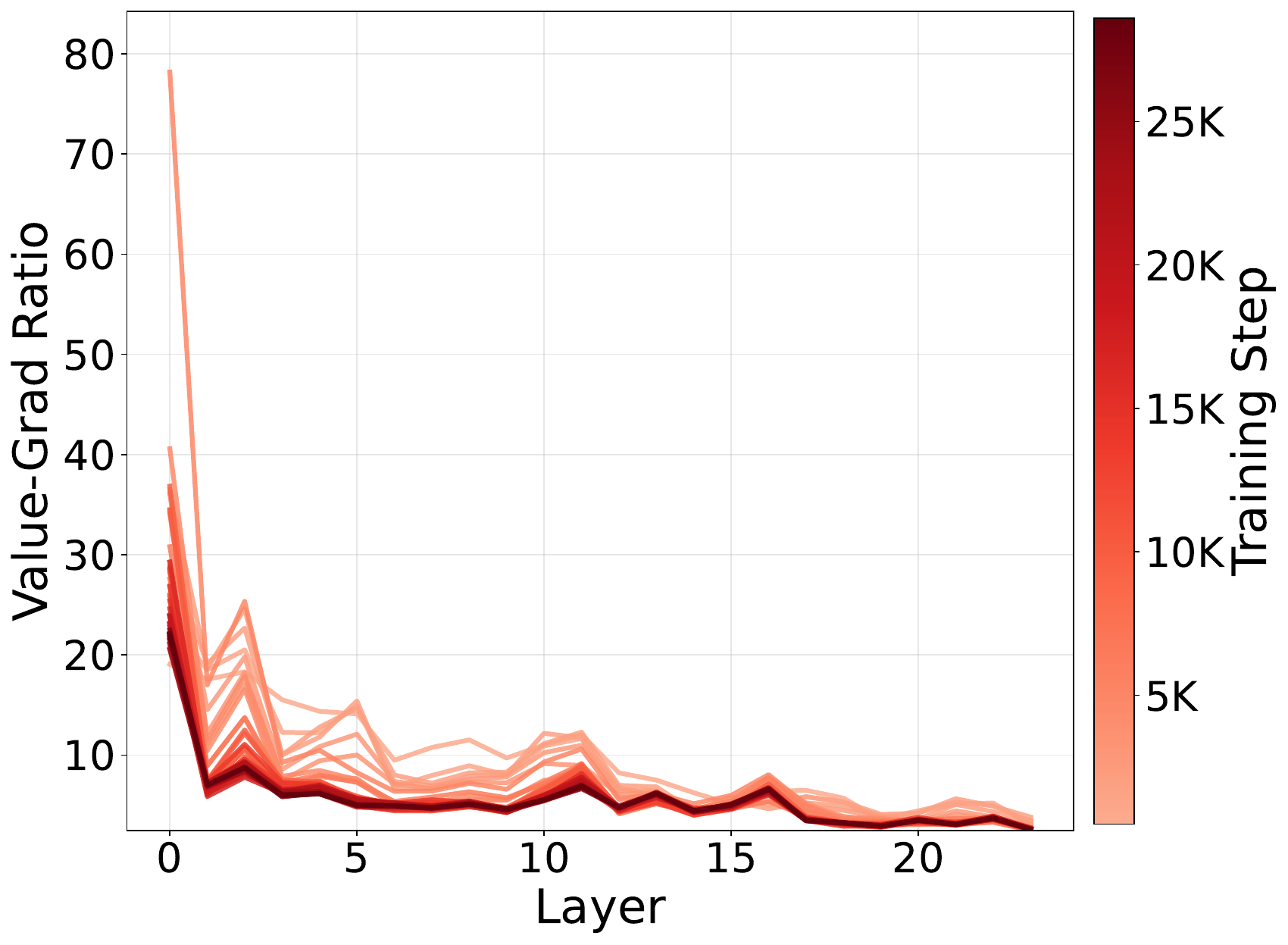}\label{fig:rho_gated}}
    \subfigure[Gated effective mass]{\includegraphics[height=\trainh,keepaspectratio]{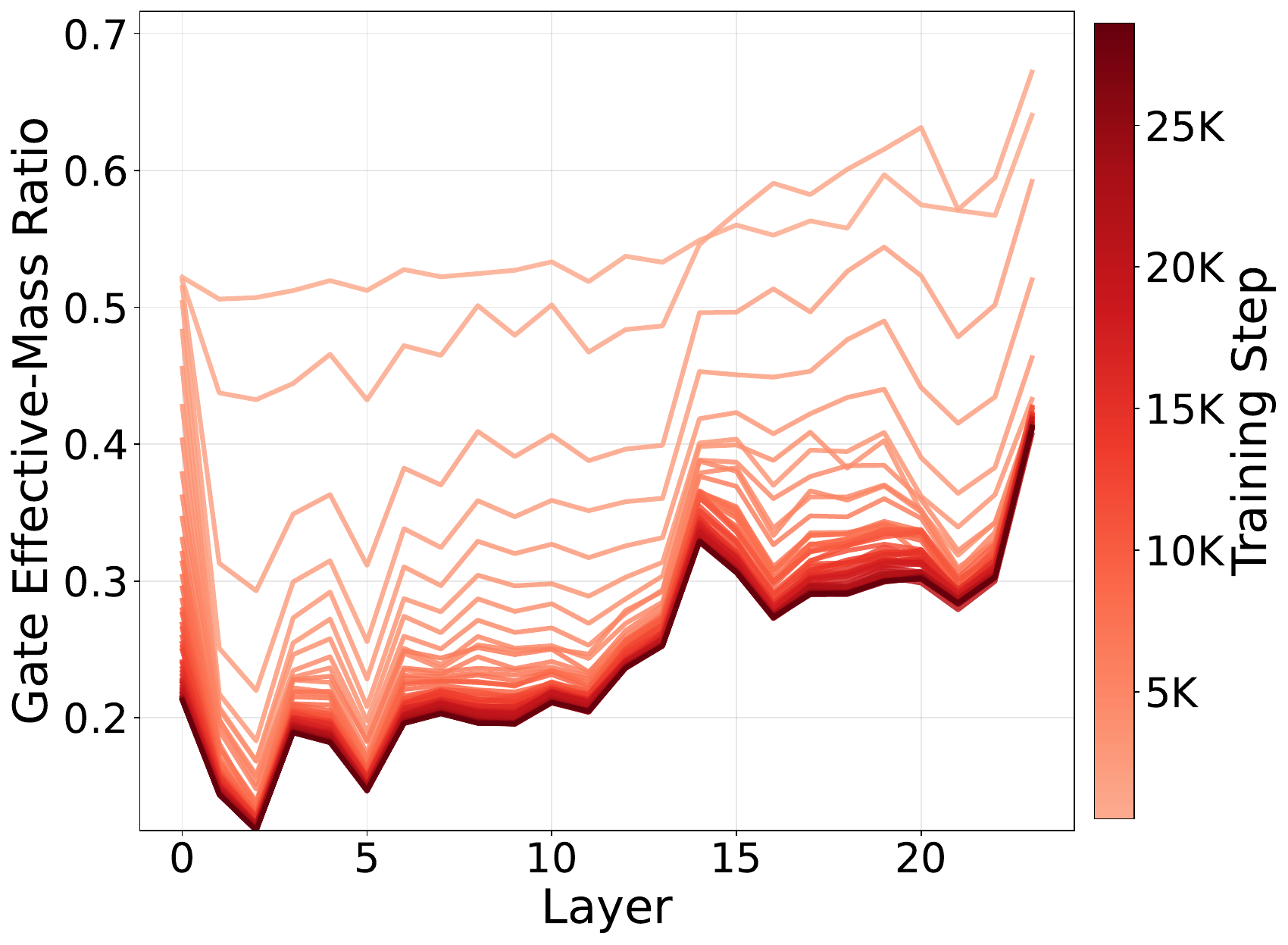}\label{fig:reff_gated}}
    \caption{\textbf{Training-time sink and gradient signatures.} Curves in (a, b, d--f) correspond to training checkpoints. (a, b) First-token attention mass: Baseline develops a middle-to-late-layer sink, whereas the post-SDPA gated model suppresses it. (c) First-token representation dominance at the final checkpoint, which is large in the baseline but attenuated by gating. (d, e) First-token value-gradient amplification \(\rho_1\): Baseline develops strong amplification in middle-to-late layers, whereas gating suppresses it. (f) Gated effective-mass ratio \(R_1^{\mathrm{rms}}\) remains well below one across training.}
    \label{fig:training}
\end{figure}

\paragraph{Diagnostics.}
For each checkpoint and layer, we track the canonical first-token sink via the causal column-mass diagnostic from Section~\ref{sec:prelim_attention_sink}. Since this corresponds to \(s=1\), we report
\begin{equation}
    m^{(\ell)}_{1}
    =
    \frac{1}{T}\sum_{t=1}^{T} A^{(\ell)}_{t,1},
    \qquad
    D^{(\ell)}_{1}
    =
    \frac{\|h^{(\ell)}_1\|_2}
    {\frac{1}{T-1}\sum_{j\ne 1}\|h^{(\ell)}_j\|_2}.
\end{equation}
The first quantity measures first-token attention concentration, and the second measures whether the corresponding hidden state becomes unusually large relative to other positions.

\paragraph{Results.}
Figure~\ref{fig:training}(a,b) tracks first-token attention mass across training checkpoints. In the baseline, first-token mass increases sharply in middle-to-late layers, indicating the emergence of a strong first-token attention sink. The post-SDPA gated model suppresses this growth and maintains substantially lower first-token mass throughout training. Figure~\ref{fig:training}(c) shows the corresponding representation-level signature at the final checkpoint: the baseline model develops pronounced first-token hidden-state dominance in the same layer range, whereas the gated model remains closer to the non-first-token scale. Thus, both attention- and representation-level sink signatures emerge during training and are substantially attenuated by post-SDPA gating. The additional analyses in Appendix~\ref{app:rmt_structural_bulk} show reduced heavy-tail dominance while the bulk spacing ratio remains stable. 
Layer-wise linear probing in Appendix~\ref{app:probing} further shows comparable next-token cross-entropy loss for the baseline and gated models once the probes are trained to convergence, with a similar result after excluding the initial source position.

\subsection{Training-Time Gradient Amplification}

We next examine whether the backward-pass signature analyzed in Section~\ref{sec:backward} also emerges during training. Because many future queries route their value-path gradients through the same high-mass source token, a forward attention sink can induce a backward gradient sink (Section~\ref{sec:backward_implication}). Using the same checkpoints, we track the first-token value-gradient amplification \(\rho_1\) in Eq.~\eqref{eq:rho1} and, for the gated model, the effective-mass ratio \(R_1^{\mathrm{rms}}\) in Eq.~\eqref{eq:rms}. As shown in Figure~\ref{fig:training}(d,e), the baseline develops large first-token value-gradient amplification in middle-to-late layers, the same range in which the first-token attention sink emerges, whereas the gated model keeps this amplification substantially lower in these layers. The effective-mass ratio remains well below one across much of training (Figure~\ref{fig:training}(f)), indicating that the active gate attenuates value-path contributions routed through the sink column, consistent with the local gradient valve in Section~\ref{sec:backward_theory}. Since \(R_1^{\mathrm{rms}}\) depends only on the attention weights and gate values, this attenuation applies even where the gated model still assigns non-negligible attention to the first token. Together with the results above, these findings show that attention concentration, value-gradient amplification, and representation dominance co-emerge in the baseline and are jointly attenuated by post-SDPA gating (additional diagnostics in Appendix~\ref{app:training_diagnostics}).

\section{Related Work}
\label{sec:related_work}

\paragraph{Attention sink phenomenon.}
Attention sinks describe the tendency of autoregressive language models to route disproportionate attention mass to a small set of tokens, often early positions, regardless of semantic relevance. Prior work shows that this phenomenon is widespread across models and inputs, emerges during pretraining, and is affected by optimization, data distribution, loss design, and attention normalization \citep{gu2025attention,peng2026attention}. Recent studies further clarify its relation to massive activations \citep{sun2024massive, yu2024super}: sinks and activation spikes often co-occur, but they are geometrically and functionally distinct \citep{sun2026spike}. Another study~\cite{chen2026gradient} complements this view from the backward pass, showing that attention sinks can induce value-path gradient concentration under causal masking.

\paragraph{Gated and normalization-based sink mitigation.}
Recent work has explored architectural interventions for reducing attention sinks, massive activations, or their co-occurrence. Post-SDPA sigmoid gating improves performance and stability while strongly reducing first-token attention concentration, attributing the effect to nonlinearity and query-dependent sparsity~\cite{qiu2025gated}. Alternative approaches modify attention normalization, as in Softpick's rectified softmax and Elastic-Softmax's learnable filtering of softmax weights~\citep{zuhri-etal-2026-softpick,fu2026attention}, or directly gate value states, as in Value-State Gated Attention (VGA)~\citep{bu2025value}. Other methods modify normalization or activation dynamics, including DyT~\citep{zhu2025transformers}, or introduce normalization and attention-bias variants that shift, suppress, or decouple sinks and massive activations~\citep{sun2026spike,chen2026gradient,ramapuram2024theory}. 
Despite these advances, the mechanisms underlying sink suppression by post-SDPA gating remain unclear. We investigate them through spectral and gradient analyses.

\paragraph{Spectral analysis on neural representations.}
Spectral methods and random matrix theory have been widely used to analyze neural network weights, activations, and learned representations, separating noise-like bulk spectra from structured low-rank components \citep{martin2021implicit,louart2018random,marvcenko1967distribution,baik2005phase}. Related studies of representation collapse also emphasize the role of low-rank structure in deep features \citep{papyan2020prevalence,queipo-de-llano2026attention}. Complementary work uses spectral or graph-signal viewpoints to design and interpret sequence models and representation dynamics \citep{shin2026graph,choi2023graph,wi2025learning}. 
In contrast, we use spectral outliers to characterize sink-aligned structure and its suppression in token-space attention representations.

\section{Conclusion}
We presented a spectral and gradient-based analysis of how post-SDPA gating mitigates attention sinks. High-mass attention columns can induce low-rank token-space modes, which learned gates attenuate beyond uniform shrinkage. In the backward pass, the gate locally reduces value-path gradient aggregation through sink columns, a gradient-valve counterpart to the forward spectral contraction. Empirically, gating reduces spectral outliers and sink-subspace alignment, attenuates first-token value-gradient amplification, and suppresses the training-time emergence of first-token attention concentration and representation dominance. Together, these results provide a unified forward--backward account of sink suppression by post-SDPA gating.
\subsubsection*{Acknowledgments}

This research was partly supported by the MSIT (Ministry of Science and ICT), Korea, under the Development and Demonstration of Agentic AI Technology for Autonomous Physics Simulation Based on MCP Standard with Multi-Agent Collaboration (RS-2026-25587486), Global Research Support Program in the Digital Field program (RS-2024-00436680), AI Research Hub Project (RS-2024-00457882), and LG AI STAR Talent Development Program for Leading Large-Scale Generative AI Models in the Physical AI Domain (RS-2025-25442149), supervised by the IITP (Institute for Information \& Communications Technology Planning \& Evaluation). This research was partly supported by the National Research Foundation of Korea(NRF) grant funded by the Korea government(MSIT) (No. RS-2026-25607520, Center for Specialized Orchestration and Training of AI for Frontier Mathematics (SOTA-MATH)). Research for this paper was also conducted under the KICT Strategic Research Program(SRP)(project no. 20260437-001, Development of a Construction Site-Specific Vertical AI Model and Sustainable Platform) funded by the Ministry of Science and ICT. This project is supported by Microsoft Research Asia.

\bibliographystyle{unsrt}
\bibliography{reference}
\clearpage
\appendix

\section{Proofs for Spectral Sink Suppression}
\label{app:spectral_proofs}

\subsection{Proof of Theorem~\ref{thm:sink_mode_contraction}}

\begin{proof}
The \(i\)-th row of the sink-induced signal component is
\begin{equation*}
    (Y_{\mathrm{sig}})_i
    =
    \sqrt das_i u^\top .
\end{equation*}
After applying the post-SDPA gate, it becomes
\begin{equation*}
    (\widetilde Y_{\mathrm{sig}})_i
    =
    \sqrt das_i(\Gamma_i u)^\top .
\end{equation*}
Therefore,
\begin{equation*}
    \frac{1}{d}\|\widetilde Y_{\mathrm{sig}}\|_F^2
    =
    a^2
    \sum_{i=1}^{n}
    s_i^2
    \|\Gamma_i u\|_2^2 .
\end{equation*}
By the definitions \(g_i=\|\Gamma_i u\|_2\), \(D=\mathrm{diag}(g_1,\ldots,g_n)\), and
\[
    \alpha_s=\|Ds\|_2^2
    =
    \sum_{i=1}^{n}g_i^2s_i^2,
\]
we obtain
\begin{equation*}
    \frac{1}{d}\|\widetilde Y_{\mathrm{sig}}\|_F^2
    =
    a^2\|Ds\|_2^2
    =
    a^2\alpha_s .
\end{equation*}
Since \(0\preceq \Gamma_i\preceq I\) and \(\|u\|_2=1\), we have
\(0\le g_i\le 1\). Hence
\begin{equation*}
    \alpha_s
    =
    \sum_{i=1}^{n}g_i^2s_i^2
    \le
    \sum_{i=1}^{n}s_i^2
    =
    1,
\end{equation*}
which gives
\[
    \frac{1}{d}\|\widetilde Y_{\mathrm{sig}}\|_F^2
    =
    a^2\alpha_s
    \le
    a^2 .
\]

It remains to prove the covariance-level spectral bound. Define
\begin{equation*}
    \widetilde C_{\mathrm{sig}}
    :=
    \frac{1}{d}
    \widetilde Y_{\mathrm{sig}}\widetilde Y_{\mathrm{sig}}^\top .
\end{equation*}
Since \(\widetilde C_{\mathrm{sig}}\) is positive semidefinite,
\begin{equation*}
    \lambda_{\max}(\widetilde C_{\mathrm{sig}})
    \le
    \operatorname{tr}(\widetilde C_{\mathrm{sig}}).
\end{equation*}
Moreover,
\begin{equation*}
    \operatorname{tr}(\widetilde C_{\mathrm{sig}})
    =
    \frac{1}{d}
    \operatorname{tr}
    \left(
        \widetilde Y_{\mathrm{sig}}
        \widetilde Y_{\mathrm{sig}}^\top
    \right)
    =
    \frac{1}{d}
    \|\widetilde Y_{\mathrm{sig}}\|_F^2
    =
    a^2\alpha_s .
\end{equation*}
Therefore,
\begin{equation*}
    \lambda_{\max}(\widetilde C_{\mathrm{sig}})
    \le
    a^2\alpha_s .
\end{equation*}
This proves the claim.
\end{proof}

\subsection{Proof of Proposition~\ref{prop:bbp_suppression_thinning}}

\begin{proof}
Let \(P_{\mathcal I}\in\{0,1\}^{n'\times n}\) be a row-selection matrix that
extracts the rows indexed by \(\mathcal I\), and suppose \(P_{\mathcal I}\)
is independent of \(Z\). Applying \(P_{\mathcal I}\) to the spiked
representation gives
\begin{equation*}
    P_{\mathcal I}Y
    =
    \sqrt d\,a\,(P_{\mathcal I}s)u^\top
    +
    \sigma P_{\mathcal I}Z.
\end{equation*}
Define
\begin{equation*}
    \alpha_s
    :=
    \|P_{\mathcal I}s\|_2^2.
\end{equation*}
If \(\alpha_s>0\), let
\begin{equation*}
    \bar s
    :=
    \frac{P_{\mathcal I}s}{\|P_{\mathcal I}s\|_2}.
\end{equation*}
Then \(P_{\mathcal I}s=\sqrt{\alpha_s}\bar s\), and hence
\begin{equation*}
    P_{\mathcal I}Y
    =
    \sqrt d\,a\sqrt{\alpha_s}\,\bar s u^\top
    +
    \sigma Z',
\end{equation*}
where \(Z':=P_{\mathcal I}Z\). Since \(Z\) has i.i.d. standard Gaussian
entries and \(P_{\mathcal I}\) selects rows independently of \(Z\), the matrix
\(Z'\in\mathbb{R}^{n'\times d}\) also has i.i.d. standard Gaussian entries.

Therefore,
\begin{equation*}
    \widetilde C
    =
    \frac{1}{d}
    (P_{\mathcal I}Y)(P_{\mathcal I}Y)^\top
\end{equation*}
is a standard rank-one spiked Wishart covariance matrix with aspect ratio
\(c'=n'/d\) and effective spike strength \(a_{\mathrm{eff}}^2=a^2\alpha_s\).
By the classical BBP phase transition, the leading eigenvalue separates from
the Marchenko--Pastur bulk only when
\begin{equation*}
    a^2\alpha_s>\sigma^2\sqrt{c'}.
\end{equation*}
Equivalently, if
\begin{equation*}
    a^2\alpha_s\le\sigma^2\sqrt{c'},
\end{equation*}
the spike is subcritical and the associated spectral outlier disappears, with
the top eigenvalue converging to
\begin{equation*}
    \lambda_+
    =
    \sigma^2(1+\sqrt{c'})^2 .
\end{equation*}
\end{proof}

\subsection{Discussion: post-SDPA Gates and the Scope of the BBP statement}
\label{app:learned_gate_bbp_scope}

Theorem~\ref{thm:sink_mode_contraction} is deterministic and applies to learned post-SDPA gates. It only uses the transformation of the sink-induced signal row
\begin{equation*}
    \sqrt das_i u^\top
    \mapsto
    \sqrt das_i(\Gamma_i u)^\top .
\end{equation*}
Thus the contraction factor
\begin{equation*}
    \alpha_s
    =
    \sum_i s_i^2\|\Gamma_i u\|_2^2
\end{equation*}
does not rely on any randomness assumption about the gate or on coordinate localization of \(s\). However, for the full representation
\(
    Y
    =
    \sqrt dasu^\top+\sigma Z,
\)
the learned gate also transforms the noise component. The \(i\)-th noise row is mapped from \(\sigma z_i\) to \(\sigma\Gamma_i z_i\). Unless the matrices \(\Gamma_i\) satisfy additional isotropy assumptions, this noise is feature-dependent and token-dependent. Consequently, the gated covariance is not exactly a standard spiked Wishart matrix, and the classical BBP transition cannot be applied as an exact theorem for learned gates.

Proposition~\ref{prop:bbp_suppression_thinning} isolates an idealized setting in which the same contraction mechanism can be connected exactly to BBP. In the row-selection model, the retained noise matrix remains i.i.d. Gaussian, so the thinned covariance is again a standard spiked Wishart model. Learned gates are not random row selectors, but the deterministic contraction result still applies. Therefore, the BBP condition should be read as a mechanism-level reference: when the contracted sink-induced signal becomes weak relative to the noise bulk, the associated spectral outlier is expected to weaken or disappear. This motivates the empirical diagnostics used in Section~\ref{sec:forward_empirical}.

\subsection{Discussion: Higher-Rank Sink Subspaces and Contraction}
\label{app:higher_rank_sinks}

Our empirical sink object is the multi-column subspace
\begin{equation}
\mathcal S_{\mathrm{sink}}
=\operatorname{span}\{\bar a_j:j\in J_{\mathrm{sink}}\},
\label{eq:appendix_sink_subspace}
\end{equation}
where $J_{\mathrm{sink}}$ may contain multiple positions, including positions other than the first. The rank-one component analyzed in the main text isolates one elementary sink contribution; it is not an assumption that all empirical sinks are literally rank one. For any combined sink component $Y_{\mathrm{sink}}$, an elementwise post-SDPA gate with entries in $[0,1]$ satisfies
\begin{equation}
\|G\odot Y_{\mathrm{sink}}\|_F^2
\leq\|Y_{\mathrm{sink}}\|_F^2.
\label{eq:appendix_higher_rank_contraction}
\end{equation}
This global inequality does not imply independent scalar contraction of every spectral spike: gating may change the relative mode strengths and rotate the resulting eigenspace. We therefore evaluate the full sink-subspace projector.

The contraction applies to the combined higher-rank sink contribution, but it does not imply that a rank-$r$ sink subspace decomposes into $r$ independently contracted eigenmodes. This distinction motivates the empirical subspace-level tests in Appendix~\ref{app:nonfirst_sinks}.

\section{Proofs for the Gated Gradient-Sink Bound}
\label{app:gradient_proofs}

\subsection{Proof of Theorem~\ref{thm:gated_value_path_gradient_bound}}

\begin{proof}
For each query position \(t\), the post-SDPA gated attention output is
\(\widetilde y_t = G_t \odot y_t\). Let
\(\delta_t=\nabla_{\widetilde y_t}\mathcal L\). By the chain rule,
\[
    \nabla_{y_t}\mathcal L = G_t\odot \delta_t.
\]
Since \(y_t = \sum_{j\le t} a_{tj}v_j\), the Jacobian of \(y_t\) with respect to
\(v_s\) is
\[
    \frac{\partial y_t}{\partial v_s}
    =
    \begin{cases}
    a_{ts} I, & s\le t,\\
    0, & s>t.
    \end{cases}
\]
Therefore,
\[
    \nabla_{v_s}\mathcal L
    =
    \sum_{t=s}^{T}
    a_{ts}\nabla_{y_t}\mathcal L
    =
    \sum_{t=s}^{T}
    a_{ts}(G_t\odot\delta_t).
\]
Taking norms and viewing \(G_t\) as a diagonal linear map gives
\[
    \|\nabla_{v_s}\mathcal L\|
    \le
    \sum_{t=s}^{T}
    a_{ts}\|G_t\odot\delta_t\|
    \le
    \sum_{t=s}^{T}
    a_{ts}\|G_t\|_{\mathrm{op}}\|\delta_t\|.
\]
If \(\|\delta_t\|\le B\) for every \(t\), then
\[
    \|\nabla_{v_s}\mathcal L\|
    \le
    B\sum_{t=s}^{T}a_{ts}\|G_t\|_{\mathrm{op}}
    =
    B M_s^{\mathrm{eff}}.
\]
Since \(G_t\in[0,1]^d\), we have \(\|G_t\|_{\mathrm{op}}\le 1\), and hence
\(M_s^{\mathrm{eff}}\le M_s\).
\end{proof}

\subsection{Proof of Corollary~\ref{cor:gated_second_moment}}
\label{app:second_moment_gradient_attenuation}

\begin{proof}
From Theorem~\ref{thm:gated_value_path_gradient_bound},
\[
    \nabla_{v_s}\mathcal L
    =
    \sum_{t=s}^{T} a_{ts}(G_t\odot\delta_t).
\]
Expanding the squared norm and using the cross-position uncorrelatedness
assumption, the cross terms vanish:
\[
    \mathbb E\|\nabla_{v_s}\mathcal L\|^2
    =
    \sum_{t=s}^{T}
    a_{ts}^2
    \mathbb E\|G_t\odot\delta_t\|^2.
\]
Since \(G_t\) acts as a diagonal linear map,
\[
    \|G_t\odot\delta_t\|^2
    \le
    \|G_t\|_{\mathrm{op}}^2\|\delta_t\|^2.
\]
Therefore,
\[
    \mathbb E\|\nabla_{v_s}\mathcal L\|^2
    \le
    B^2
    \sum_{t=s}^{T}
    a_{ts}^2\|G_t\|_{\mathrm{op}}^2
    =
    B^2 S_s^{\mathrm{eff}}.
\]
Finally, \(G_t\in[0,1]^d\) implies \(\|G_t\|_{\mathrm{op}}\le 1\), and hence
\(S_s^{\mathrm{eff}}\le S_s\).
\end{proof}

\subsection{Local Value-path Attenuation versus Global Training Dynamics}
\label{app:local_vs_global_gradient_dynamics}

Our gradient results are local statements about the value-path Jacobian. Conditioned on fixed attention weights and upstream gradients \(\delta_t=\nabla_{\widetilde y_t}\mathcal L\), post-SDPA gating filters each contribution from query position \(t\) to source value \(v_s\). Equivalently, the raw sink-column statistics \(M_s\) and \(S_s\) are replaced by their gated counterparts \(M_s^{\mathrm{eff}}\) and \(S_s^{\mathrm{eff}}\). This does not imply that all gradients in a trained model with post-SDPA gating are uniformly smaller than in a model without post-SDPA gating, since gating also changes the forward computation, upstream gradients, and learned representations. Rather, the structural claim is that for a fixed upstream signal, the value-path map through a sink column is contractive. The deterministic and second-moment bounds cover coherent and incoherent aggregation regimes, respectively, and in both cases
\[
    M_s^{\mathrm{eff}}\le M_s,
    \qquad
    S_s^{\mathrm{eff}}\le S_s.
\]

Thus, post-SDPA gating weakens the attention sink to gradient sink link without necessarily removing the attention sink itself: a sink column may still receive attention, but each value-path contribution is filtered before reaching the source value. This provides a local mechanism for reducing gradient pressure in the attention-sink--gradient-sink--massive-activation pathway, without claiming a full theory of global training dynamics.

\section{Experimental Setting} \label{app:analysis_setting}
\paragraph{Hardware.}
All experiments were conducted on a server equipped with two AMD EPYC~9654 96-core CPUs (192 physical cores / 384 threads in total, 1.5--3.7\,GHz, 768\,MiB L3 cache) and 1\,TB of system memory. The server is configured with eight NVIDIA RTX~PRO~6000 Blackwell Max-Q Workstation GPUs (96\,GB GDDR7 each, driver 580.95.05). All code was implemented in PyTorch~2.9.1 built against CUDA~12.8 (toolkit \texttt{nvcc} 12.8.93) on Python~3.13.5.

\paragraph{Model checkpoints.}
For the Qwen analyses, we use the public checkpoints released by a previous work~\citep{qiu2025gated} in the \texttt{QwQZh/gated\_attention} HuggingFace repository.\footnote{\url{https://huggingface.co/QwQZh/gated_attention}} Specifically, we analyze \texttt{1B\_baseline} and \texttt{1B\_gate\_elementwise}. All models are evaluated in inference mode, with no additional training or parameter updates.

\paragraph{Training setup for training-time diagnostics.}
For the training-time diagnostics, we train Pythia-410M baseline and post-SDPA gated variants using the HuggingFace training framework, following the language-modeling training protocol of~\cite{behrouz2024titans}. The two models are trained under matched data, optimization, and checkpointing conditions. The baseline model uses 24 Transformer layers, hidden size \(1024\), 16 attention heads, and MLP intermediate size \(4096\). The elementwise gated variant uses the same configuration except for the post-SDPA elementwise gate and a reduced MLP intermediate size of \(3584\), which compensates for the additional gate parameters. Both models use the \texttt{EleutherAI/gpt-neox-20b} tokenizer and are trained on 15B tokens from FineWeb-Edu~\cite{penedo2024fineweb} with causal language modeling, sequence length \(4096\), AdamW optimization, and peak learning rate \(4\times 10^{-4}\). We save checkpoints throughout training and evaluate without further parameter updates.

To examine robustness beyond the original setting, we also analyze matched baseline and post-SDPA elementwise-gated Pythia-2.8B models; matched Pythia-410M models trained with an 8K context length and evaluated at 4K or 8K; and matched Pythia-410M models trained with document-boundary EOS separators. The plain sequence convention uses \texttt{add\_bos\_token = False} and \texttt{add\_eos\_token = False}; the EOS-separated convention uses \texttt{add\_bos\_token = False} and \texttt{add\_eos\_token = True}. Protocol-specific evaluation conditions and measurements are provided in Appendices~\ref{app:scale_28b}--\ref{app:special_tokens}. 

\paragraph{Language modeling and zero-shot commonsense reasoning.}
Following prior works~\citep{behrouz2024titans,shin2026graph}, we evaluate baseline and post-SDPA gated variants across two evaluation categories: WikiText (Wiki.;~\cite{wikitext}) perplexity and zero-shot commonsense reasoning tasks. The commonsense tasks include LAMBADA (LMB.;~\cite{paperno-etal-2016-lambada}), PIQA~\cite{bisk2019piqareasoningphysicalcommonsense}, HellaSwag (Hella.;~\cite{zellers-etal-2019-hellaswag}), WinoGrande (Wino.;~\cite{sakaguchi2019winograndeadversarialwinogradschema}), ARC-easy (ARC-e) and ARC-challenge (ARC-c)~\cite{clark2018thinksolvedquestionanswering}, BoolQ~\cite{clark2019boolq}, and OpenbookQA (OBQA.;~\cite{mihaylov2018can}).

\begin{figure}[t]
    \centering
    \subfigure[Outlier mass]{\includegraphics[width=0.31\linewidth]{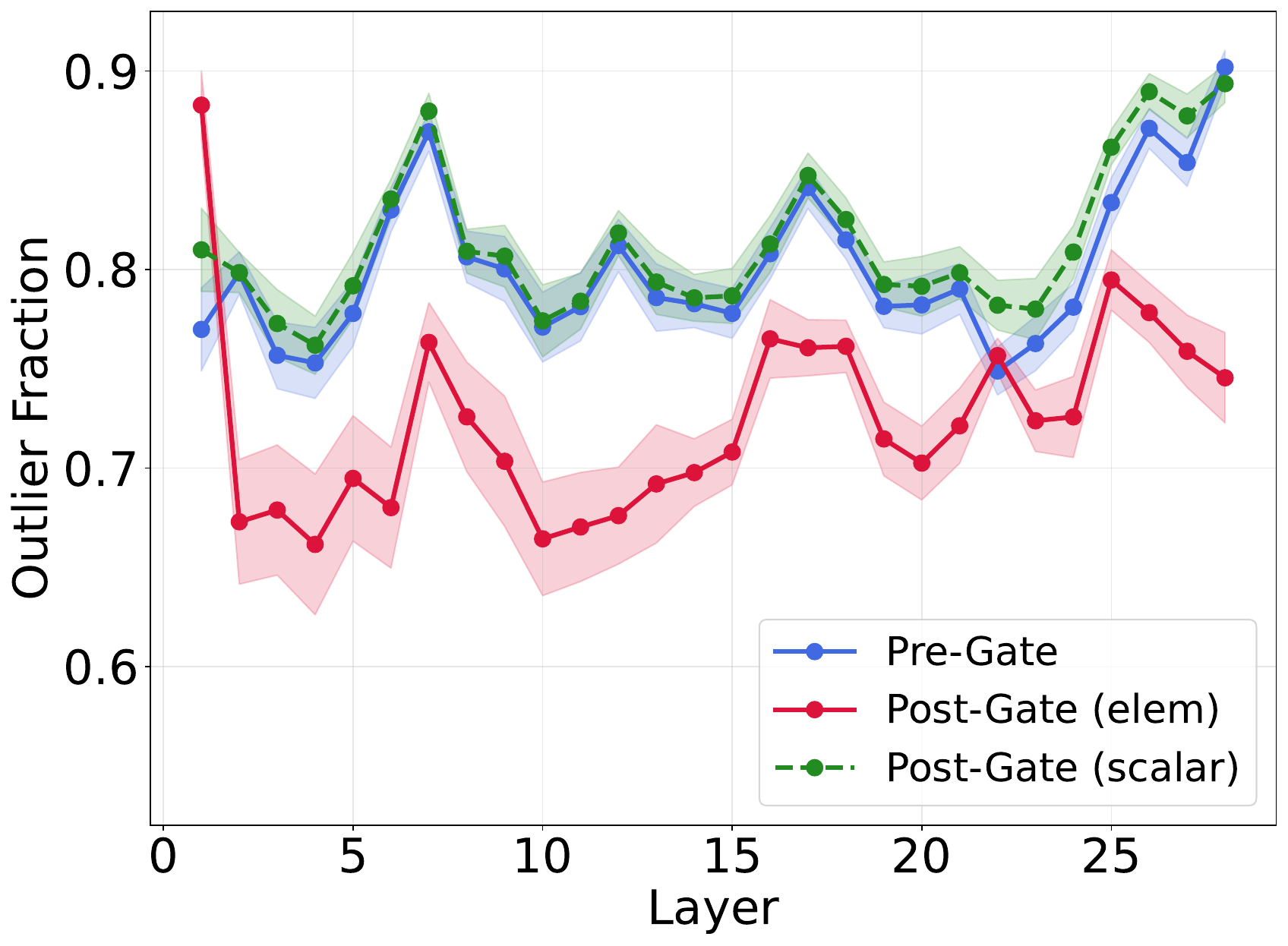}
        }
    \hfill
    \subfigure[Effective rank]{
        \includegraphics[width=0.31\linewidth]{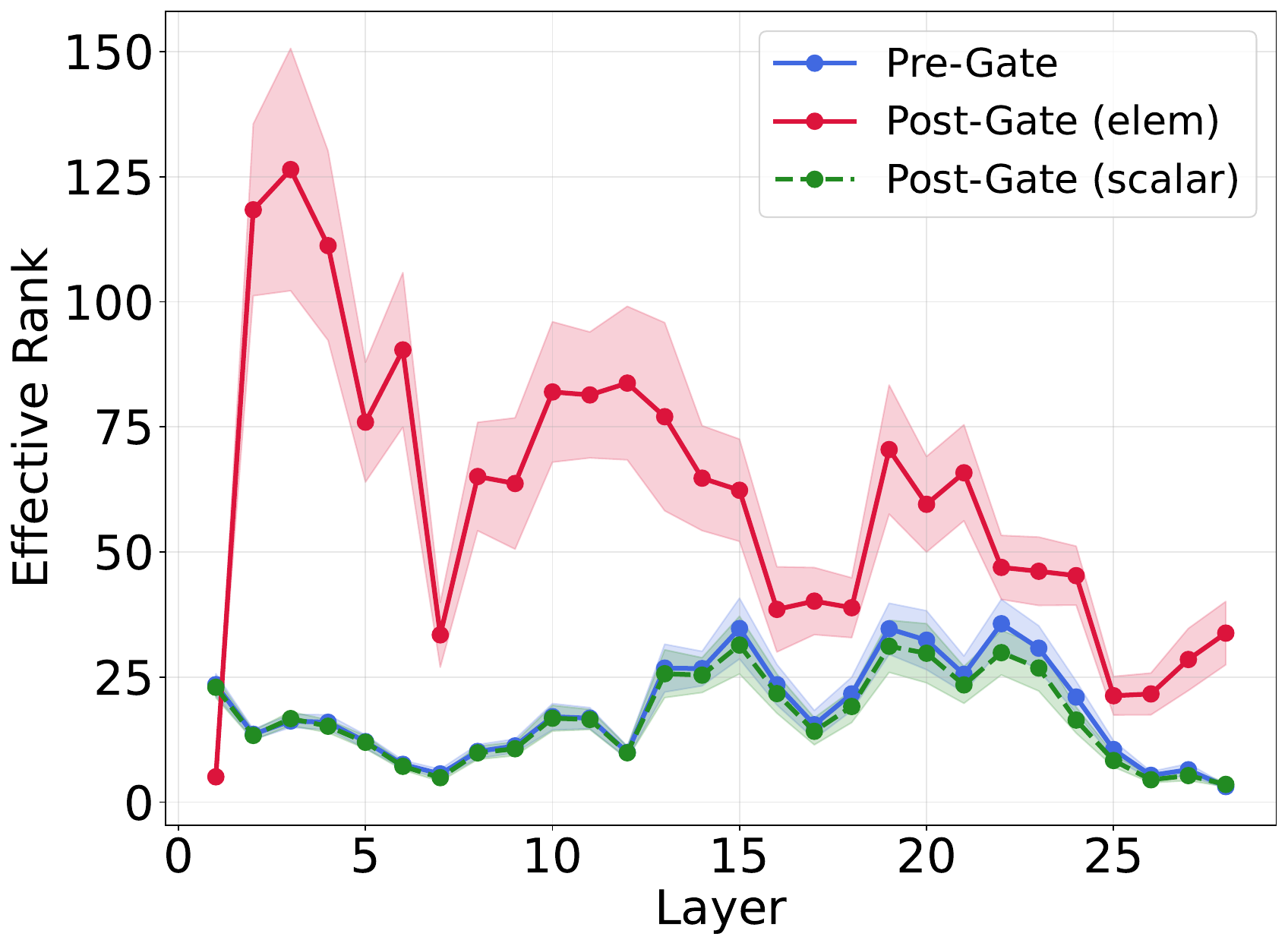}}
    \hfill
    \subfigure[\(\lambda_{\max}\) ratio]{
        \includegraphics[width=0.31\linewidth]{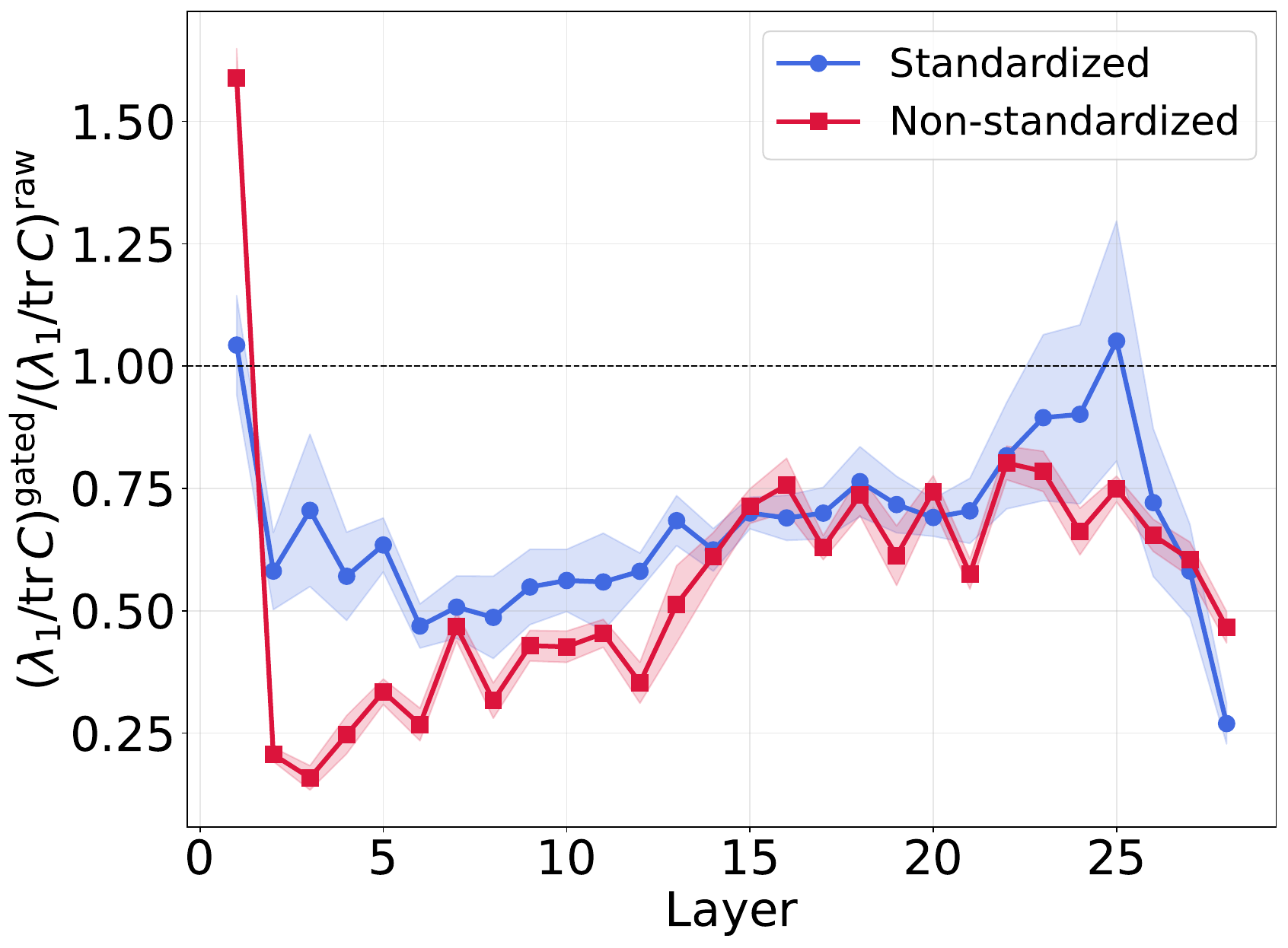}}
    \caption{
    \textbf{Robustness to standardization and gate control.}
    Without standardization, the post-SDPA gate also reduces outlier mass and increases effective rank. A feature-averaged scalar gate preserves token-wise gate strength but removes feature-direction-specific selectivity, and does not reproduce the same spectral suppression.
    }
    \label{fig:nonstd_robustness}
\end{figure}

\section{Additional RMT Diagnostics}
\label{app:rmt_controls}

This appendix provides additional diagnostics supporting the RMT analysis in Section~\ref{sec:forward_empirical}. The main text focuses on spectral suppression and sink-subspace alignment. We report robustness checks, gate statistics, sink-subspace counterfactuals, contraction diagnostics, and bulk-level spectral analyses.

\subsection{Robustness to Covariance Normalization and Feature-Averaged Gate Control}
\label{app:rmt_nonstd}
In the main text, we compute token-space covariance after feature-wise standardization across tokens. This removes trivial feature-scale effects and focuses the diagnostics on token-space correlation structure. To verify that the observed suppression is not an artifact of this normalization, we repeat the analysis using the non-standardized covariance
\(
    C_Y^{\mathrm{raw}} = \frac{1}{D}YY^\top .
\)
Since the overall scale of \(C_Y^{\mathrm{raw}}\) is not fixed, we rescale the MP edge by \(\operatorname{tr}(C_Y^{\mathrm{raw}})/T\) so that the outlier mass is invariant to global rescaling.
We also compare the learned elementwise gate with a feature-averaged gate control. For each token, we replace the feature-wise gate vector \(G_i\in[0,1]^D\) by its feature average
\(
    \bar G_i = \frac{1}{D}\sum_{c=1}^{D}G_{i,c},
\)
and apply the scalar gate \(\bar G_i I\) to the SDPA output. This control preserves token-wise gate strength, so it can still suppress particular token positions, but removes feature-direction-specific selectivity.

Figure~\ref{fig:nonstd_robustness} shows that the post-SDPA gate still suppresses dominant spectral structure without feature-wise standardization. In particular, compared with Pre-Gate, Post-Gate (elem) has lower outlier mass and higher effective rank in the non-standardized covariance across most layers. By contrast, Post-Gate (scalar) does not reproduce the same suppression pattern: its outlier mass and effective rank remain close to those of Pre-Gate, indicating that average token-wise shrinkage alone is not sufficient. Figure~\ref{fig:nonstd_robustness}(c) further shows that the reduction in leading-mode concentration \(\lambda_1/\operatorname{tr}(C)\) from Pre-Gate to Post-Gate (elem) is at least as strong without standardization as with it, so feature-wise standardization does not create or inflate the observed suppression. These results indicate that the spectral suppression is not merely a covariance-normalization artifact or a consequence of average token-wise shrinkage. Rather, the feature-selective action of the learned post-SDPA gate is important for suppressing dominant token-space spectral structure. The following sections further examine where this selective action appears in gate statistics and how it relates to sink-induced subspaces.

\subsection{Learned Selectivity Beyond Generic Mode Contraction}
\label{app:gate_selectivity}

The contraction result in Theorem~\ref{thm:sink_mode_contraction} applies to any individual mode; it does not, by itself, establish preferential suppression of sink-aligned directions. To make this distinction explicit, consider a rank-one pre-gate component $Y_k=\sigma_k s_k u_k^{\top}$, where $\|s_k\|_2=\|u_k\|_2=1$. For the elementwise gate $G$, with $\Gamma_i$ denoting the diagonal matrix of gate values at token $i$, its raw-energy retention is
\begin{equation}
\alpha_k
=\frac{\|G\odot Y_k\|_F^2}{\|Y_k\|_F^2}
=\sum_i s_{k,i}^2\|\Gamma_i u_k\|_2^2\leq 1.
\label{eq:appendix_mode_retention}
\end{equation}
This inequality does not order sink- and non-sink-aligned modes by their retention. 
We evaluate this distinction using the 64 leading pre-gate modes of the elementwise-gated Qwen-1B model, comparing the top and bottom terciles of sink alignment \(\|P_{\mathrm{sink}}s_k\|_2^2\) (Table~\ref{tab:appendix_gate_selectivity}).

\begin{table}[!htbp]
\centering
\caption{\textbf{Mode retention for strongly vs. weakly sink-aligned pre-gate directions in Qwen-1B (Gated).} Raw-energy retention is computed on individual rank-one components; the other rows use feature-standardized token-space covariance.}
\label{tab:appendix_gate_selectivity}
\small
\begin{tabular}{lccc}
\toprule
Retention metric & High alignment & Low alignment & Difference \\
\midrule
Raw activation energy ($\alpha_k$) & 0.047 & 0.044 & $+0.003$ \\
Standardized covariance ($R_k$) & 0.740 & 0.945 & $-0.205$ \\
Scalar gate ($R_k$) & 0.957 & 0.990 & $-0.033$ \\
\bottomrule
\end{tabular}
\end{table}

We therefore additionally examine retention in the feature-standardized token-space covariance geometry used for the main spectral analysis. For each pre-gate covariance eigenvector $q_k$, define
\begin{equation}
R_k=\frac{q_k^\top C_{\mathrm{post}}q_k}
{q_k^\top C_{\mathrm{pre}}q_k},
\label{eq:appendix_standardized_retention}
\end{equation}
where $C_{\mathrm{pre}}$ and $C_{\mathrm{post}}$ are the pre- and post-gate feature-standardized token-space covariance matrices. This standardized geometry reveals a directional difference that is absent from raw-energy retention. 
The negative association between sink alignment and retention persists after controlling for pre-gate spectral energy, so it cannot be explained merely by larger initial eigenvalues. 
As an ablation, we replace each token's featurewise gate values by their feature average, removing within-token feature selectivity. 
The resulting retention gap becomes substantially smaller, as shown in Table~\ref{tab:appendix_gate_selectivity}. Thus, the gate is nearly non-selective in raw activation energy but selectively reshapes the feature-standardized covariance through its learned feature-dependent structure. This empirical selectivity is additional to, rather than a consequence of, the unconditional mode-wise contraction bound.
\subsection{Post-SDPA Gate-score Statistics}
\label{app:gate_score_stats}

\begin{figure}[t]
    \centering
    \subfigure[Gate at sink vs. non-sink positions]{
        \includegraphics[width=0.31\linewidth]{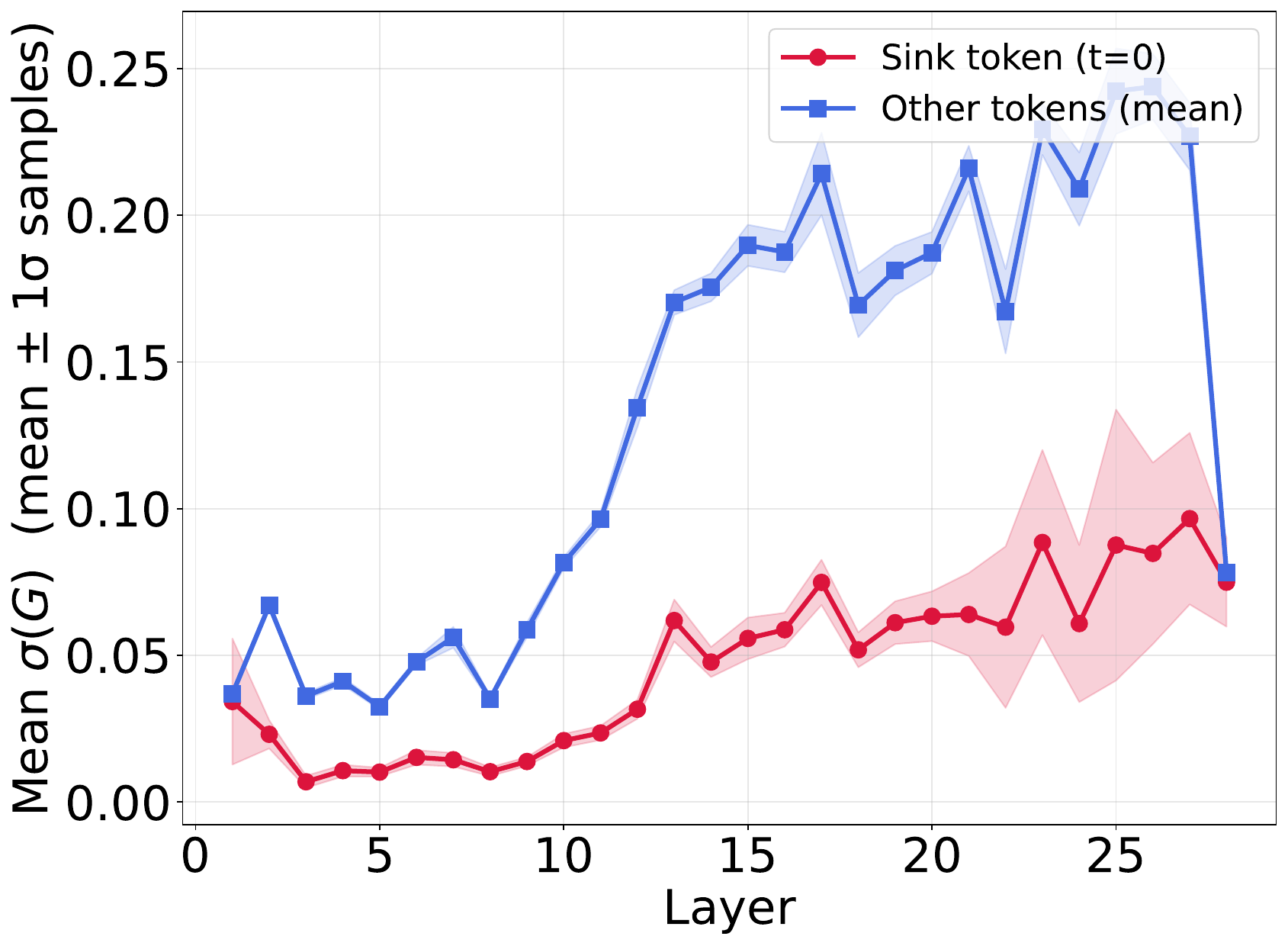}}
    \hfill
    \subfigure[Token-wise gate heatmap]{
        \includegraphics[width=0.31\linewidth]{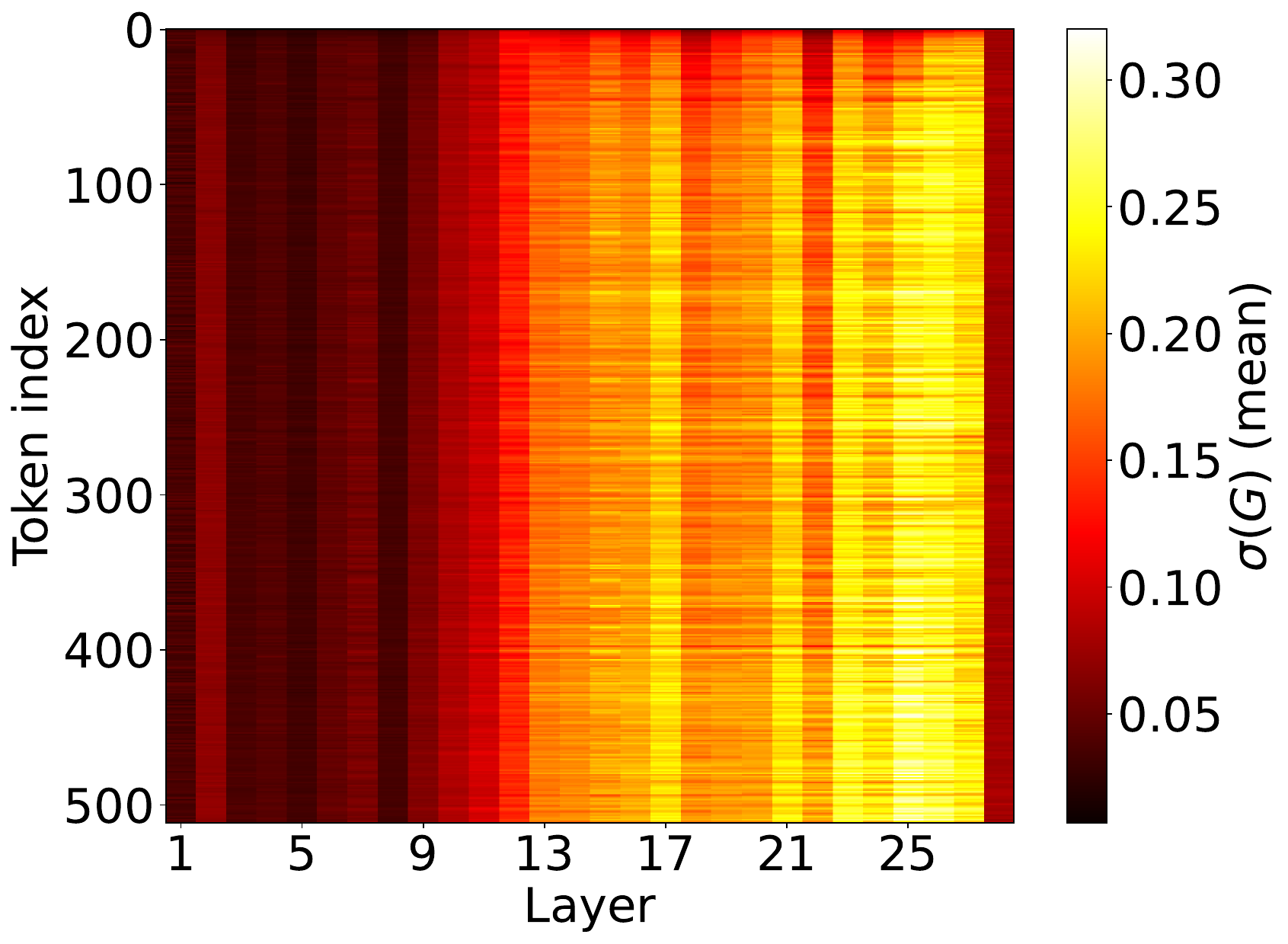}}
    \hfill
    \subfigure[Head-wise sink/non-sink gate ratio]{ \includegraphics[width=0.31\linewidth]{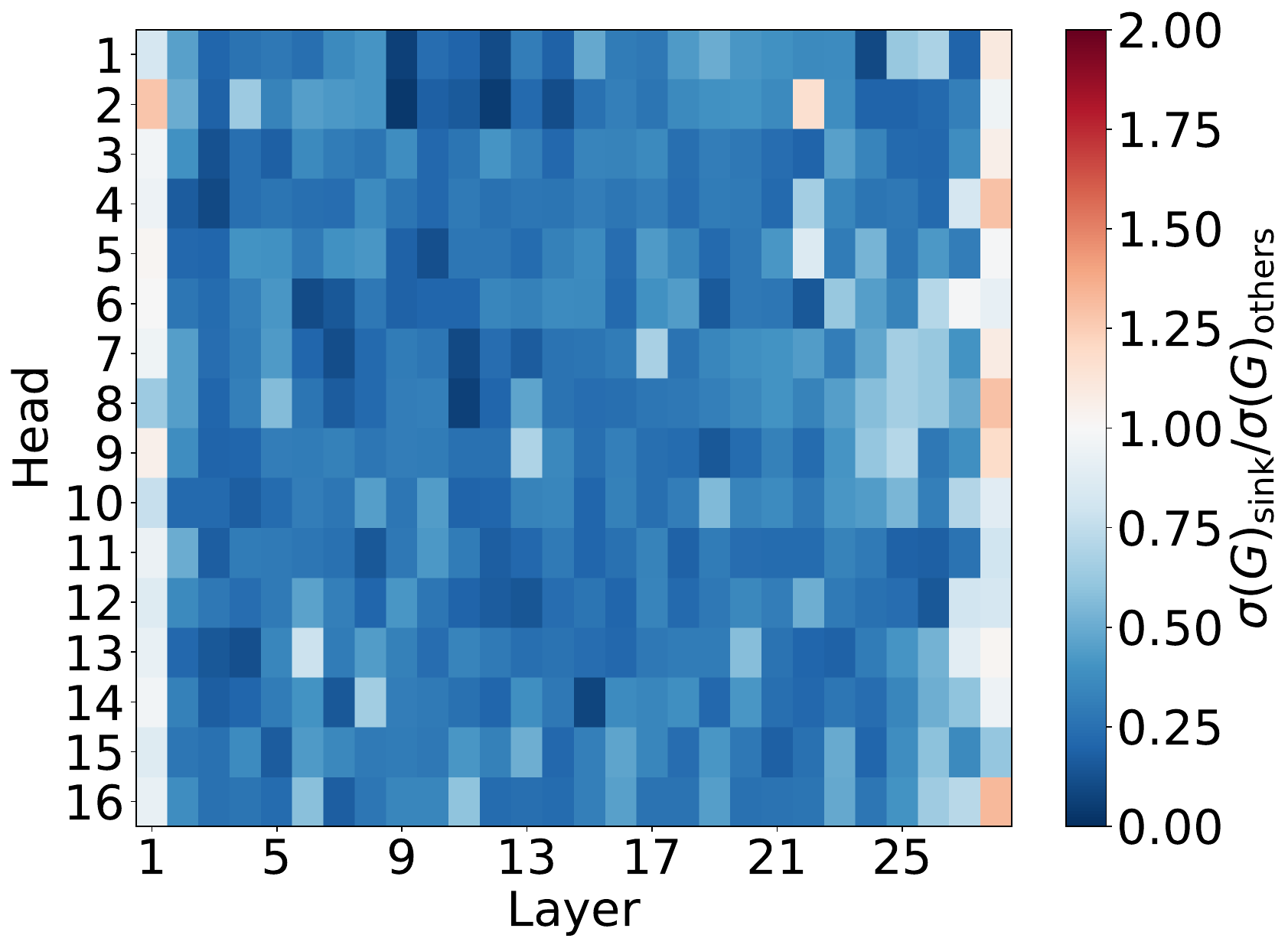}}
    \caption{
    \textbf{Gate-score statistics.} The post-SDPA gate is not spatially uniform: its strength varies across depth and is often concentrated around sink-related positions and hotspot heads. These statistics provide an interpretable view of the learned filtering pattern.
    }
    \label{fig:gate_score_stats}
\end{figure}
We also inspect the post-SDPA gate scores directly. While average gate values do not by themselves determine spectral suppression, they provide an interpretable view of where the model applies stronger filtering. We compute gate statistics across layers, tokens, and heads, and compare the gate values at sink-related positions with those at non-sink positions. Figure~\ref{fig:gate_score_stats} shows that gate scores vary substantially across depth and heads. In layers where spectral spikes are pronounced, the gate tends to apply stronger filtering to sink-related positions and hot-spot heads. This is consistent with the contraction analysis in Section~\ref{sec:forward_spectral}: suppression depends not only on the average gate magnitude, but on how the learned gate acts on the dominant representation modes.

\subsection{Structural and Bulk Diagnostics}
\label{app:rmt_structural_bulk}

\begin{figure}[t]
    \centering
    \subfigure[Sink-direction alignment]{
    \includegraphics[width=0.23\linewidth]{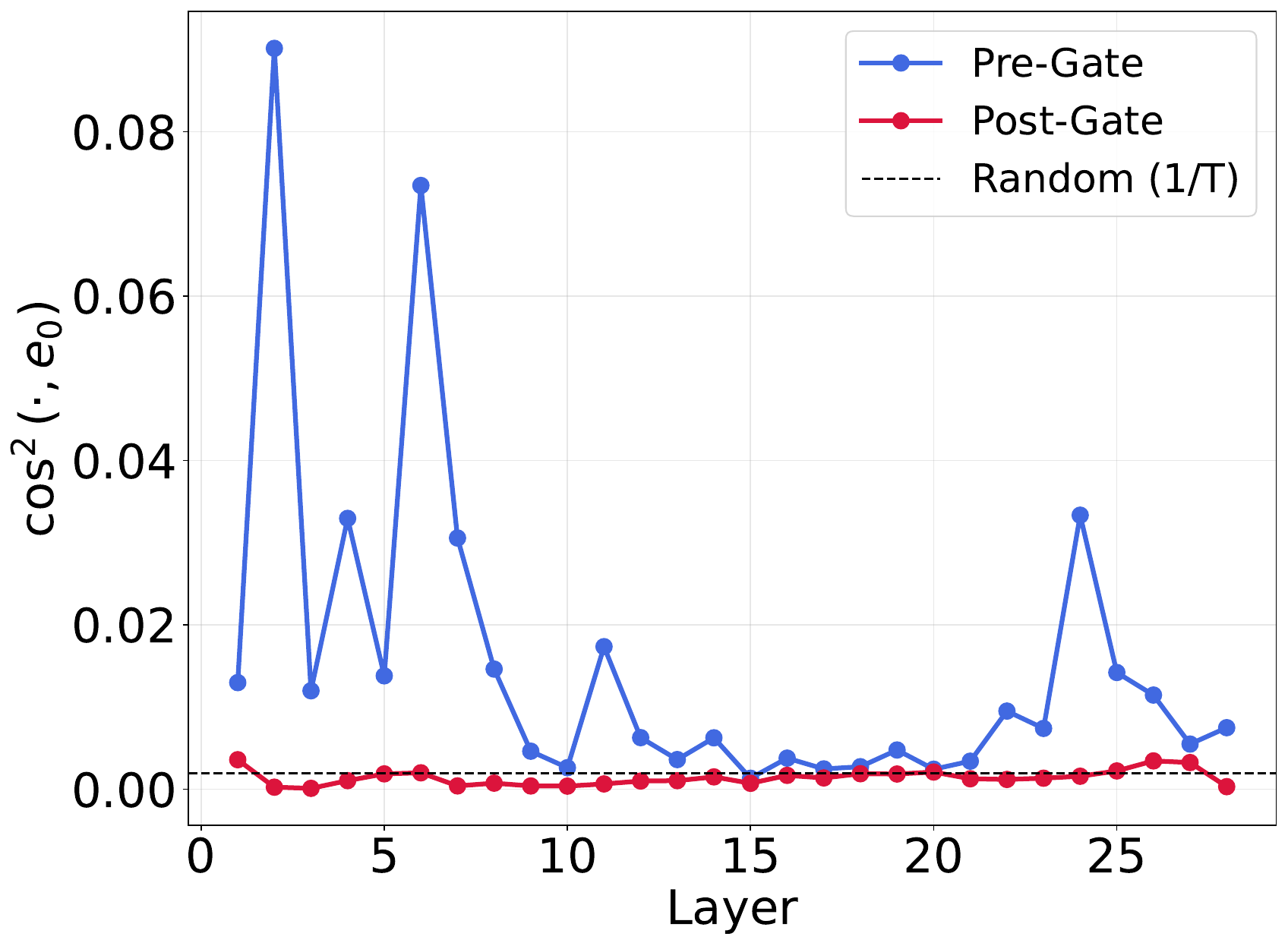}}
    \subfigure[Head-wise \(\lambda_1\)]{
        \includegraphics[width=0.23\linewidth]{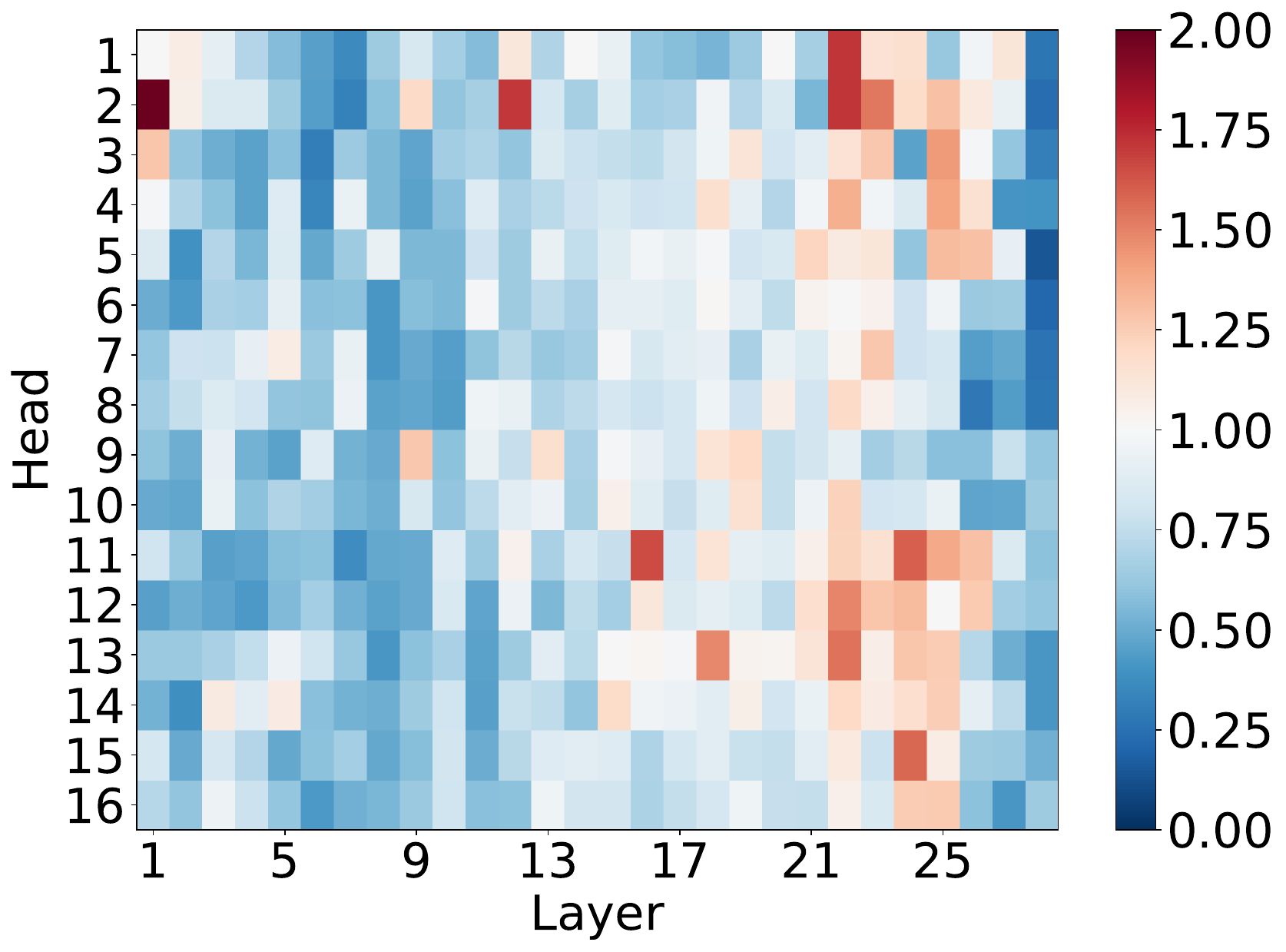}}
    \subfigure[Tail CCDF]{
        \includegraphics[width=0.23\linewidth]{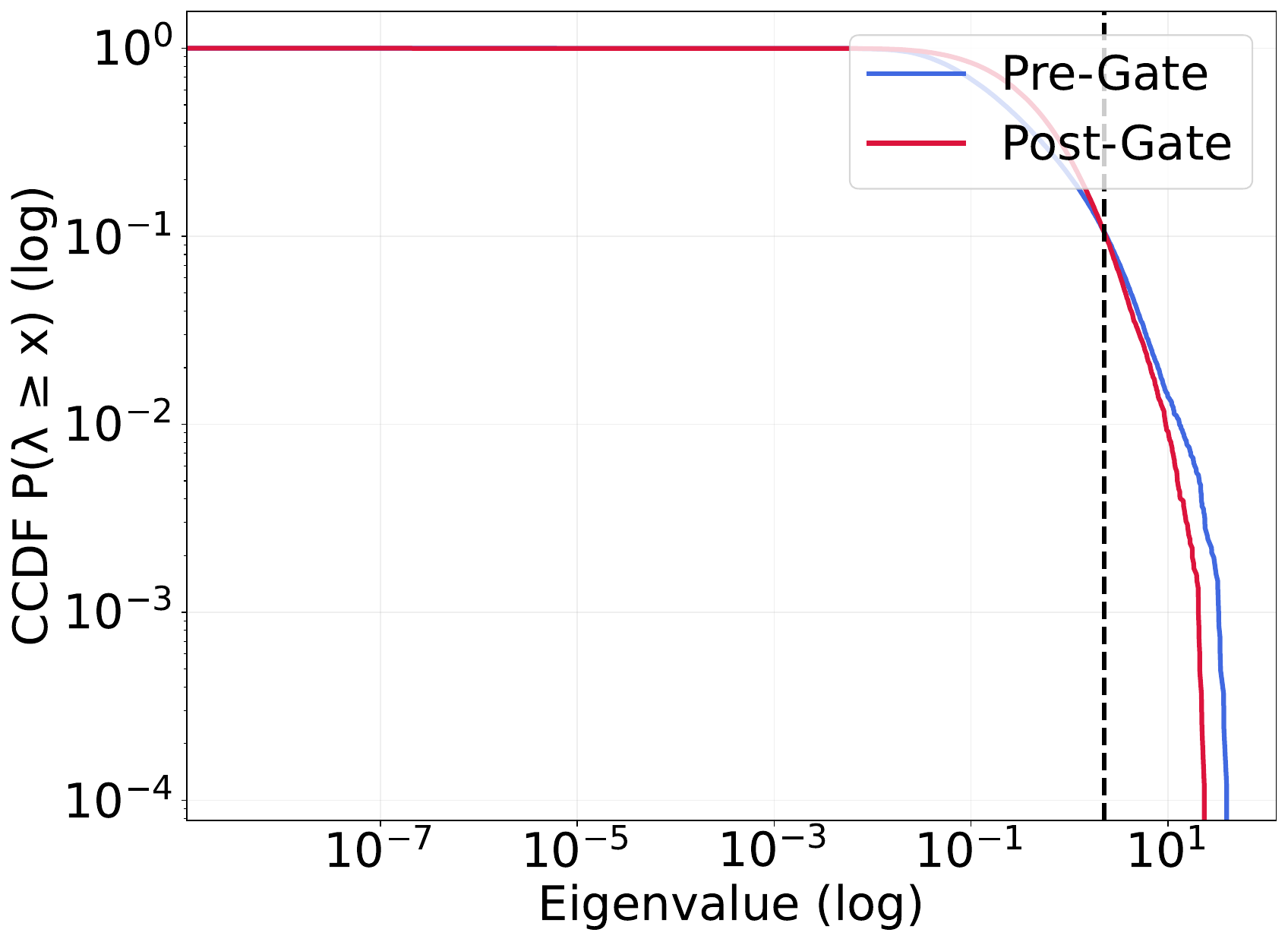}}
    \subfigure[Bulk spacing ratio]{
        \includegraphics[width=0.23\linewidth]{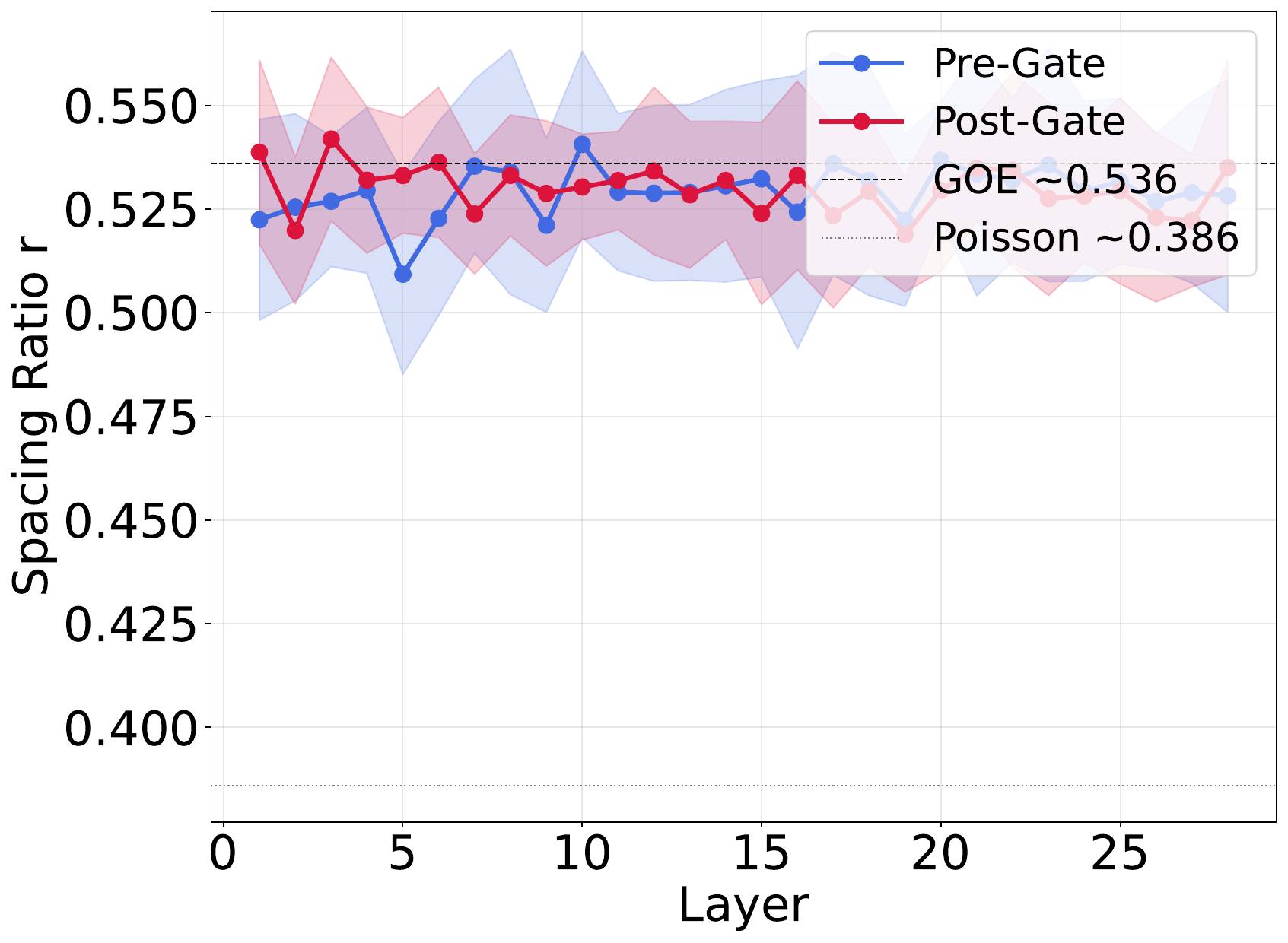}}
    \caption{
    \textbf{Structural and bulk diagnostics.}
    (a) Leading-eigenvector alignment with the first-token direction is reduced where such alignment is present.
    (b) Head-wise $\lambda_1$ ratios show that localized hotspot heads with large leading eigenvalues tend to be suppressed by gating.
    (c) The eigenvalue-tail CCDF \(P(\lambda \ge x)\) shows reduced heavy-tail dominance after gating.
    (d) Bulk spacing ratios remain stable, indicating that gating suppresses outlier structure while largely preserving the spectral bulk.
    }
    \label{fig:structural_bulk}
\end{figure}
We report a set of complementary diagnostics that probe different facets of the post-SDPA spectrum, separating outlier-level structure from bulk behavior. Where the main analysis (Section~\ref{sec:forward_empirical}) used the data-adaptive sink subspace 
$P_\text{sink}$, here we additionally measure the alignment of the leading eigenvector with the simpler first-token reference direction $e_0$.
Second, we compute head-wise token-space spectra to identify whether dominant modes are distributed uniformly across heads or concentrated in a small number of hotspot heads. Third, we examine the eigenvalue tail using the complementary cumulative distribution function (CCDF), \(P(\lambda \ge x)\), which summarizes how much spectral mass remains in large eigenvalues. Finally, after removing eigenvalues above the Marchenko–Pastur upper edge $\lambda_+$ (i.e., the bulk only), we compute the adjacent spacing ratio
\[
    r_i=\frac{\min(\Delta_i,\Delta_{i+1})}{\max(\Delta_i,\Delta_{i+1})},
    \qquad
    \Delta_i=\lambda_{i+1}-\lambda_i.
\] and report its layer-wise mean. This statistic tests whether gating disrupts the local bulk spectrum, with typical reference values near the GOE and Poisson limits shown in the figure.

Figure~\ref{fig:structural_bulk} summarizes these diagnostics. Leading eigenspace alignment decreases in layers where sink-related alignment is elevated, and head-wise spectra show that large leading eigenvalues are localized to a small number of heads. Gating reduces these localized spectral hotspots. The tail CCDF shows reduced heavy-tail dominance after gating, while the bulk spacing ratio remains stable. Together, these results support the view that post-SDPA gating primarily suppresses outlier and heavy-tail components associated with sink structure, rather than broadly disrupting the spectral bulk.

\subsection{Sink-subspace Counterfactuals and Contraction Diagnostics}
\label{app:sink_subspace_diagnostics}

\begin{wrapfigure}{r}{0.4\linewidth}
    \vspace{-1.0em}
    \centering
    \includegraphics[width=0.9\linewidth]{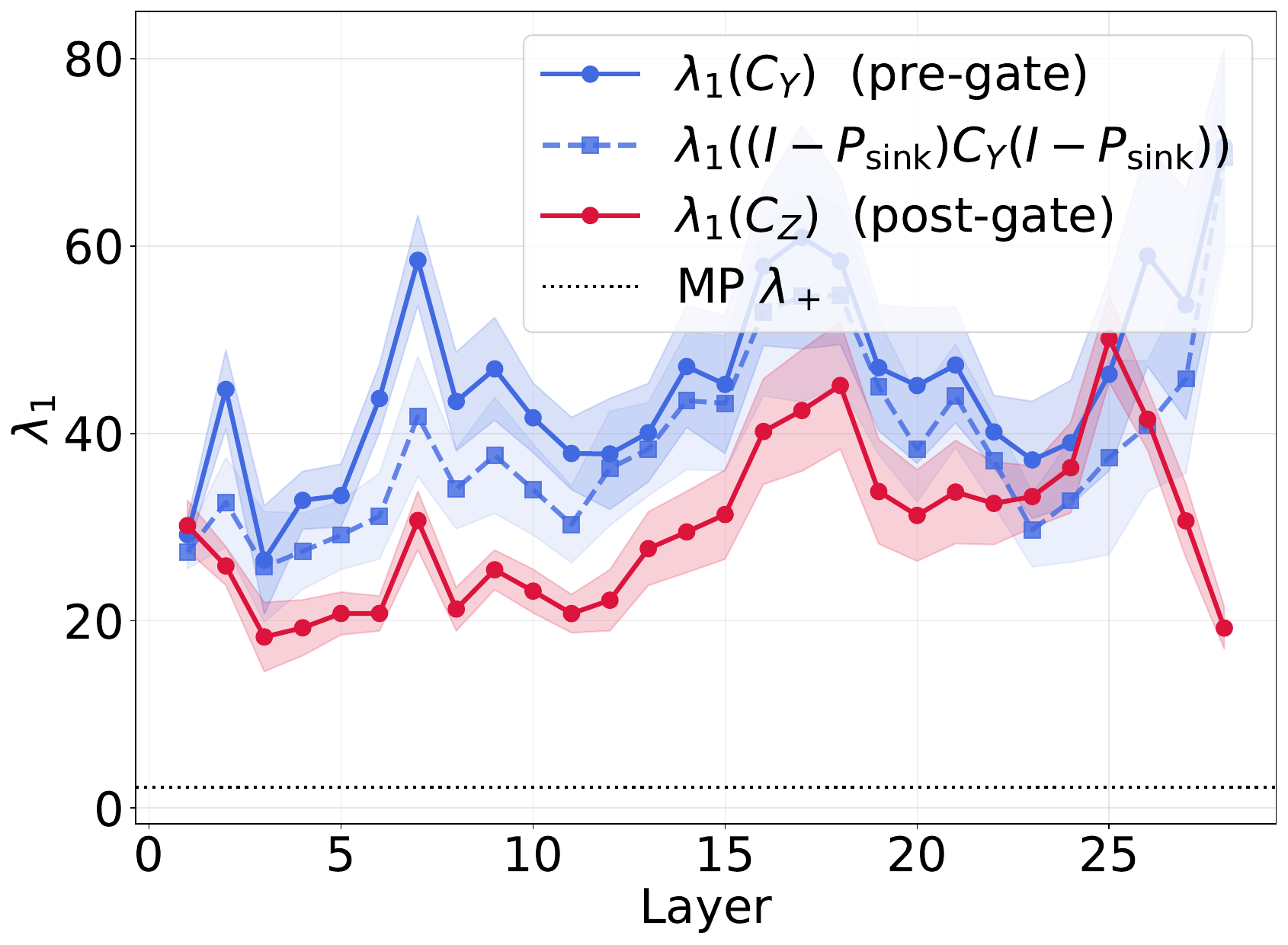}
    \vspace{-1.0em}
    \caption{
    \textbf{Sink-subspace diagnostic.} Removing \(P_{\mathrm{sink}}\) lowers
    \(\lambda_1\).
    }
    \label{fig:sink_subspace_diagnostics}
    \vspace{-1.0em}
\end{wrapfigure}

We next examine whether the spectral suppression observed in Section~\ref{sec:forward_empirical} is attributable to the sink-induced subspace. Using the data-adaptive sink projector \(P_{\mathrm{sink}}\), we compare three quantities per layer: the pre-gate leading eigenvalue \(\lambda_1(C_Y)\), the same quantity after explicitly removing the sink subspace,
\(
\lambda_1\bigl((I-P_{\mathrm{sink}})C_Y(I-P_{\mathrm{sink}})\bigr),
\)
and the post-gate value \(\lambda_1(C_Z)\). 
As shown in Figure~\ref{fig:sink_subspace_diagnostics}, projecting out the sink subspace lowers \(\lambda_1\) toward the Marchenko--Pastur edge \(\lambda_+\) in most layers, indicating that sink-induced directions account for part of the dominant pre-gate spectral component.
The post-gate value \(\lambda_1(C_Z)\) lies below this sink-ablated value in most layers, suggesting that the gate also attenuates leading structure outside the selected sink subspace.

\subsection{Attention--Value Pairing Permutation Control}
\label{app:value_permutation}

The alignment of the dominant token-space direction with a sink subspace could, in principle, arise from a special value state associated with a sink position rather than from the high-mass attention-column geometry. To test this alternative, we preserve the entire attention matrix $A$ and empirical value matrix $V$ while randomly permuting their positionwise pairing:
\begin{equation}
Y_\pi=A\Pi V
=\sum_j A_{:,j}v_{\pi(j)}^\top,
\label{eq:appendix_permutation}
\end{equation}
where $\Pi$ is a permutation matrix. The original sink-subspace projector is held fixed, and the alignment is recomputed using the same feature-standardized diagnostic as in the main spectral analysis. We average over $20$ random permutations for each of the same $16$ sequences used in the original analysis packed with an EOS separator.

\begin{table}[!htbp]
\centering
\caption{\textbf{Sink-subspace alignment under attention--value pairing permutation.} The attention matrix, empirical value matrix, and original sink projector are preserved; only the column--value pairing is randomized. Results are averaged over 20 permutations per sequence and 16 sequences. Alignment is the feature-standardized sink alignment of the pre-gate output, as in the main spectral analysis.}
\label{tab:appendix_value_permutation}
\small
\begin{tabular}{lccc}
\toprule
Representation & Original & Permuted & Layers with increase \\
\midrule
Baseline & 0.062 & 0.196 & $100\%$ \\
Gated (pre-gate) & 0.220 & 0.510 & $96\%$ \\
\bottomrule
\end{tabular}
\end{table}

Table~\ref{tab:appendix_value_permutation} shows that sink alignment increases rather than decreases when the original attention-column--value pairing is permuted. Therefore, the sink-aligned leading direction does not require a privileged value vector attached to the original sink position. 

\subsection{Robustness Beyond the Isotropic-Value Approximation}
\label{app:value_anisotropy}

The isotropic-value approximation is an interpretable reference rather than a literal assumption about trained values. For an attention output $Y=AV$, write
\begin{equation}
B_V=\frac{1}{d}VV^\top
=\hat\sigma_V^2I+\Delta_V.
\label{eq:appendix_value_gram}
\end{equation}
Then the exact raw token-space covariance is
\begin{equation}
C_Y^{\mathrm{raw}}
=\frac{1}{d}YY^\top
=AB_VA^\top
=\hat\sigma_V^2AA^\top+A\Delta_VA^\top.
\label{eq:appendix_anisotropic_decomp}
\end{equation}
The first term isolates the attention-column reference; the second captures learned value anisotropy, which can strengthen, weaken, or rotate the resulting spectral directions.
\begin{table}[!htbp]
\centering
\caption{\textbf{Value anisotropy and sink alignment.} Alignment at the endpoints $\alpha=0$ (isotropic) and $\alpha=1$ (learned values) of the value-Gram interpolation in Qwen-1B, computed on the non-standardized pre-gate covariance $C_\alpha$ and thus not directly comparable to Table~\ref{tab:appendix_value_permutation}. An isotropic Gaussian null gives cross-token correlation $\approx 1$ and normalized feature rank $\approx 0.88$.}
\label{tab:appendix_value_anisotropy}
\resizebox{\linewidth}{!}{%
\begin{tabular}{lccccc}
\toprule
Model & Cross-token corr. & Norm. feature rank & Align. ($\alpha=0$) & Align. ($\alpha=1$) & Random align. \\
\midrule
Baseline & 4.9 & 0.44 & 0.97 & 0.64 & $\approx 0.003$ \\
Gated & 14.4 & 0.33 & 0.79 & 0.32 & $\approx 0.009$ \\
\bottomrule
\end{tabular}%
}
\end{table}

On the Qwen-1B checkpoints and sequences used for the original spectral analysis, learned values are strongly anisotropic. Table~\ref{tab:appendix_value_anisotropy} reports the dimension-normalized cross-token correlations and normalized feature effective ranks, alongside a matched isotropic Gaussian reference. These diagnostics show substantial learned-value anisotropy in both checkpoints.
To assess sensitivity to this anisotropy while holding $A$ fixed, we interpolate the value Gram matrix in the positive semidefinite cone:
\begin{equation}
B_\alpha=(1-\alpha)\hat\sigma_V^2I+\alpha B_V,
\qquad C_\alpha=AB_\alpha A^\top,
\quad\alpha\in[0,1].
\label{eq:appendix_value_interpolation}
\end{equation}
The endpoint $\alpha=0$ is the isotropic attention-column reference, whereas $\alpha=1$ uses the empirical learned-value Gram matrix. 
As Table~\ref{tab:appendix_value_anisotropy} shows, moving from the isotropic reference to the empirical value geometry reduces alignment in both models, but the empirical endpoint remains far above the corresponding random-subspace reference in every evaluated layer and sample. The isotropic approximation is therefore useful as a structural prediction for sink-aligned spectral modes. Importantly, the deterministic gate contraction theorem does not rely on value isotropy; isotropy enters only the interpretable $AA^\top$ reference and the idealized BBP analysis.
\clearpage
\subsection{Naturally Occurring Non-First-Position Sinks}
\label{app:nonfirst_sinks}

\begin{wraptable}{r}{0.55\textwidth}
\centering
\vspace{-1.5em}
\caption{\textbf{Non-first-position sink structure.} The first position is excluded from the sink subspace. Alignment is compared before and after gating, with pre-gate alignment normalized by the random-subspace baseline. Projection measures the reduction in the leading covariance eigenvalue after removing the selected sink subspace.}
\label{tab:appendix_nonfirst}
\vspace{.5em}
\small
\resizebox{\linewidth}{!}{%
\begin{tabular}{lcc}
\toprule
Metric & Qwen-1B & Pythia-410M \\
\midrule
Align. (pre $\to$ post) & $0.280\to0.186$ & $0.262\to0.154$ \\
Pre/random & $21\times$ & $20\times$ \\
Align. drop & $33.4\%$ & $41.2\%$ \\
$\lambda_1$ drop (proj.) & $19.4\%$ & $19.2\%$ \\
\bottomrule
\end{tabular}%
}
\vspace{-.1em}
\end{wraptable}

We test whether the sink-subspace spectral signature persists beyond the first position and beyond a literal rank-one object. The general higher-rank contraction argument is discussed in Appendix~\ref{app:higher_rank_sinks}.
We exclude position $0$ when selecting sink columns in normally trained checkpoints and evaluate their contribution to the dominant token-space covariance direction. Table~\ref{tab:appendix_nonfirst} reports the observed alignment and the effect of projecting out the selected sink subspace.
Across the two model families, the selected sink sets contain approximately $6.7$ non-first columns on average. Their leading-eigenvector alignment is roughly $20\times$ the random-subspace reference, and removing these subspaces reduces $\lambda_1$ by about $19\%$. Approximately $43$--$45\%$ of the selected non-first column instances correspond to structural delimiters, including periods, commas, and newlines. Within the gated models, post-SDPA gating reduces the corresponding alignment by $33$--$41\%$. These observations show that the forward spectral signature extends to multi-column, delimiter-rich sink structure.

\section{Additional Training-Time Diagnostics} \label{app:training_diagnostics}
\subsection{Training-Time Sink and Gradient Diagnostics}
\label{app:existing_training_diagnostics}

\begin{figure}[t]
    \centering
    \subfigure[Baseline Sink mass]{
        \includegraphics[width=0.23\linewidth]{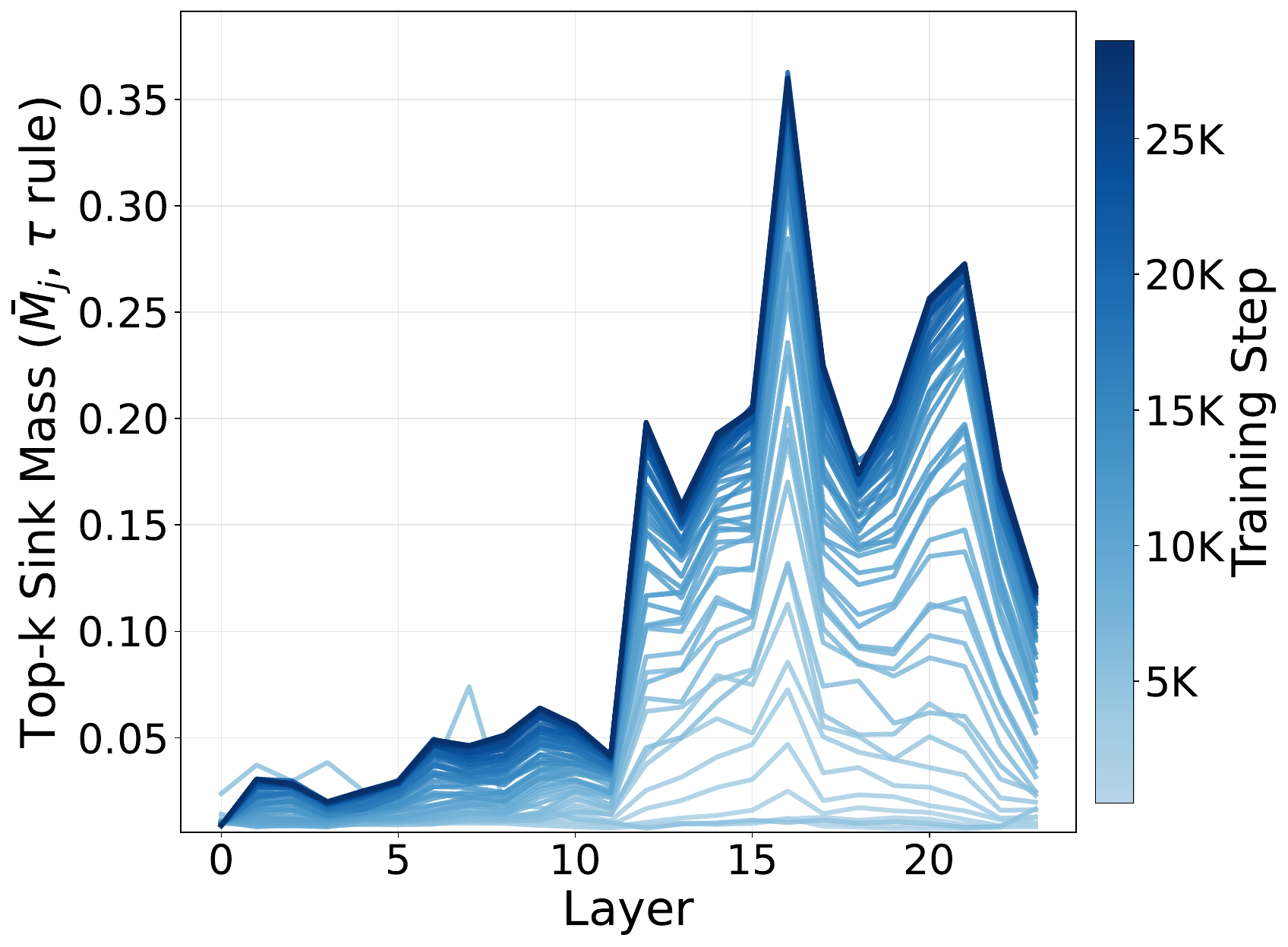}}
    \subfigure[Gated Sink mass]{
        \includegraphics[width=0.23\linewidth]{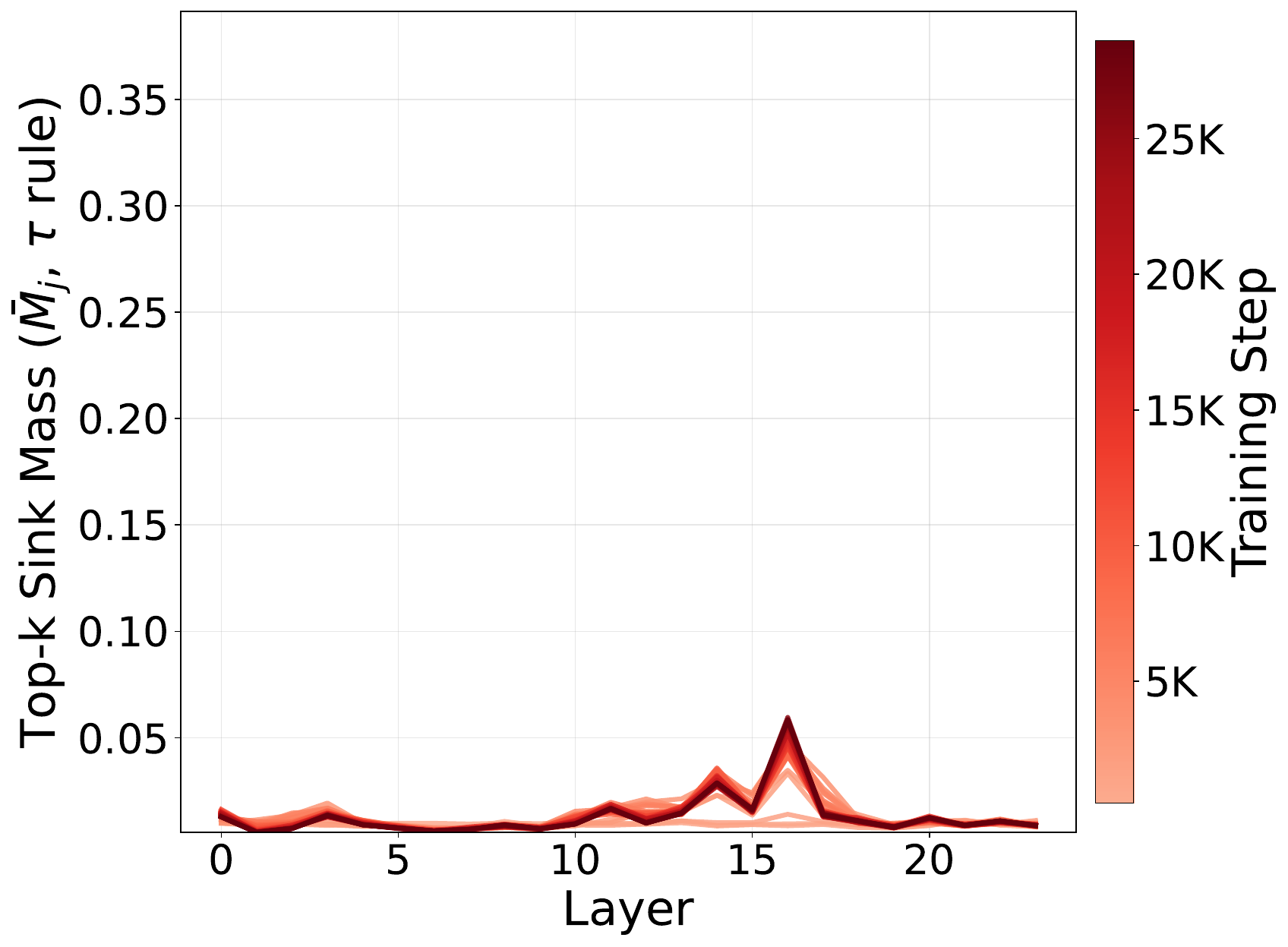}}
    \subfigure[Baseline Outlier mass]{
        \includegraphics[width=0.23\linewidth]{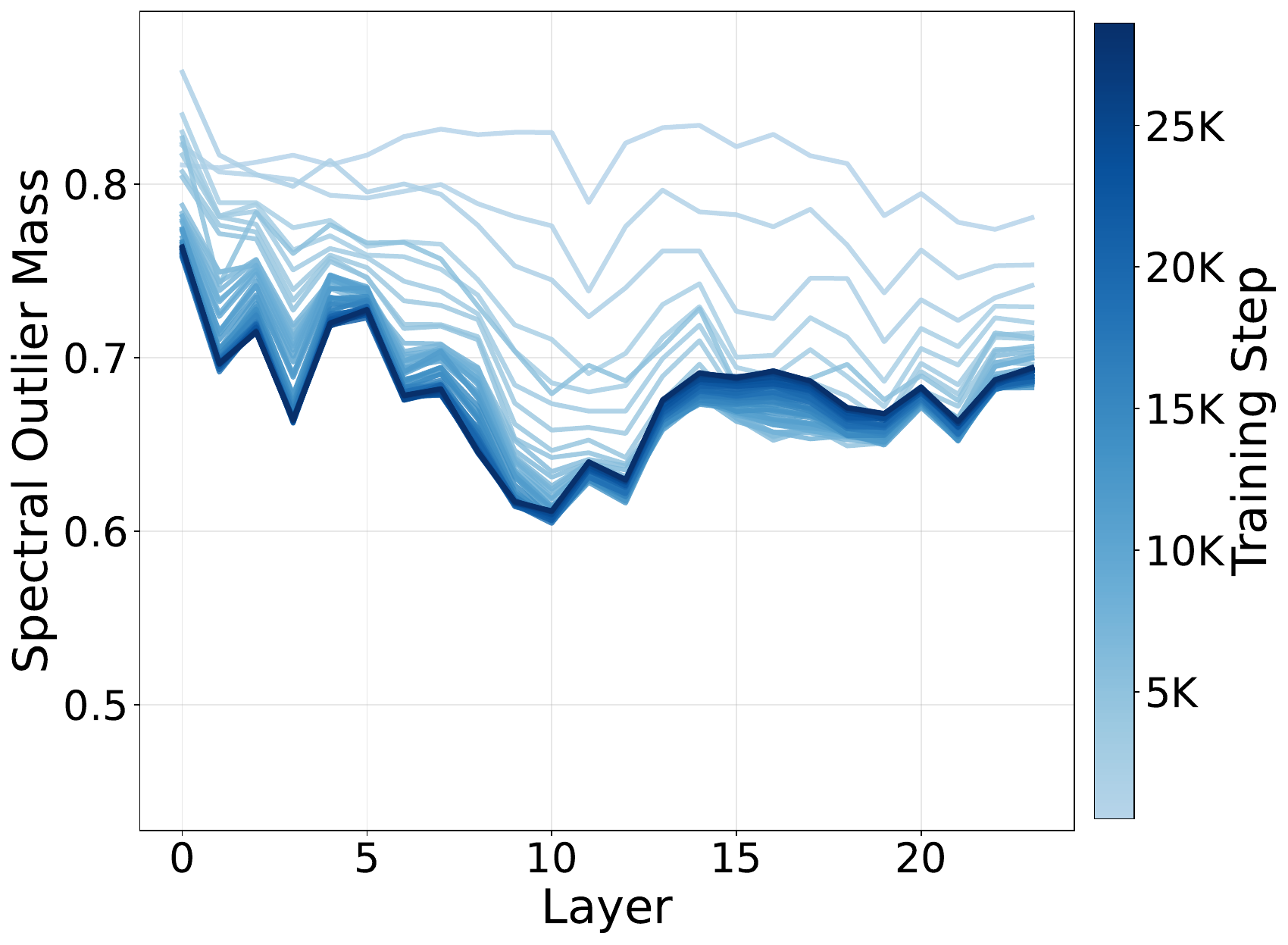}}
    \subfigure[Gated Outlier mass]{
        \includegraphics[width=0.23\linewidth]{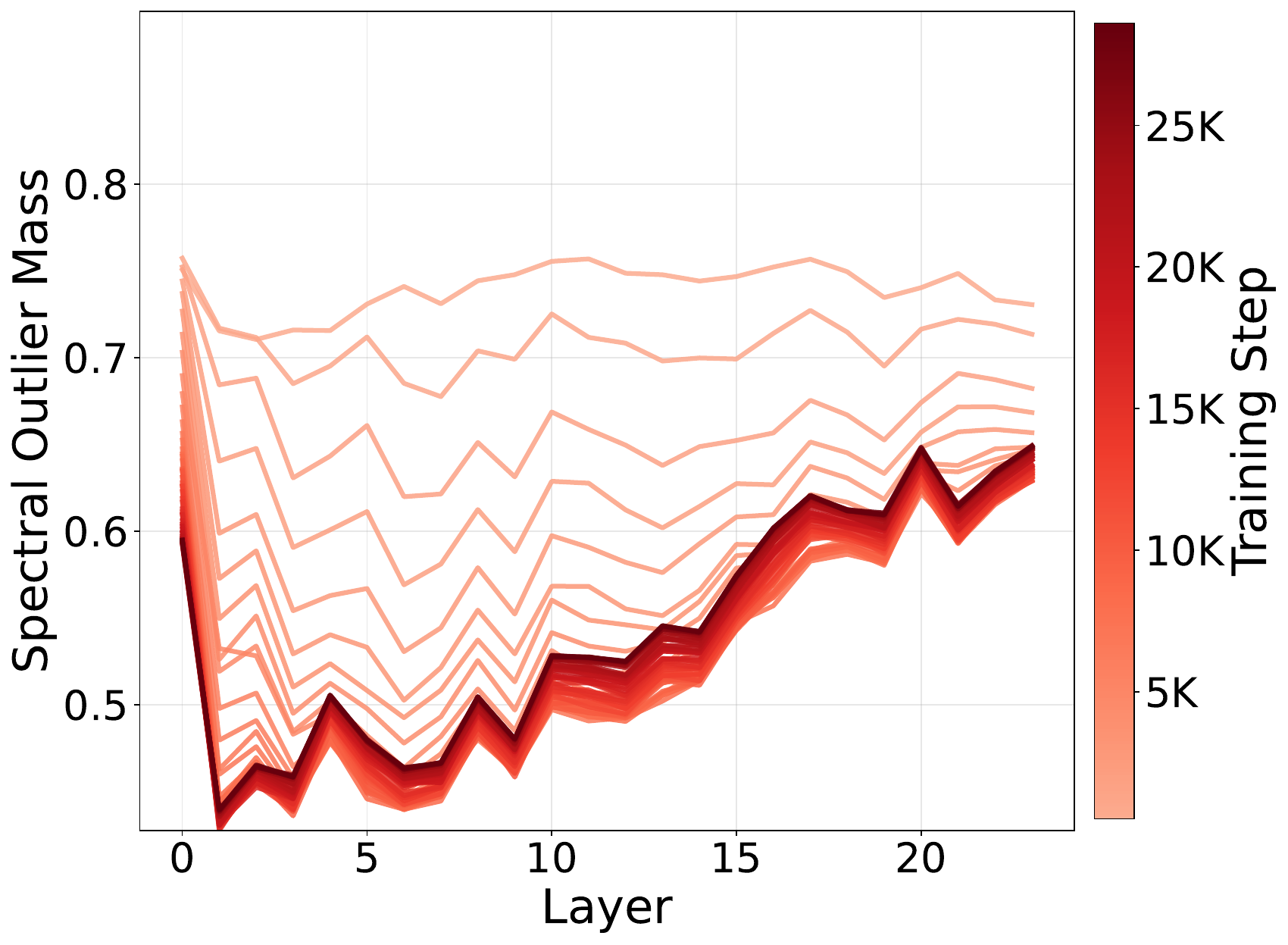}}
    \caption{
    \textbf{Training-time sink and spectral diagnostics.} 
    Each curve corresponds to a training checkpoint. 
    (a,b) Top-\(k\) sink mass is the mean \(\bar M_j\) over the sink columns selected as in Section~\ref{sec:forward_empirical}, while (c,d) outlier mass measures the fraction of token-space covariance spectrum above the MP edge. 
    Compared with the baseline, post-SDPA gating keeps both high-mass sink columns and out-of-bulk spectral concentration substantially weaker across training.
    }
    \label{fig:training_sink_spectral}
\end{figure}

\begin{table}[t]
\centering
\caption{
\textbf{Downstream evaluation of Pythia-410M.} Baseline and post-SDPA gated variants trained on 15B FineWeb-Edu tokens. All variants use matched data and optimization settings, with gated models adjusting MLP size to compensate for gate parameters. The best results are highlighted in \textbf{bold}. Avg. denotes the average of accuracies and normalized accuracies over 8 tasks.}
\label{tab:eval_results}
\resizebox{\textwidth}{!}{
\begin{tabular}{lccccccccccc}
\toprule
\multirow{2}{*}{Model} & Wiki. & LMB. & LMB. & PIQA & Hella & Wino & ARC-e & ARC-c & OBQA & BoolQ & \multirow{2}{*}{Avg.$\uparrow$} \\
      & ppl$\downarrow$ & ppl$\downarrow$ & acc$\uparrow$ & acc$\uparrow$ & acc\_n$\uparrow$ & acc$\uparrow$ & acc$\uparrow$ & acc\_n$\uparrow$ & acc\_n$\uparrow$ & acc$\uparrow$ &  \\
\midrule
Baseline          & 29.53 & 55.21 & 29.94 & 65.83 & 36.76 & 51.70 & 56.02 & 26.54 & 30.80 & \textbf{62.11} & 44.96 \\
Gated       & \textbf{27.99} & \textbf{40.52} & \textbf{32.47} & \textbf{66.43} & \textbf{37.93} & \textbf{52.49} & \textbf{57.79} & \textbf{27.65} & \textbf{32.00} & 61.71 & \textbf{46.06} \\
\bottomrule
\end{tabular}
}
\end{table}

\begin{table}[t]
\centering
\caption{
\textbf{Final-checkpoint diagnostics.} Layer-mean values for Pythia-410M at the final checkpoint. Post-SDPA gating reduces attention concentration, weakens dominant spectral structure, reduces out-of-bulk mass, increases effective rank, and substantially lowers first-token value-gradient amplification. Spectral metrics for the gated model are computed on the post-gate output.
}
\label{tab:final_checkpoint_summary}
\begin{tabular}{lcccccc}
\toprule
Variant & \(m_1\) & \(m_{\mathrm{top}k}\) & \(\lambda_1/\lambda_+\) & OutMass & \(r_{\mathrm{eff}}\) & \(\rho_1\) \\
\midrule
Baseline & 0.084 & 0.125 & 18.42 & 0.679 & 89.14 & 25.40 \\
Gated    & 0.013 & 0.013 & 12.49 & 0.545 & 153.20 & 5.67 \\
\bottomrule
\end{tabular}
\end{table}

We provide additional training-time diagnostics that complement the final-checkpoint analyses in the main text. These diagnostics track how attention concentration, token-space spectral structure, and value-gradient amplification evolve across checkpoints in the baseline and post-SDPA gated models. To verify that the gated variant remains competitive under the matched 15B-token training setup, we also report standard language-modeling and downstream evaluation results in Table~\ref{tab:eval_results}. 
Unless otherwise stated, all diagnostics are evaluated on sixteen 512-token WikiText-103 validation sequences, with an EOS token between passages. Table~\ref{tab:final_checkpoint_summary} reports their layer means at the final checkpoint, and Figure~\ref{fig:training_sink_spectral} shows them layer-wise across checkpoints.

Table~\ref{tab:final_checkpoint_summary} summarizes the same trend observed in the layer-wise diagnostics. Post-SDPA gating reduces first-token and top-\(k\) sink mass, lowers the leading-eigenvalue ratio, reduces outlier mass, increases effective rank, and substantially reduces first-token value-gradient amplification. Thus, the gated model shows weaker attention concentration, weaker dominant spectral structure, and weaker sink-directed gradient amplification at the final checkpoint.

Figure~\ref{fig:training_sink_spectral} tracks the emergence of sink and spectral signatures during training. In the baseline, top-\(k\) sink mass grows in middle-to-late layers, indicating the formation of high-mass attention columns. The same layers also show stronger out-of-bulk spectral mass, consistent with the sink-induced spectral-spike view. In contrast, the post-SDPA gated model keeps both quantities lower across checkpoints, suggesting that gating suppresses the co-emergence of attention concentration and token-space spectral outliers.

\subsection{Direct Measurements of Massive Activations and Their Relation to Sinks}
\label{app:massive_activations}

To measure massive activations directly, rather than relying solely on relative first-token dominance, we record residual-stream hidden states $h_t$ after each Transformer block. Let $\mathcal{V}$ denote the set of non-padding token positions in each sample. We measure the following massive-activation (MA) diagnostics:
\begin{align}
M_2 &= \max_{t\in\mathcal{V}}\|h_t\|_2,
&&\text{(maximum token hidden-state norm)},\\
M_\infty &= \max_{t\in\mathcal{V},\,c}|h_{t,c}|,
&&\text{(maximum coordinate magnitude)}.
\label{eq:appendix_activation_diagnostics}
\end{align}
For Pythia-410M and Qwen-1B, we evaluate both MA diagnostics on the same sixteen 512-token WikiText-103 validation blocks without padding using model-specific tokenization.  
Each metric is computed per sample and layer, then averaged across samples and layers for Table~\ref{tab:appendix_massive_activations}. 

\begin{table}[t]
\centering
\small
\caption{\textbf{Massive-activation (MA) diagnostics.} Entries are sample- and layer-mean values on sixteen 512-token WikiText-103 validation blocks without padding. The last column reports baseline-to-gated ratios.}
\label{tab:appendix_massive_activations}
\begin{tabular}{llrrr}
\toprule
Model & Diagnostic & Baseline & Gated & Baseline/gated \\
\midrule
Pythia-410M 
           & $M_2$ (MA) & 379.2 & 108.4 & $3.50\times$ \\
           & $M_\infty$ (MA) & 320.0 & 67.4 & $4.75\times$ \\
\midrule
Qwen-1B    & $M_2$ (MA) & 11,854.5 & 146.3 & $81.1\times$ \\
           & $M_\infty$ (MA) & 11,839.5 & 87.2 & $135.7\times$ \\
\bottomrule
\end{tabular}
\end{table}

In Pythia-410M, post-SDPA gating jointly suppresses AS, GS, and MA (Tables~\ref{tab:final_checkpoint_summary} and~\ref{tab:appendix_massive_activations}). Their baseline peaks occur in middle-to-late layers (17--21), while the layerwise maxima of $M_2$ and $M_\infty$ decrease from $726.7$ to $186.6$ and from $643.0$ to $157.4$, respectively. Across 56 training checkpoints, the baseline's layer-mean $M_\infty$ grows from $8.7$ to $320.0$, whereas the gated model never exceeds $69.4$.
Qwen-1B further confirms cross-model MA suppression, with both $M_2$ and $M_\infty$ reduced by over an order of magnitude (Table~\ref{tab:appendix_massive_activations}). The maximum observed coordinate magnitude is $17{,}536$ in the baseline (first token, layer 8), compared with $378$ in the gated model across the evaluated samples and layers.
Together, these measurements show lower attention concentration, value-gradient amplification, and massive-activation magnitudes in the gated Pythia-410M model, while the Qwen-1B results confirm massive-activation suppression. Combined with the fixed-upstream-gradient counterfactual in Section~\ref{sec:backward_empirical}, these findings are consistent with attenuation along the AS--GS--MA pathway. 

\subsection{Additional Next-Token Probing Diagnostic} \label{app:probing}

We additionally examine whether the sink-related changes observed under gating are accompanied by changes in linearly accessible next-token information. Conceptually, this asks whether target-related information remains recoverable from hidden states across layers, especially around layers where sink signatures are pronounced.

\paragraph{Practical proxy.}
Since the conditional uncertainty of the next token given a hidden state is difficult to estimate directly, we use layer-wise linear probing as a practical proxy for next-token decodability. For each layer \(\ell\), we train a linear probe to predict the next token from the hidden state \(H_\ell\), and report cross-entropy loss on the test set. Lower cross-entropy indicates that next-token information is more linearly recoverable from the hidden representation.

\paragraph{Experimental setting.}
We use the Qwen-1B models with and without post-SDPA gating and the WikiText-2 dataset with maximum sequence length \(128\). For each model and layer, we train a linear probe and evaluate it on the test split, where layer 0 denotes the embedding output. Each probe is trained on raw or per-dimension standardized hidden states, either for one epoch with learning rate \(10^{-3}\) or for five epochs with learning rate \(10^{-4}\). We report results both with and without the initial source position to check whether the observed trend is driven solely by the canonical sink position.

\begin{figure}[t]
    \centering
    \subfigure[Raw, one epoch]{
        \includegraphics[width=0.32\linewidth]{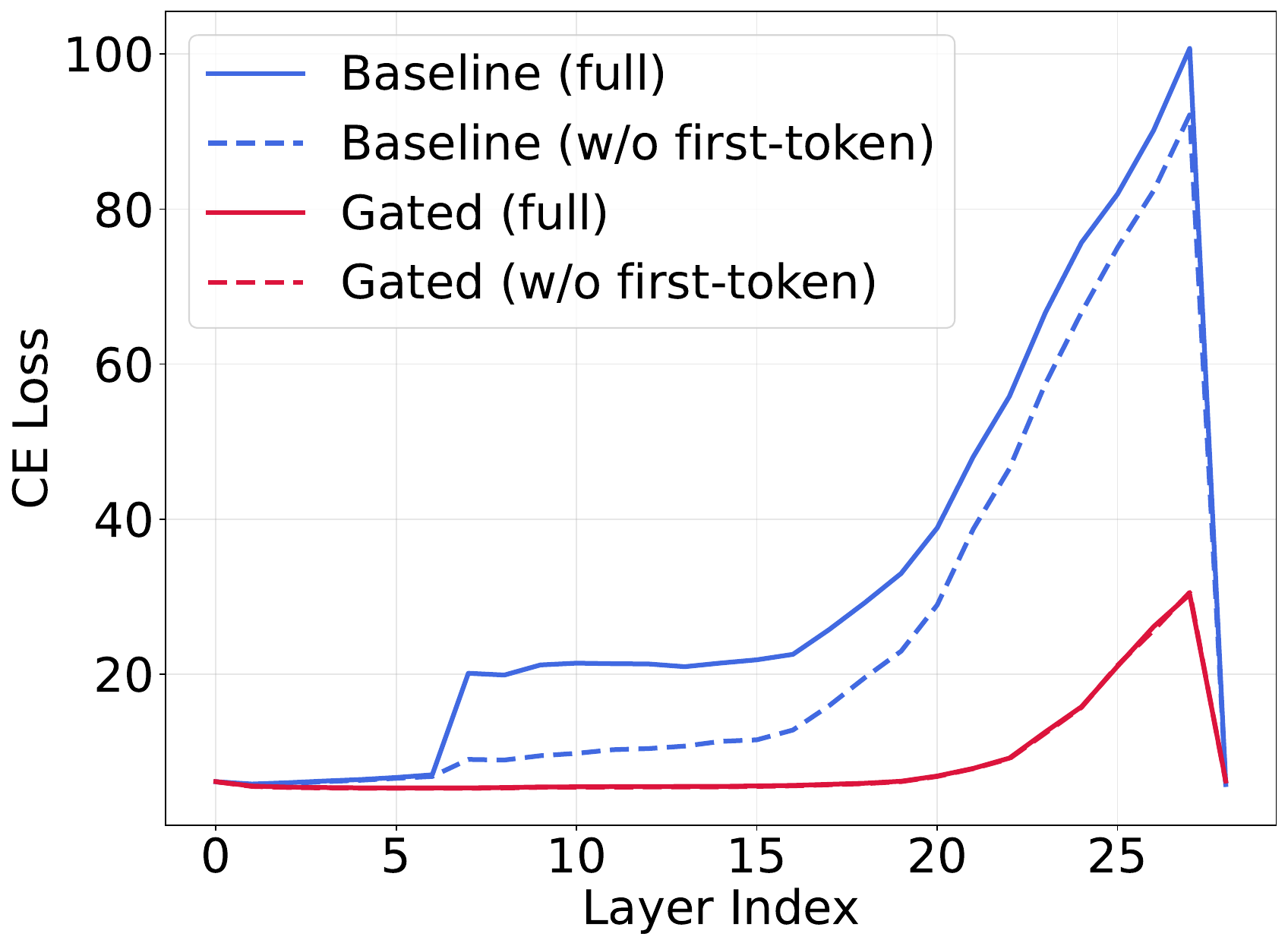}}
    \hfill
    \subfigure[Standardized, one epoch]{
        \includegraphics[width=0.32\linewidth]{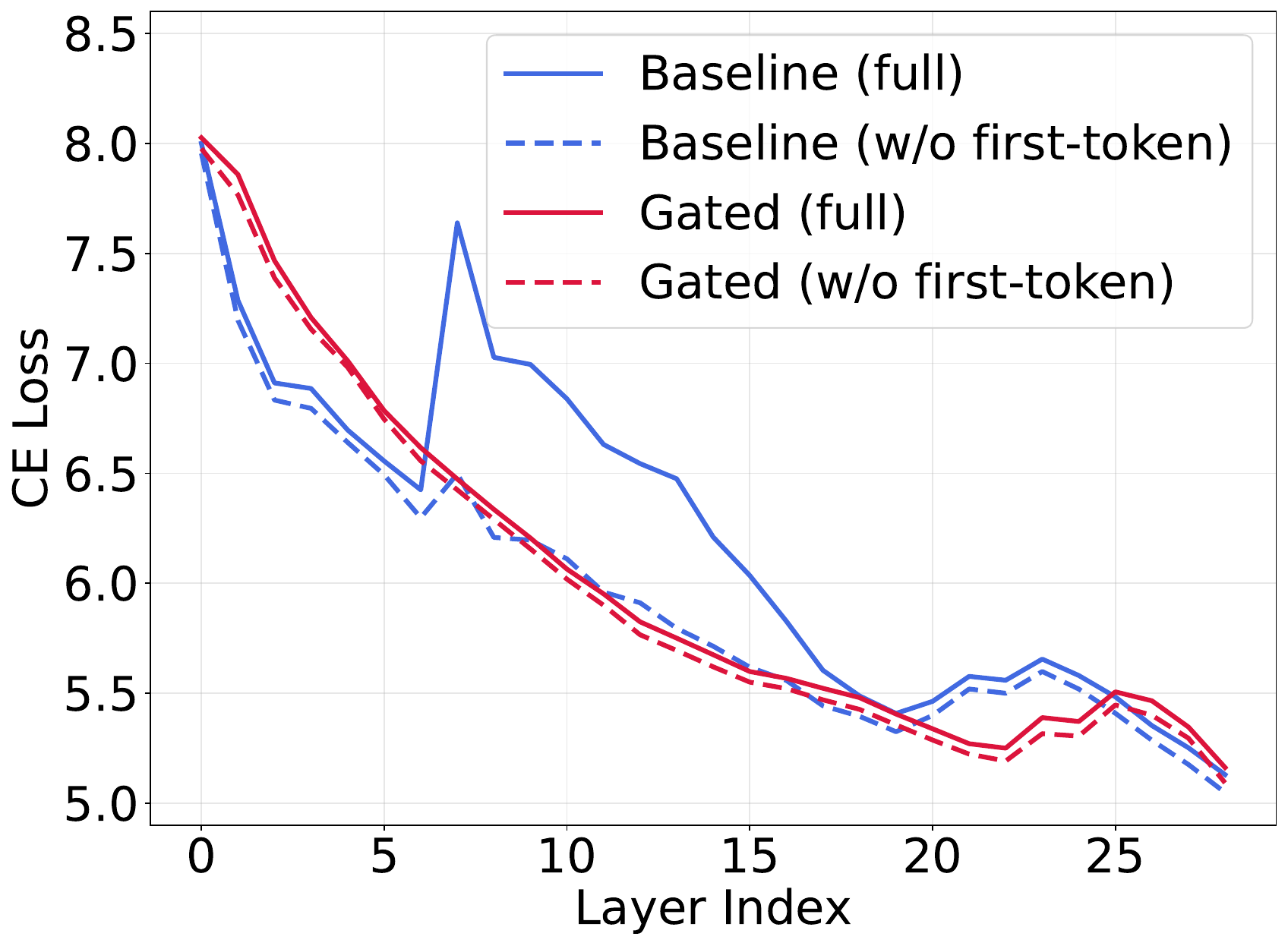}}
    \hfill
    \subfigure[Standardized, five epochs]{
        \includegraphics[width=0.32\linewidth]{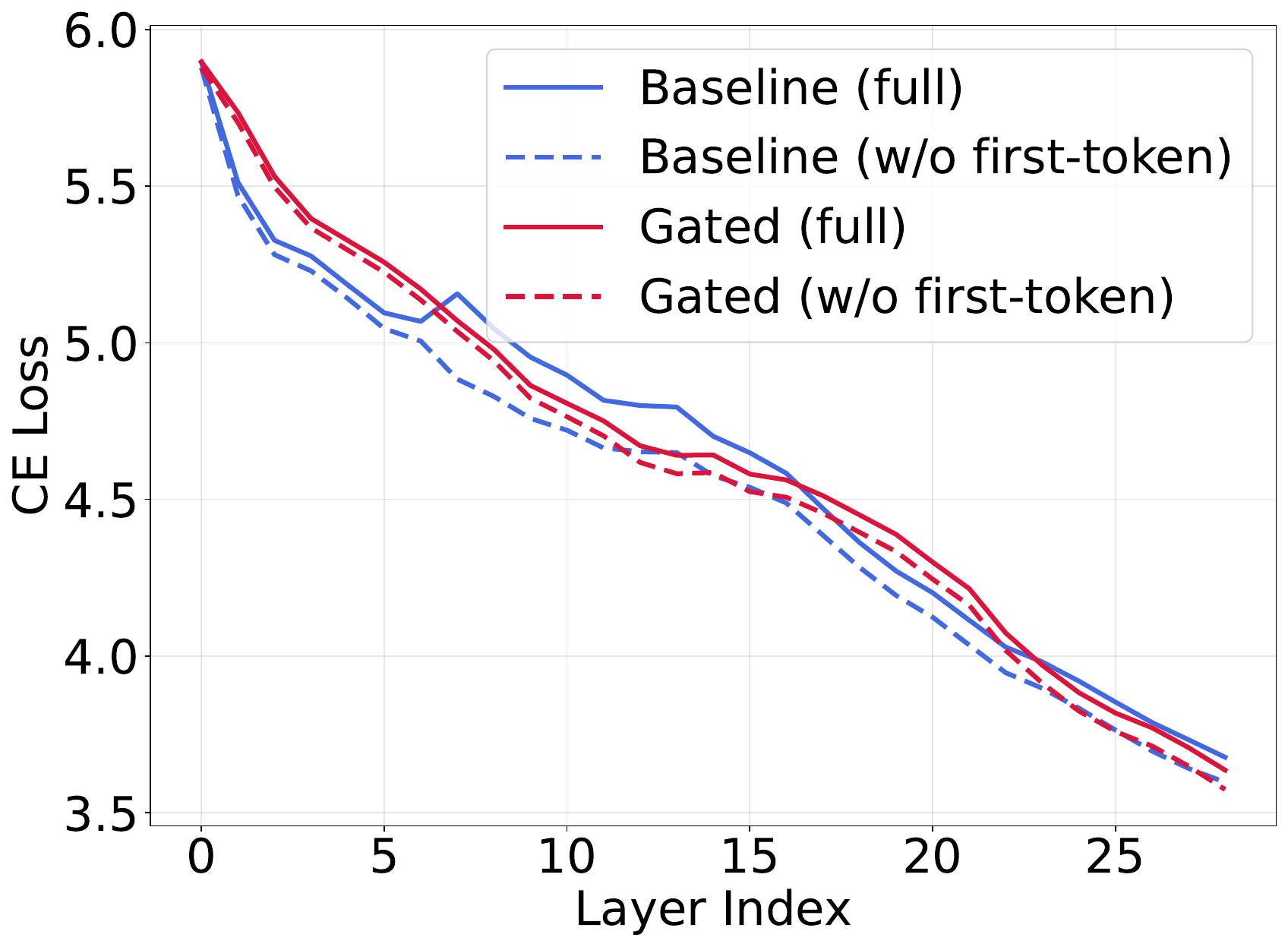}}
    \caption{\textbf{Layer-wise probing cross-entropy loss.} Baseline and gated models are evaluated with and without the initial source position. Probes are trained for one epoch on (a) raw and (b) per-dimension standardized hidden states, and (c) for up to five epochs with a lower learning rate on standardized hidden states. Lower is better.}
    \label{fig:probing}
\end{figure}

\paragraph{Experimental results.}
Under the original protocol (Figure~\ref{fig:probing}(a)), baseline probes are much worse than gated probes in layers 8--28, where massive activations emerge in the baseline (Appendix~\ref{app:massive_activations}). Massive activations place a few hidden dimensions at a far larger scale than the rest, which makes a linear probe difficult to optimize under a fixed training budget; in these layers, the baseline probes are even worse than a uniform prediction over the vocabulary. Gated probes, whose hidden states have much smaller activations, are far less affected.
To test whether this gap reflects missing next-token information or only this optimization difficulty, we standardize each hidden dimension before probing, which leaves the class of linear probes unchanged. Under the same training budget, the large gap in deep layers mostly disappears (Figure~\ref{fig:probing}(b)). A smaller gap remains in layers 8--17, where the first-token massive activation appears, and it vanishes once the initial source position is excluded. With a lower learning rate and longer training, this remaining gap also largely closes (Figure~\ref{fig:probing}(c)), and the two models reach comparable probing loss across depth, with or without the initial source position.
These results indicate that the gap under the original protocol arises from the difficulty of training linear probes on hidden states with massive activations, rather than from a lack of next-token information. Post-SDPA gating suppresses massive activations and thereby makes next-token information easier to read out with a linear probe, while both models retain comparable linearly recoverable next-token information.

\section{Robustness Across Models and Settings}
\label{app:additional_robustness}

We examine whether the attention, spectral, and value-gradient signatures extend beyond the original model scale, context length, and sequence-formatting convention. The following analyses compare matched baseline and post-SDPA elementwise-gated models. These experiments use separately constructed WikiText-103 test inputs, so absolute values differ slightly from Table~\ref{tab:final_checkpoint_summary}.

\subsection{Robustness Across Model Scale}
\label{app:scale_28b}

\begin{table}[t]
\centering
\caption{\textbf{Model-scale robustness.} The first two rows are absolute layer-mean values at $2.8$B; the last two rows report gated/baseline ratios. Ratios below $1$ indicate decreases under gating.}
\label{tab:appendix_scale}
\resizebox{\linewidth}{!}{%
\begin{tabular}{lrrrrrrrr}
\toprule
Model & $m_1$ & $m_{\mathrm{top}k}$ & $\rho_1$ & $\rho_{\mathrm{top}k}$ & OutMass & $\lambda_1/\lambda_+$ & Eff. rank & Sink alignment \\
\midrule
Baseline 2.8B & 0.147 & 0.156 & 28.24 & 27.04 & 0.684 & 24.21 & 111.2 & 0.174 \\
Gated 2.8B & 0.011 & 0.010 & 4.29 & 3.61 & 0.566 & 15.54 & 185.0 & 0.156 \\
2.8B ratio & 0.077 & 0.067 & 0.152 & 0.133 & 0.827 & 0.642 & 1.664 & 0.895 \\
410M ratio & 0.129 & 0.101 & 0.195 & 0.160 & 0.796 & 0.658 & 1.717 & 0.657 \\
\bottomrule
\end{tabular}%
}
\end{table}

We trained matched ungated and post-SDPA elementwise-gated Pythia-2.8B models and repeated the attention, value-gradient, and token-space spectral diagnostics. Table~\ref{tab:appendix_scale} reports absolute layer means for the $2.8$B models and gated-to-baseline ratios for both $2.8$B and the corresponding $410$M comparison.
At $2.8$B, gating reduces first-token attention mass by approximately $13\times$, top-$k$ attention mass by approximately $15\times$, and first-token and top-$k$ value-gradient amplification by $6.6\times$ and $7.5\times$, respectively. The leading-outlier ratio decreases from $24.21$ to $15.54$, while effective rank increases from $111.2$ to $185.0$. All diagnostics change in the same qualitative direction at $2.8$B and $410$M. The attention- and gradient-related reductions are at least as strong at $2.8$B, and the leading-outlier and effective-rank ratios are similar at the two scales. 
Thus, the attenuation of sink-related attention, gradient, and spectral signatures by post-SDPA gating is not specific to the 410M scale.

\subsection{Robustness to Longer Training and Evaluation Contexts}
\label{app:long_context}

\begin{table}[t]
\centering
\caption{\textbf{Robustness across training and evaluation context lengths.} Baseline and elementwise-gated models are compared under matched context-length settings. Top-$k$ value-gradient amplification is omitted for the 8K/8K setting because the backward pass exceeded available device memory.}
\label{tab:appendix_context}
\resizebox{\linewidth}{!}{%
\begin{tabular}{llccccccc}
\toprule
Train/eval & Model & $m_1$ & $m_{\mathrm{top}k}$ & Sink columns & $\lambda_1/\lambda_+$ & Norm. eff. rank & Alignment/random & $\rho_{\mathrm{top}k}$ \\
\midrule
\multirow{2}{*}{4K/4K}
 & Baseline & 0.025 & 0.071 & 2.5 & 44.1 & 0.159 & 180.4 & 58.4 \\
 & Gated    & 0.002 & 0.003 & 6.5 & 29.1 & 0.288 & 58.5 & 6.2 \\
\midrule
\multirow{2}{*}{8K/4K}
 & Baseline & 0.015 & 0.058 & 3.3 & 53.0 & 0.118 & 203.8 & 54.7 \\
 & Gated    & 0.002 & 0.003 & 5.7 & 35.4 & 0.237 & 50.7 & 7.5 \\
\midrule
\multirow{2}{*}{8K/8K}
 & Baseline & 0.008 & 0.058 & 3.1 & 59.0 & 0.134 & 349.9 & --- \\
 & Gated    & 0.001 & 0.002 & 5.8 & 42.2 & 0.262 & 68.7 & --- \\
\bottomrule
\end{tabular}%
}
\end{table}

To test context-length robustness beyond the original $4$K setting, we trained matched baseline and elementwise-gated Pythia-410M models with an $8$K context window. We evaluate the $8$K-trained checkpoints at both $4$K and $8$K, and compare them to $4$K-trained checkpoints evaluated at $4$K. All use the original plain sequence convention, in which the first token is an ordinary content token. Table~\ref{tab:appendix_context} reports baseline-to-gated values. Effective rank and sink alignment are normalized by \(\min(T,D)\) and \(k/T\), respectively, for comparison across sequence lengths.

At both evaluation lengths, gating reduces the first-position and adaptive top-$k$ sink masses, the normalized leading spectral outlier, and sink-subspace alignment, while increasing normalized effective rank. Within the $8$K-trained baseline, extending evaluation from $4$K to $8$K reduces first-position mass from $0.015$ to $0.008$, even though $m_{\mathrm{top}k}$ stays at $0.058$. This suggests that the longer-context sink pattern is less exclusively concentrated at position $0$ and is better captured by adaptive, potentially non-first sink columns. The $8$K-trained model also exhibits strong value-gradient amplification at $4$K evaluation, reduced by gating from $54.7$ to $7.5$. We do not report an $8$K/$8$K backward measurement because retaining the attention states for backpropagation scales quadratically with sequence length and exceeded device memory. Consequently, forward attention and spectral robustness are established at $8$K evaluation, while the gradient result for the $8$K-trained checkpoint is directly verified at $4$K evaluation.

\subsection{Robustness to Sequence Formatting and Special Tokens}
\label{app:special_tokens}
\paragraph{Sequence formatting conventions.}

\begin{table}[t]
\centering
\caption{\textbf{Robustness across natural sequence-formatting conventions.} All entries are gated/baseline ratios.}
\label{tab:appendix_conventions}
\resizebox{\linewidth}{!}{%
\begin{tabular}{lrrrrrrrr}
\toprule
Convention & $m_1$ & $m_{\mathrm{top}k}$ & $\rho_1$ & $\rho_{\mathrm{top}k}$ & OutMass & $\lambda_1/\lambda_+$ & Eff. rank & Sink alignment \\
\midrule
Plain & 0.135 & 0.108 & 0.206 & 0.175 & 0.797 & 0.649 & 1.723 & 0.660 \\
EOS-separated & 0.183 & 0.226 & 0.253 & 0.320 & 0.786 & 0.644 & 1.735 & 0.385 \\
\bottomrule
\end{tabular}%
}
\end{table}
We additionally trained matched baseline and post-SDPA gated Pythia-410M models under two sequence conventions: (i) \emph{plain}, with \texttt{add\_bos\_token = False} and \texttt{add\_eos\_token = False}; and (ii) \emph{EOS-separated}, with EOS appended at document boundaries (\texttt{add\_eos\_token = True}, \texttt{add\_bos\_token = False}). Table~\ref{tab:appendix_conventions} shows gated/baseline ratios, not absolute metric values. Values below $1$ indicate attenuation, except that an effective-rank ratio greater than $1$ indicates increased effective rank.
Across both conventions, gating decreases attention-sink mass, value-gradient amplification, outlier mass, the leading-outlier ratio, and sink alignment, while increasing effective rank. The direct comparison before and after the gate within each gated model also reproduces the main spectral pattern: $\lambda_1/\lambda_+$ falls to $0.608\times$ under both conventions, effective rank increases by $1.81$--$1.89\times$, and sink alignment decreases to $0.51$--$0.53\times$.

\paragraph{Controlled EOS-separator sinks.} We also train matched baseline and elementwise-gated Pythia-410M models with EOS tokens inserted at document boundaries, evaluating each model under the same convention in Table~\ref{tab:appendix_conventions}. Position $0$ is excluded when constructing the sink subspace. The selected non-first sink sets contain an average of $2.64$ columns in the baseline and $3.54$ in the gated model, with EOS tokens accounting for $58\%$ and $71\%$ of the selected columns, respectively. Their pre-gate or ungated alignments with the leading covariance direction are $0.184$ (baseline) and $0.190$ (gated, before the gate), corresponding to $28$--$36\times$ their random-subspace references. Projecting out these sink subspaces reduces $\lambda_1$ by $12$--$14\%$. Within the gated model, the alignment falls from $0.190$ to $0.099$ after gating, a factor of $0.52$. Repeating the analysis with the column-energy selector $\|A_{:,j}\|_2^2$ yields a reduction from $0.182$ to $0.069$. Thus, learned EOS-related sinks can occupy multiple non-first columns and remain attenuated by gating.

\subsection{Position-Controlled EOS Counterfactual}
\label{app:eos_position_control}

\begin{table}[t]
\centering
\caption{\textbf{Position-controlled EOS counterfactual.} Baseline and post-SDPA gated models are compared across EOS insertion positions. Without EOS, the target is position $0$; otherwise, the target is the inserted EOS token. The strongest sink is selected among positions with at least $T/8$ future queries. Spectral alignment for the gated model uses the post-gate output.}
\label{tab:appendix_eos_position}
\resizebox{\linewidth}{!}{%
\begin{tabular}{llcccc}
\toprule
Input condition & Model
& \shortstack{$P(\text{target strongest})$}
& \shortstack{Target attn.\\mass}
& \shortstack{Target spectral\\alignment}
& \shortstack{Target gradient\\amplification} \\
\midrule
\multirow{2}{*}{No EOS; position $0$}
 & Baseline & 0.583 & 0.130 & 0.204 & 31.39 \\
 & Gated    & 0.174 & 0.015 & 0.031 & 6.38 \\
\midrule
\multirow{2}{*}{EOS at $0$}
 & Baseline & 0.930 & 0.200 & 0.229 & 40.97 \\
 & Gated    & 0.706 & 0.037 & 0.056 & 10.77 \\
\midrule
\multirow{2}{*}{EOS at $0.25T$}
 & Baseline & 0.878 & 0.181 & 0.042 & 28.64 \\
 & Gated    & 0.753 & 0.041 & 0.023 & 10.20 \\
\midrule
\multirow{2}{*}{EOS at $0.50T$}
 & Baseline & 0.911 & 0.199 & 0.131 & 25.30 \\
 & Gated    & 0.885 & 0.055 & 0.031 & 10.03 \\
\midrule
\multirow{2}{*}{EOS at $0.75T$}
 & Baseline & 0.953 & 0.237 & 0.140 & 20.88 \\
 & Gated    & 0.990 & 0.086 & 0.052 & 8.33 \\
\bottomrule
\end{tabular}%
}
\end{table}

To isolate the role of token identity from absolute position, we evaluate otherwise identical content after replacing one token with EOS at the specified relative position. The strongest sink is selected by availability-normalized causal column mass, i.e., the column mass divided by the number of queries that can attend to each position. Unlike the other analyses, we use this normalization here so that sink strength is comparable across insertion positions. Since it divides by only a few queries near the end of the sequence, we restrict candidates to positions with at least $T/8$ future queries. For the no-EOS condition, the target is position $0$. Table~\ref{tab:appendix_eos_position} reports baseline and gated results.
A first-position sink arises even without EOS. When EOS is inserted, the strongest attention sink largely follows it in both models: the inserted EOS is the strongest sink in $88$--$95\%$ of ungated and $71$--$99\%$ of gated evaluations. Gating therefore leaves the sink location largely unchanged but reduces its strength: target attention mass decreases by $2.8$--$5.4\times$ and target gradient amplification by $2.5$--$3.8\times$. When EOS appears away from the start, position $0$ is no longer the dominant attention sink but retains a residual value-gradient signature, with amplification of $24.47$--$28.84$ reduced to $6.00$--$6.18$ by gating. Thus, sink location is affected by both sequence position and special-token identity, whereas gate-induced attenuation persists under these changes.

\section{Extensions Beyond Post-SDPA Gating}
\label{app:intervention_extensions}

We use the spectral and gradient diagnostics to characterize intervention classes beyond post-SDPA gating. We distinguish direct empirical comparisons of attention operators from theoretical comparisons of local gradient pathways and gate placements.

\subsection{Softpick: An Attention-Operator Comparison}
\label{app:softpick}

The raw token-space spectral identity
\begin{equation}
C_Y^{\mathrm{raw}}=\frac{1}{d}AVV^\top A^\top
\label{eq:appendix_any_attention_operator}
\end{equation}
applies to any attention operator $A$, including Softpick~\citep{zuhri-etal-2026-softpick}. The same token-space spectral diagnostics can therefore characterize normalization-based sink mitigations even though their mechanisms need not match post-SDPA gating. In contrast, the deterministic mode contraction result in Theorem~\ref{thm:sink_mode_contraction} concerns an already-formed attention output acted on by a downstream gate. Here we separate the effect of changing the normalization operator at fixed learned scores from that of training a model using the alternative operator.

\paragraph{Protocols and metrics.} In a frozen-operator swap, we hold Qwen-1B baseline scores $S=QK^\top/\sqrt d$ and values $V$ fixed, changing only Softmax to the Softpick operator. In a separate trained-model comparison, we analyze released Vanilla and Softpick-340M checkpoints\footnote{\url{https://huggingface.co/collections/zaydzuhri/softpick}}. Sink-bearing layers are selected independently of spectral outcomes using only the Vanilla raw-score geometry: the first position must be the raw-logit argmax for more than $50\%$ of queries. This yields layers $8$--$28$ for Qwen-1B and $8$--$24$ for the matched $340$M models.

Since Softpick need not produce attention rows summing to $1$, we report both absolute top-$k$ column mass and a normalized fraction of routed mass. With
\begin{align}
M_j&=\sum_{t\ge j}A_{tj},
&J_{\mathrm{top}k}&=\operatorname{TopK}_j(M_j),\\
m_{\mathrm{top}k}&=\frac{1}{kT}\sum_{j\in J_{\mathrm{top}k}}M_j,
&R_{\mathrm{top}k}&=\frac{\sum_{j\in J_{\mathrm{top}k}}M_j}{\sum_j M_j}.
\label{eq:appendix_softpick_mass}
\end{align}
Both statistics use the same top-$k$ columns per head ($k=8$), ranked by the column mass $M_j$. This is the quantity used for the sink subspace in Section~\ref{sec:forward_empirical} up to the constant factor $1/T$, but with a fixed $k$ instead of the adaptive threshold. Ranking by $M_j$ avoids selecting late positions whose few valid queries make availability-normalized mass unreliable. The leading-outlier ratios below use the feature-standardized covariance and MP edge $\lambda_+$ employed in the main spectral analysis.

\begin{table}[t]
\centering
\caption{\textbf{Comparison of Softmax and Softpick under frozen-operator and trained-model settings.} The frozen control uses Qwen-1B without retraining, while the trained comparison uses released Vanilla/Softpick-340M checkpoints. Effects indicate percentage decreases (frozen) or Vanilla-to-Softpick ratios (trained).}
\label{tab:appendix_softpick}
\small
\setlength{\tabcolsep}{4pt}
\begin{tabular}{lrrr|lrrr}
\toprule
\multicolumn{4}{c}{Operator Swap (Qwen-1B, layers 8--28)}
& \multicolumn{4}{c}{Trained models (340M, layers 8--24)} \\
\midrule
Metric & Softmax & Softpick & Effect
& Metric & Vanilla & Softpick & Effect \\
\midrule
$m_{\mathrm{top}k}$ & 0.0767 & 0.0754 & $1.7\%$
& $m_{\mathrm{top}k}$ & 0.0786 & 0.0032 & $24.3\times$ \\
$\lambda_1/\lambda_+$ & 19.86 & 19.69 & $0.9\%$
& $\lambda_1/\lambda_+$ & 15.39 & 13.74 & $1.12\times$ \\
- & - & - & -
& $R_{\mathrm{top}k}$ & 0.629 & 0.199 & $3.2\times$ \\

\bottomrule
\end{tabular}
\end{table}

\paragraph{Results.}
In the frozen swap, all reported diagnostics change by less than $2\%$. The first token is already the largest raw score for $79$--$94\%$ of queries in the selected Qwen layers, so replacing Softmax after training largely preserves the same dominant source. In contrast, the Softpick-trained model selects the first token as raw-logit argmax for fewer than $1\%$ of queries and exhibits a negative first-token logit margin. Its top-$k$ mass and routed-mass concentration decrease by $24.3\times$ and $3.2\times$, respectively, relative to the trained Vanilla model.
The leading-outlier ratio changes more modestly, from $15.39$ to $13.74$. Thus, Softpick substantially reduces sink concentration, although structured spectral modes in the representation largely remain. The strong trained effect is primarily associated with the score geometry learned under Softpick, rather than a large post-hoc effect of swapping normalizers on frozen Softmax-trained scores. These observations distinguish upstream operator-mediated prevention of sink formation from downstream contraction of an already-formed sink contribution. The algebraic covariance framework also accommodates other attention operators, including Elastic-Softmax~\citep{fu2026attention}, but the empirical comparison here is limited to Softpick.

\subsection{Gradient-Valve Interpretation of Value-State Gated Attention}
\label{app:vga}

The local value-gradient interpretation can also be applied to Value-State Gated Attention (VGA)~\citep{bu2025value}, while distinguishing source-conditioned and query-conditioned gates. Using column-vector notation, the local value Jacobian of a post-SDPA gate for source token $s$ and query token $t$ is
\begin{equation}
J^{\mathrm{post}}_{t,s}
=\frac{\partial\widetilde y_t}{\partial v_s}
=a_{ts}\Gamma_t,
\qquad
\nabla_{v_s}\mathcal L
=\sum_{t\ge s}a_{ts}\Gamma_t^\top\delta_t,
\quad\delta_t:=\nabla_{\widetilde y_t}\mathcal L.
\label{eq:appendix_post_jacobian}
\end{equation}
Here $\Gamma_t$ is the diagonal, query-dependent sigmoid gate. Since $\|\Gamma_t\|_{\mathrm{op}}\leq1$, each individual value-path contribution is directly contracted at fixed upstream gradient.

For a simplified scalar VGA gate, let
\begin{equation}
\phi(v_s)=g_sv_s,
\qquad g_s=\operatorname{sigmoid}(w_g^\top v_s).
\label{eq:appendix_vga_gate}
\end{equation}
Because the gate depends on its own value state, $\nabla_{v_s}g_s=g_s(1-g_s)w_g$, so its Jacobian is
\begin{equation}
J_\phi(v_s)
=\frac{\partial\phi(v_s)}{\partial v_s}
=g_sI+g_s(1-g_s)v_sw_g^\top.
\label{eq:appendix_vga_jacobian}
\end{equation}
Consequently,
\begin{equation}
J^{\mathrm{VGA}}_{t,s}=a_{ts}J_\phi(v_s),
\qquad
\nabla_{v_s}\mathcal L
=\sum_{t\ge s}a_{ts}J_\phi(v_s)^\top\delta_t.
\label{eq:appendix_vga_gradient}
\end{equation}
The first term, $g_sI$, is the gated content path; the second, rank-one term, $g_s(1-g_s)v_sw_g^\top$, is a self-regulatory path because the source value also controls its gate. Both architectures therefore share the local form
\begin{equation}
\nabla_{v_s}\mathcal L
=\sum_{t\ge s}a_{ts}B_{t,s}^\top\delta_t,
\quad
B_{t,s}=
\begin{cases}
\Gamma_t,&\text{post-SDPA gating},\\
J_\phi(v_s),&\text{VGA}.
\end{cases}
\label{eq:appendix_general_valve}
\end{equation}
High attention-column mass does not necessarily imply proportionally high value-gradient mass when these local Jacobians suppress the corresponding contributions.

The conditioning and guarantees differ. Post-SDPA gating acts separately at each receiving query after aggregation and has the unconditional local operator-norm bound $\|\Gamma_t\|_{\mathrm{op}}\leq1$ for its value path. VGA gates each source value before aggregation and includes an additional derivative term. In general,
\begin{equation}
\|J_\phi(v_s)\|_{\mathrm{op}}
\leq g_s+g_s(1-g_s)\|v_s\|_2\|w_g\|_2,
\label{eq:appendix_vga_bound}
\end{equation}
so $g_s<1$ alone does not guarantee operator-norm contraction at intermediate gate values. In the closed-gate limit $g_s\to0$, with the remaining factors bounded,
\begin{equation}
g_sI\to0,
\qquad g_s(1-g_s)v_sw_g^\top\to0,
\qquad J_\phi(v_s)\to0.
\label{eq:appendix_vga_closed}
\end{equation}
Thus, complete severing is a limiting case of local Jacobian suppression, not a requirement for substantial sink-gradient attenuation. Our fixed-upstream-gradient counterfactual already shows considerable attenuation under ordinary sigmoid post-SDPA gates. This comparison is mechanistic; it is not an experimental claim that VGA and post-SDPA gating have identical training dynamics.

\subsection{Mechanistic Comparison of Gate Placements}
\label{app:gate_placements}

The five gate placements studied in gated-attention architectures intervene at different points along the sink pathway. For one head, let $y_t=\sum_j a_{tj}v_j$; multi-head outputs are mixed as $o_t=\sum_hW_O^{(h)}y_t^{(h)}$. Denote multiplicative gates at the query, key, value, attention-output, and mixed-output positions by $\Gamma^Q,\Gamma^K,\Gamma^V,\Gamma^Y,$ and $\Gamma^O$:
\begin{align}
\text{G4 (query):}\quad&q_t\mapsto\Gamma_t^Qq_t,\\
\text{G3 (key):}\quad&k_j\mapsto\Gamma_j^Kk_j,\\
\text{G2 (value):}\quad&v_j\mapsto\Gamma_j^Vv_j,\\
\text{G1 (post-SDPA):}\quad&y_t\mapsto\Gamma_t^Yy_t,\\
\text{G5 (post-output):}\quad&o_t\mapsto\Gamma_t^Oo_t.
\label{eq:appendix_five_placements}
\end{align}
For sink source $s$, the attention column creates a rank-one forward contribution $Y_s=A_{:,s}v_s^\top$, while the same attention weights accumulate value-path gradients across receiving queries.

Query/key gating (G4/G3) intervenes before the attention matrix is formed; the coupled changes to logits and normalization mean that attenuating queries or keys does not guarantee monotone attenuation of a particular sink column. Value gating (G2) directly transforms the source contribution to
\begin{equation}
Y_s^{\mathrm{G2}}
=A_{:,s}(\Gamma_s^Vv_s)^\top,
\label{eq:appendix_g2_sink}
\end{equation}
which, for a fixed attention matrix, retains the same token-space column direction $A_{:,s}$. It also uses a single source-side gate for all queries receiving that value.

Post-SDPA gating (G1) instead transforms the sink contribution at each receiving query:
\begin{equation}
y_{t,s}^{\mathrm{G1}}=a_{ts}\Gamma_t^Yv_s,
\qquad
\nabla_{v_s}\mathcal L
=\sum_ta_{ts}(\Gamma_t^Y)^\top\delta_t.
\label{eq:appendix_g1_sink}
\end{equation}
It can therefore suppress the same source differently across receiving queries, after the sink contribution is formed but before head mixing. The contraction theorem directly describes its forward attenuation, and its local value Jacobian attenuates query-to-source contributions before they accumulate at a sink value. G5 gates only after the head-specific outputs have been mixed by $W_O$; contraction in that mixed output space does not guarantee contraction of each individual sink-carrying head or of the corresponding projected value path.

\section{Limitations and Future Work} \label{app:limitation}
Our analysis is intended as a mechanism-level account rather than a complete theory of gated training dynamics. The BBP interpretation relies on an idealized isotropic reference model; learned feature-wise gates also reshape the noise component, so the resulting covariance need not be a standard spiked Wishart model. Similarly, our gradient result is local: it shows that gating contracts value-path gradients for fixed attention weights and upstream gradients, but does not imply that all gradients in a trained gated model are uniformly smaller. Empirically, gating reduces sink-subspace alignment and dominant spectral outliers, but not all spectral suppression is explained by sink-induced modes. Learned gates may also affect non-sink structured directions, head-specific outliers, and anisotropic noise. 
Thus, our findings provide a mechanism-level account within the evaluated settings, while broader generalization remains to be established.

\section{Broader Impact} \label{app:broader_impact}
This work contributes to the mechanistic understanding of gated attention and attention-sink mitigation in Transformer language models. By connecting sink-induced spectral structure with value-gradient attenuation, our analysis may help develop more stable, interpretable, and efficient architectures, particularly for large-scale and long-context training. The societal impact of this work is indirect. We do not introduce a new deployed system or task-specific capability, but architectural insights that improve training stability and scalability may contribute to stronger foundation models. Such improvements can support beneficial applications, but may also amplify risks associated with misuse or unsafe deployment. These risks should be addressed through appropriate evaluation, safety, and governance practices alongside architectural research.

\end{document}